\documentclass[twoside,11pt]{article}

\usepackage{jmlr2e}
\newtheorem{assumption}[theorem]{Assumption}

\usepackage{microtype}
\usepackage{graphicx}
\usepackage{subcaption}
\usepackage{booktabs}
\usepackage{xcolor}
\usepackage{hyperref}
\usepackage{float}

\usepackage{amsmath}
\usepackage{amssymb}
\usepackage{adjustbox}

\usepackage{mathtools}
\usepackage{multirow}
\usepackage{multicol}
\usepackage{makecell}
\usepackage{xcolor}
\usepackage{arydshln}
\usepackage{algorithm}
\usepackage{algorithmic}
\usepackage{enumitem}
\usepackage{wrapfig}

\usepackage{subcaption}

\usepackage[capitalize,noabbrev]{cleveref}
\makeatletter
\renewenvironment{proof}[1][Proof]{%
  \par
  \noindent
  {\bf #1\ } 
}{%
  \hfill\BlackBox\\[2mm]
}
\makeatother

\usepackage[normalem]{ulem}

\newcommand{\param}{\theta}            
\newcommand{\wdCoeff}{\lambda}         
\newcommand{\renyiOrder}{\alpha}       
\newcommand{\smooth}{\beta}            
\newcommand{\penDecay}{\gamma}         
\newcommand{\epsPriv}{\epsilon}        
\newcommand{\dpDelta}{\delta}          
\newcommand{\penCoeff}{\rho}           
\newcommand{\noiseStd}{\sigma_{\text{noise}}}         
\newcommand{\noiseLStd}{\sigma}   

\newcommand{\paramVec}{\bm{\theta}}    
\newcommand{\LagMult}{\bm{\pi}}        
\newcommand{\auxParam}{\bm{\theta}_k}  
\newcommand{\penVec}{\bm{\rho}}        

\newcommand{\effStep}{\widehat{\eta}}  
\newcommand{\minCurv}{\tau}   

\newcommand{\Dset}{\mathcal{D}}        
\newcommand{\Aset}{\mathcal{A}}        
\newcommand{\loss}{R_{\text{erm}}}     

\newcommand{\lossk}{R_k}              
\newcommand{\Lag}{\mathcal{L}}         
\newcommand{\Pset}{\mathcal{P}}        
\newcommand{\PiSet}{\Pi}               
\newcommand{\Ndist}{\mathcal{N}}       

\newcommand{\clipBound}{C}             
\newcommand{\numAux}{K}                
\newcommand{\numIter}{M}               
\newcommand{\numEpochs}{E}             

\newcommand{\prox}[2]{\text{Prox}_{#1,#2}} 
\newcommand{\lipF}{L_F}                
\newcommand{\lipT}{L_T}                
\newcommand{\lsiConst}{c}              
\newcommand{\rdpEps}{\varepsilon}      

\newcommand{\lrGlobal}{\eta_{\param}}
\newcommand{\lrAux}{\eta_{k}}
\newcommand{\rhoK}{\rho_{k}}

\def\vtheta{{\bm{\theta}}}
\def\vpi{{\bm{\pi}}}
\def\E{\mathbb{E}}

\DeclareMathOperator*{\argmin}{arg\,min}

\usepackage{amsmath,amsfonts,bm}

\def\eqref#1{equation~\ref{#1}}

\def\1{\bm{1}}

\def\rmI{{\mathbf{I}}}

\def\rmV{{\mathbf{V}}}

\def\vx{{\bm{x}}}

\def\vz{{\bm{z}}}

\DeclareMathAlphabet{\mathsfit}{\encodingdefault}{\sfdefault}{m}{sl}
\SetMathAlphabet{\mathsfit}{bold}{\encodingdefault}{\sfdefault}{bx}{n}

\def\gA{{\mathcal{A}}}

\def\gD{{\mathcal{D}}}

\def\gF{{\mathcal{F}}}

\def\gL{{\mathcal{L}}}
\def\gM{{\mathcal{M}}}
\def\gN{{\mathcal{N}}}

\def\gP{{\mathcal{P}}}

\def\gS{{\mathcal{S}}}

\def\sP{{\mathbb{P}}}

\newcommand{\reg}{\lambda}                    

\usepackage{lastpage}
\jmlrheading{27}{2026}{1-\pageref{LastPage}}{XX/XX; Revised XX/XX}{XX/XX}{XX-XXXX}{Ding Chen, Haochen Luo, Xiaofei Wang, Chen Liu.}
\ShortHeadings{title}{Chen, Wang, Liu}

\ShortHeadings{DP Training Under the Hidden State Assumption}{Chen, Luo, Wang, Liu.}
\firstpageno{1}

\def\precondition{\rmV_k}

\begin{document}

\title{Differentially Private Neural Network Training \\ Under the Hidden State Assumption}

\author{\name Ding Chen$^{1}$ \email ding.chen@my.cityu.edu.hk
        \AND
        \name Haochen Luo$^1$ \email chester.hc.luo@my.cityu.edu.hk
        \AND
        \name Xiaofei Wang$^1$ \email xwang4279-c@my.cityu.edu.hk
        \AND
        \name Chen Liu$^1$ \email chen.liu@cityu.edu.hk \\
        \addr $^1$ Department of Computer Science, City University of Hong Kong \\
        83 Tai Chee Ave, Kowloon Tong, Hong Kong, China \\
        }

\editor{}

\setlength{\parindent}{0pt}
\setlength{\parskip}{0.5em}

\maketitle

\begin{abstract} %
Current differentially private learning paradigms face a severe utility
bottleneck: DP-SGD degrades performance through noise accumulation over
training steps, while aggregation-based approaches such as PATE suffer from data inefficiency due to disjoint data partitioning. We propose \textbf{Differentially Private Decoupled Training (DP-DT)}, a framework that decouples representation learning from privacy enforcement. DP-DT confines noise injection to the weight aggregation stage and employs auxiliary models, continuously synchronized with a global model, to perform noise-free feature extraction on private data shards. We prove that DP-DT converges globally to a limit point under non-convex objectives, via a Lyapunov potential analysis combined with the Kurdyka--\L{}ojasiewicz property. Under the Hidden State Assumption, where adversaries observe only the final published model, we further prove that DP-DT's privacy loss can converge to a
constant bound rather than accumulating with iterations. Empirical results across vision and language benchmarks confirm that DP-DT significantly mitigates utility degradation, achieving state-of-the-art privacy-utility trade-offs.

\end{abstract}


\section{Introduction}
Deep neural networks have demonstrated exceptional performance and are now integral to applications such as face recognition \cite{he2016deep} and large language models \cite{kandpal2022deduplicating}. However, their success relies heavily on large-scale training data, which introduces serious privacy risks. Notably, research \cite{shokri2017membership} indicates that these models can memorize training data, allowing adversaries to extract sensitive information from the model parameters. 

To address data leakage, Differential Privacy (DP) provides a theoretical framework to bound the privacy loss of individuals. Due to its post-processing property and quantifiable guarantees, DP has become the de facto standard for protecting machine learning models. Generally, privacy-preserving training algorithms fall into two main categories: 

\textbf{Gradient Perturbation(e.g. DP-SGD)} Gradient perturbation methods, such as DP-SGD \cite{abadi2016deep}, enforce differential privacy guarantees by injecting calibrated noise into clipped gradients. This mechanism inherently couples privacy preservation with the gradient-based learning process. However, this coupling introduces a severe utility-privacy trade-off: since neural network training requires many iterations, the privacy budget allocated to each step must be extremely small, necessitating the addition of large noise that significantly degrades model performance.

\textbf{Output aggregation(e.g. PATE)} The \textit{sample and aggregation} framework provides the DP guarantee to the final output rather than the training process. It trains multiple models on disjoint subsets of the private data and applies the DP guarantee only during the aggregation of their predictions. However, its utility is strictly constrained by the data volume. Splitting the dataset into many disjoint parts drastically reduces the training samples available for each teacher model, leading to weak individual predictors. Consequently, to achieve acceptable performance, this framework heavily relies on public unlabeled data to assist training.

To address these fundamental challenges, we propose \textbf{Differentially Private Decoupled Training (DP-DT)}. Our core motivation is to separate the \textit{representation learning} capability from the \textit{privacy-preserving mechanism}. Specifically, DP-DT maintains two distinct sets of parameters:
\begin{enumerate}
    \item \textbf{Auxiliary Models:} These models are trained on disjoint private data shards without noise injection, serving as ``clean learners'' for precise data features.
    \item \textbf{Global Model:} A single global model is maintained and updated by coordinating with the auxiliary models through a privacy-preserving consensus scheme, ensuring the final release remains differentially private.
\end{enumerate}

Crucially, DP-DT employs an \textbf{iterative coordination protocol} by using the Alternating Direction Method of Multipliers(ADMM) algorithm \cite{boyd2011distributed}. Specifically, In each round, auxiliary models are first re-initialized with global parameters to facilitate knowledge sharing across data shards, mitigating the inefficiency of one-off aggregation methods like PATE. Next, these models train on private data without noise injection to capture precise features. Finally, the global model aggregates these auxiliary states via a privacy-preserving consensus mechanism. This design avoids the utility degradation associated with gradient noising in DP-SGD by confining privacy mechanisms to the aggregation step. We prove DP-DT can converge globally to a limit point under non-convex objectives.

Furthermore, DP-DT inherently aligns with the \textit{Hidden State Assumption (HSA)}, which is proposed by recent works ~\citep{feldman2018privacy,altschuler2022privacy,ye2022differentially} due to its benign property in privacy loss analysis. Specifically, HSA is a more realistic threat model where the adversary observes only the final published model rather than intermediate training states. This contrasts with standard DP-SGD analysis based on \textit{composition theorems}, which assume \textit{continuous adversarial access} to the entire optimization trajectory. The composition theorems are inherently pessimistic, as they accumulate privacy costs for intermediate updates that are never released. Prior works~\citep{feldman2018privacy,altschuler2022privacy,ye2022differentially} have attempted to apply HSA to DP-SGD; their research indicates that privacy loss under the HSA can converge to a stable bound, while the privacy loss under the composition theorem depends on the number of iterations. However, the benign property relies on strict \textit{convexity and smoothness} assumptions, which prevents its application in deep learning. In contrast, our DP-DT framework overcomes these limitations. By proposing a more general privacy accounting method under the HSA, we theoretically prove that under our framework, the privacy loss \textit{converges} to a tight constant bound, making HSA effective for general non-convex deep learning.

We summarize the contribution of this paper as follows. \textbf{(1) Decoupled Framework:}  We propose \textit{Differentially Private Decoupled Training (DP-DT)}, a framework that decouples feature learning from privacy noise. By confining noise injection solely to the weight aggregation stage and utilizing auxiliary models that are continuously synchronized with the global model to perform noise-free representation extraction, DP-DT substantially improves the utility of the private model. Convergence analysis indicates that DP-DT can converge to a limit point under non-convex objectives. \textbf{(2) Generalized Hidden State Accounting:} We derive a rigorous privacy accountant for DP-DT under the HSA, theoretically proving that the privacy loss converges to a stable bound. To the best of our knowledge, this \textbf{represents the first extension} of HSA's applicability to deep learning.
\textbf{(3) Empirical Superiority:}  Experiments on CIFAR-10 and CIFAR-100 demonstrate that DP-DT significantly outperforms baselines such as DP-SGD and PATE and achieves better privacy-utility trade-off.

\section{Related Work}
\label{sec:related_work}
\textbf{Differential Privacy Machine Learning.} 
While initially formulated for statistical database queries~\citep{dwork2006differential}, DP has evolved into the standard framework for trustworthy machine learning. 
Existing approaches predominantly follow two strategies. 
The dominant method, \textbf{Gradient Perturbation} (e.g., DP-SGD~\citep{song2013stochastic}), secures training by injecting noise into clipped gradients. While rigorous, it inherently couples privacy with learning dynamics; the cumulative noise required over thousands of iterations often overwhelms gradient signals, strictly bounding the privacy-utility trade-off~\citep{jayaraman2019evaluating}. Some recent work tries to mitigate the trade-off by projecting the gradient to subspace \cite{yu2021not,mulrooney2025memory}.
Conversely, the \textbf{Sample-and-Aggregation} framework \cite{nissim2007smooth} (e.g., PATE) provides DP guarantee to the final output. Specifically, S\&A framework decomposes a dataset into disjoint subsets, computes a target function on each subset independently, and subsequently aggregates individual results with DP guarantee. The PATE framework \cite{papernot2016semi,papernot2018scalable} applies this framework in deep learning. They first train teacher models on disjoint data partitions. Then, they apply a noisy voting mechanism to aggregate predictions, which are used to train a student model. Although this avoids direct noise injection during representation learning, it suffers from severe \textit{data inefficiency}: partitioning the dataset dilutes the signal for individual teachers, necessitating large-scale public data for knowledge distillation to recover performance.

\textbf{Differential Privacy Accounting.} 
Tracking privacy budget consumption is fundamental to DP mechanisms.
Early work relied on loose composition theorems~\citep{dwork2006differential}, later refined by the Moments Accountant (MA)~\citep{abadi2016deep}, R\'enyi DP~\citep{mironov2017renyi} and Gaussian DP~\citep{dong2022gaussian} for tighter bounds. 
However, these methods inherently assume a transparent optimization process where all intermediate states are visible to adversaries. Consequently, privacy loss increases with the number of iterations, limiting training duration. In practice, while the adversary typically lacks access to all intermediate states, the released model itself is often accessible. Thus, recent studies~\citep{feldman2018privacy,  chourasia2021differential, altschuler2022privacy,ye2022differentially} explore privacy accounting under the \textit{Hidden State Assumption} (HSA), where intermediate dynamics are concealed. 
The key finding is that privacy loss can \textit{converge} rather than accumulate—but only if the loss function satisfies strict properties like strong convexity and $\beta$-smoothness. These conditions are incompatible with the non-convex nature of deep learning. 
While variants like \citet{chien2024convergent} relax this to $(L,\wdCoeff)$-H\"older continuity, they depend on estimating complex continuity constants, leaving a gap for a generalized, practical accountant for deep neural networks.



\textbf{Augmented Lagrangian Method.} While SGD dominates unconstrained optimization, the Augmented Lagrangian Method (ALM)~\citep{hestenes1969multiplier} remains pivotal for constrained learning problems~\citep{zhang2018systematic}. 
To enable parallelization, the Alternating Direction Method of Multipliers (ADMM)~\citep{boyd2011distributed} was introduced to decompose global objectives into local sub-problems. 
However, classical ADMM relies on sequential updates (Gauss-Seidel) and computationally expensive exact minimization, rendering it impractical for high-dimensional neural networks. 
To overcome these bottlenecks, modern variants such as the \textit{Auxiliary Problem Principle} (APP)~\citep{zhu2024first} and Linearized ADMM~\cite{liu2019linearized} employ first-order approximations with proximal terms. Moreover, ADMM now is also widely used in quantization~\citep{ye2018unified} and federative learning~\citep{zhou2025preconditioned}.

\section{Methodology}

\subsection{Preliminary} \label{subsec:pre}

\textbf{Privacy Definitions.} \citet{dwork2006differential} proposes $(\epsPriv, \dpDelta)$-DP to quantitatively measure the privacy of an algorithm. That is, a random mechanism $\gM$ is $(\epsPriv, \dpDelta)$-DP if $\sP(\gM(\Dset) \in \gS) \leq e^\epsPriv \sP(\gM(\Dset') \in \gS) + \dpDelta$ holds for any neighboring datasets $\Dset$, $\Dset'$ and any possible output set $\gS$.

$(\epsPriv, \dpDelta)$-DP measures the bound in max-divergence between the output distributions of two neighboring inputs.
\citet{mironov2017renyi} proposes R\'enyi differential privacy (RDP) and use R\'enyi divergence to replace max-divergence.
Specifically, a mechanism $\gM$ is $(\renyiOrder , \rdpEps)$-RDP if $D_\renyiOrder (\gM(\Dset)||\gM(\Dset')) \leq \rdpEps$ holds for any neighboring datasets $\Dset$, $\Dset'$ where metric $D_\renyiOrder $ represents the R\'enyi divergence of order $\renyiOrder  > 1$.
For iterative algorithms like deep learning training, RDP enables tighter privacy accounting when it is converted back to $(\epsPriv, \dpDelta)$-DP.
Therefore, we adopt RDP for privacy accounting in this work.
For a given $\renyiOrder $, we call $\rdpEps$ the \textit{privacy loss} if the algorithm satisfies $(\renyiOrder , \rdpEps)$-RDP.

\textbf{Notation and Threat Model.} We consider training neural networks of general architectures.
Let $\Dset_{\text{train}}$ denote the entire training set partitioned into $\numIter$ mini-batches $\Dset_{\text{train}} = \{\Dset_1, \Dset_2, ..., \Dset_\numIter\}$.
We use $\ell(\cdot)$ as the loss function, and $f(\paramVec, \vx)$ to represent the output of a neural network parameterized by $\paramVec$ given the input $\vx$.
For any dataset $\Dset$, the empirical risk minimization aim to solve the following optimization problem:
\begin{equation}
\begin{aligned} \label{eq:erm}
    \min_{\paramVec} \loss(\paramVec; \Dset) = \frac{1}{|\Dset|}\sum_{(\vx, y) \in \Dset} \ell(f(\paramVec, \vx), y)
\end{aligned}
\end{equation}
When training deep neural networks, we usually only save the final model parameters as checkpoints.
By contrast, the training trajectory, optimization states, and other variables beyond the model parameters are generally not included when models are published. 
Consequently, for privacy analysis, we adopt \textit{the hidden state assumption}~\citep{feldman2018privacy,altschuler2022privacy, ye2022differentially}, which restricts the adversary's view strictly to the final published model parameters.

\subsection{Decoupling with Consistency}
\label{subsec:decoupling}
Our algorithm is inspired by the \textit{sample-and-aggregation framework (S\&A)}~\citep{nissim2007smooth},  which decomposes a dataset into disjoint subsets and releases the aggregated result with a DP guarantee. However, rather than employing S\&A for a one-shot model publication as in~\cite{nissim2007smooth}, we utilize this framework for the iterative training phase. This adaptation allows us to address a significantly more challenging scenario: enabling deep learning optimization that strictly adheres to the \textit{Hidden State Assumption}.

\begin{figure}[htbp] %
    \centering
    \begin{subfigure}[b]{0.35\columnwidth}
        \centering
        \includegraphics[width=\linewidth]{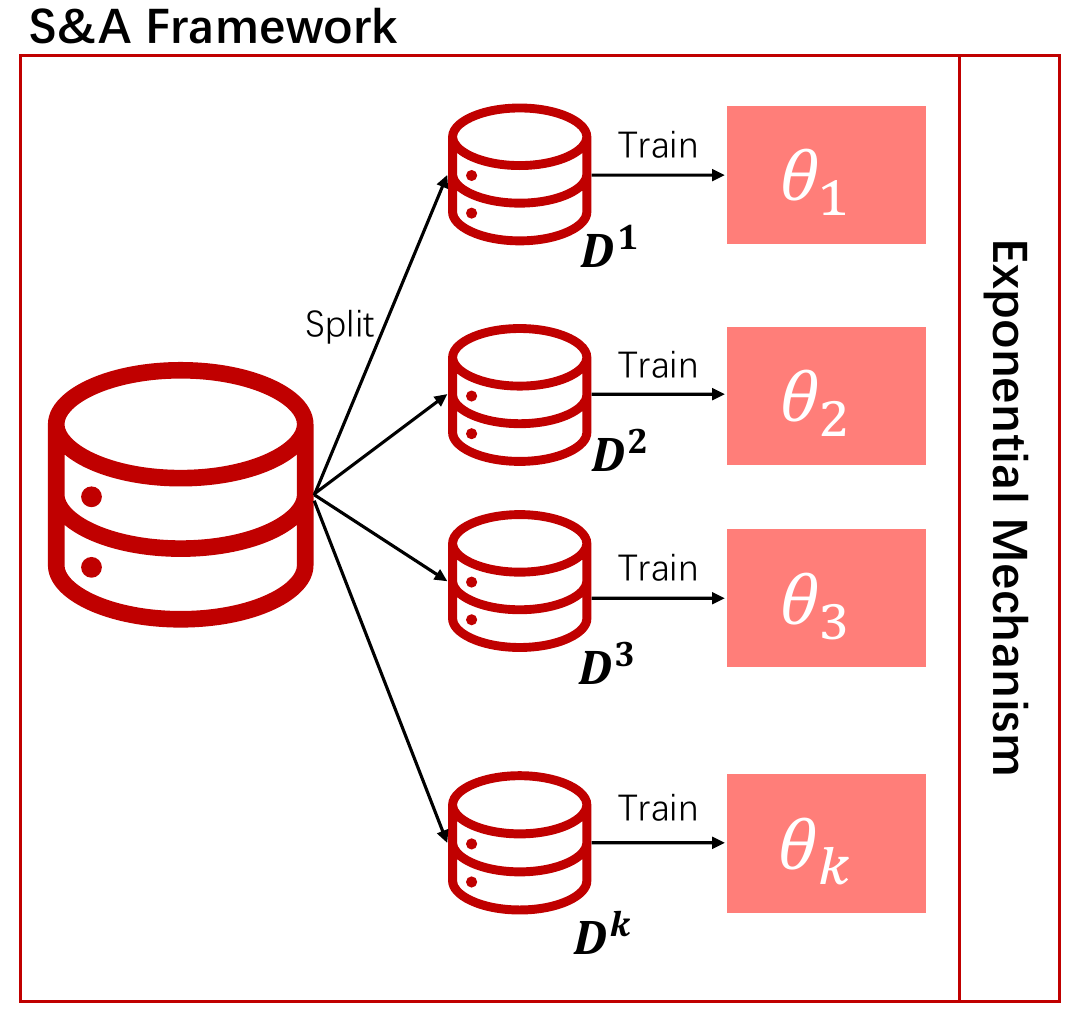}
        \caption{S\&A Framework}
        \label{fig:sa}
    \end{subfigure}
    \hfill 
    \begin{subfigure}[b]{0.6\columnwidth}
        \centering
        \includegraphics[width=\linewidth]{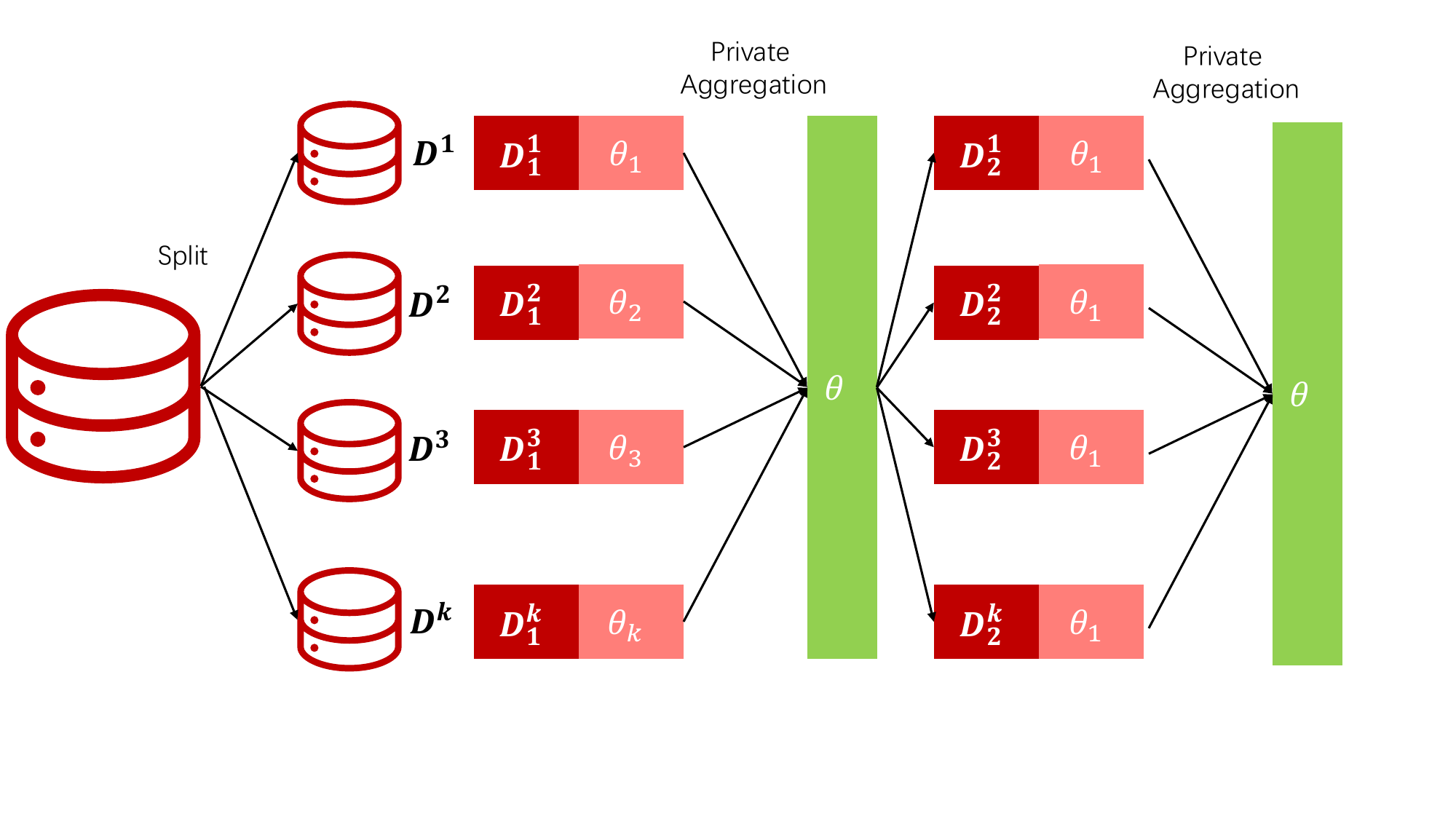}
        \caption{Decoupling with Consistency Strategy}
        \label{fig:dpdt}
    \end{subfigure}
    
    \caption{\small (a) The S\&A framework: the DP mechanism is applied to the final publication.  (b) Decoupling with Consistency Strategy: the DP mechanism is applied to the private aggregation. }
    \label{fig:sa_and_dc} 
\end{figure}

As demonstrated in Figure~\ref{fig:sa_and_dc}, we propose a \textit{decoupling with consistency} strategy: decoupling the privacy-sensitive training process from the final published model, with consistency between the two components enforced through soft constraints.
In addition to \textit{the global publishable model} parameterized by $\vtheta$, we maintain $K$ \textit{auxiliary models} of the same architecture parameterized by $\gA = \{\vtheta_k\}_{k=1}^K$, which update their own parameters independently.
Specifically, we partition each mini-batch data into $K$ disjoint micro-batches:
\begin{equation}
\begin{aligned}
    \gD_m = \{\gD_i^k\}_{k = 1}^K,~~~~m = 0, 1, ..., M
\end{aligned}
\end{equation}
Each auxiliary model $\vtheta_k$ is trained on its micro-batches $\{\gD_i^k\}_{i = 1}^M$, while the publishable model serves as the consensus of the auxiliary models.
This design achieves the best of both worlds: on one hand, the auxiliary models are not published, eliminating the need for privacy-enhancing noise during their optimization and thus boosting their utility; on the other hand, the publishable model is updated based on the parameters of the auxiliary models rather than directly from the training data, bypassing the challenges of intractable non-convex optimization and thereby facilitating privacy accounting under the hidden state assumption.
Formally, our goal is to minimize the aggregate empirical risk across all auxiliary models while enforcing consistency between the auxiliary parameters in $\Aset$ and the global parameters $\paramVec$. 
Therefore, training neural networks can be formulated as the following optimization problem:
\begin{equation}
\begin{aligned}
    \min_{\paramVec, \Aset} \quad & \frac{1}{\numIter\numAux} \sum_{m=1}^\numIter \sum_{k=1}^\numAux  \loss(\auxParam;\Dset_{m}^k) + r(\paramVec) \quad
    \text{s.t.}~~ \auxParam = \paramVec,~ \forall k \in \{1, \ldots, \numAux\}.
\end{aligned} 
\label{eq:constrained_prob}
\end{equation}
Problem~(\ref{eq:constrained_prob}) ensure the consistency between the publishable model and the auxiliary models with hard constraints. Therefore, it is equivalent to empirical risk minimization in (\ref{eq:erm}) with a regularization term $r(\paramVec)$ which is an extended-valued convex regularization function, possibly non-smooth. In practice, $r(\paramVec)$ is typically chosen as the $\ell_2$-norm penalty.

In our decoupling with consistency framework, we replace the hard constraints in (\ref{eq:constrained_prob}) with soft ones and consider its augmented Lagrangian function as below:
\begin{equation}
\begin{aligned}
\label{equ:alm_form}
    \Lag(&\paramVec,\Aset,\PiSet,\Pset) =  \frac{1}{\numIter\numAux} \sum_{m=1}^\numIter \sum_{k=1}^\numAux \loss(\auxParam;\Dset_{m}^k) + r(\paramVec) +\frac{1}{\numAux} \sum_{k=1}^\numAux \underbrace{\left[\langle \LagMult_k, \auxParam-\paramVec \rangle + \frac{\rhoK}{2} \|\auxParam-\paramVec\|^2\right]}_{h(\paramVec, \auxParam, \LagMult_k, \rho_k)} 
\end{aligned}
\end{equation}
where $\PiSet=\{\LagMult_k\}_{k=1}^\numAux$ is the set of Lagrangian multipliers, and $\Pset=\{\rhoK\}_{k=1}^\numAux$ is the set of penalty coefficients.
Based on the augmented Lagrangian function $\Lag$ in (\ref{equ:alm_form}), the optimization objective of the auxiliary model parameters $\auxParam$ includes not only the empirical risk minimization term, the regularization term but also proximity terms $h(\paramVec, \auxParam, \LagMult_k, \rho_k)$, which ensure that $\auxParam$ does not deviate significantly from the global publishable parameters $\paramVec$.
Furthermore, when optimizing the global publishable parameters $\paramVec$ with $\gA$, $\Pi$ and $\gP$ fixed, the function $\Lag$ becomes a convex form.
This property facilitates the privacy analysis detailed in the following sections.

\subsection{Differentially Private Decoupled Training (DP-DT)} \label{subsec:dpdt}



For brevity, we suppress the fixed parameters in the notation for $\Lag$ and $h$ defined in (\ref{equ:alm_form}).
For example, $\Lag(\paramVec)$ refers to $\Lag$ as a function of $\paramVec$, with other parameters held constant.
Given the properties of $\Lag$ described above, we adopt a variant of alternating direction method of multipliers (ADMM) \citep{zhu2024first} to solve the optimization problem (\ref{equ:alm_form}), which decomposes the complex problem into smaller, more manageable sub-problems.
In this context, we propose \textbf{differentially private decoupled training (DP-DT)}, which iteratively updates the parameters of $\Lag$ per iteration: alternatively updating the auxiliary model parameters $\Aset$, multipliers $\PiSet$, global publishable parameters $\paramVec$ and penalty coefficients $\Pset$.
The pseudo-code of DP-DT is outlined in Algorithm~\ref{alg:dp_decoupled}.

\begin{algorithm}[ht]
 \caption{Differential Private Decoupled Training (DP-DT)}
   \label{alg:dp_decoupled}
\begin{algorithmic}[1]
\STATE {\bfseries Input:} Dataset $\Dset_{\text{train}} = \{\Dset_m\}_{m = 1}^\numIter = \{\{\gD_m^k\}_{k = 1}^K\}_{m = 1}^\numIter$; number of auxiliary models $\numAux$ and epochs $\numEpochs$; learning rates $\lrGlobal$ for $\paramVec$ and $\{\lrAux\}_{k = 1}^\numAux$ for $\gA$; clipping threshold $\clipBound^{(t)}$; noise magnitude $\noiseStd$; penalty coefficient $\{\rhoK\}_{k = 1}^\numAux$ and their rate $\{\penDecay_k\}_{k = 1}^\numAux$.

\STATE {\bfseries Initialization:} Initialize parameters $\paramVec^{(0)}$ and $\{\auxParam^{(0)}\}_{k = 1}^\numAux$, multipliers $\{\LagMult_k^{(0)}\}_{k = 1}^\numAux$, global step index $t = 0$.

\FOR{epoch index $l = 1$ {\bfseries to} $\numEpochs$}
    \FOR{mini-batch index $m = 1$ {\bfseries to} $\numIter$}    
    \STATE \textcolor{gray}{\% Auxiliary Model and Multiplier Update can be in Parallel}
    \FOR{auxiliary model index $k = 1$ {\bfseries to} $\numAux$}
    \STATE Set $\auxParam^{(t)} = \paramVec^{(t)}$
    \STATE Compute gradient $\nabla_{\auxParam} \gL(\auxParam^{(t)}; \Dset_{m}^k)$
    \STATE Compute $\precondition^{(t)}$ based on adopted optimizer.
    \STATE Obtain the adaptive learning rate $\eta_k^{(t)}$ based on (\ref{eq:st})
    \STATE Update $\auxParam^{(t+1)}$ and $\LagMult_k^{(t+1)}$ based on (\ref{eq:update_aux})
    \ENDFOR

    \STATE \textcolor{gray}{\% Privacy-preserving Publishable Model Update}
    \STATE Update global model $\paramVec^{(t+1)}$ based on (\ref{equ:theta_update}).
    \STATE \textcolor{gray}{\% Penalty Coefficient Update}
    \STATE Update $\rhoK^{(t+1)} = \rhoK^{(t)} / \penDecay_k$ periodically
    \STATE $t = t + 1$
    \ENDFOR
\ENDFOR
\STATE {\bfseries Output:} The final published model $\paramVec^{(T)}$.
\end{algorithmic}
\end{algorithm}
\textbf{Top-Level Design.} Among the parameters in $\Lag$, only the global parameters $\paramVec$ are public and accessible to the adversary under the hidden state assumption, so we employ privacy-enhancing mechanisms when updating $\paramVec$. 
Considering that $r(\paramVec)$ may be non-smooth, we utilize noisy proximal gradient descent for updating $\paramVec$. 
To ensure differential privacy, it is essential to bound the gradient magnitude of the remaining components apart from $r(\paramVec)$ w.r.t. $\paramVec$, i.e., $\|\nabla_{\paramVec} \sum_{k = 1}^\numAux h(\paramVec, \auxParam)\|_2$ must be bounded. 
In this regard, we adopt two strategies: (1) reset the auxiliary model parameters to the global parameters before updating them, i.e., $\forall \auxParam \in \Aset$, we set $\auxParam \leftarrow \paramVec$; (2) apply clipping to the updates of the auxiliary model parameters and multipliers. Both strategies aim to prevent divergence and control the sensitivity which is related to $\|\auxParam - \paramVec\|$.

As demonstrated in Algorithm~\ref{alg:dp_decoupled}, each iteration of DP-DT involves several stages: alternative updates on auxiliary model parameters $\gA$, multipliers $\Pi$, publishable parameters $\paramVec$ and penalty coefficients $\gP$. Same as in Algorithm~\ref{alg:dp_decoupled}, we use superscript $\cdot^{(t)}$ to indicate the value of a variable in iteration $t$. The following are detailed discussions.

\textbf{Stage 1: Auxiliary Model and Multiplier Updates.} 

The parameters and the multiplier of each auxiliary model can be updated in parallel, as the global publishable parameter $\paramVec^{(t)}$ remains fixed during this stage, and the data used for each auxiliary model is disjoint. 
Without loss of generality, we focus on the $k$-th auxiliary model with parameters $\auxParam$ and the multiplier $\LagMult_k$. 
As outlined in Algorithm~\ref{alg:dp_decoupled}, $\auxParam$ is set to $\paramVec$ before its update, i.e. $\auxParam^{(t)} = \paramVec^{(t)}$.
Therefore, the local gradient w.r.t. $\auxParam$ is $\nabla_{\auxParam} \loss(\auxParam^{(t)};\Dset_m^k) + \LagMult_k^{(t)}$.
Considering the high dimensionality of the parameters $\auxParam$, we usually employ optimizers with pre-conditioners, like RMSProp or Adam~\citep{hinton2012neural, kingma2014adam}, to update $\auxParam$. 
For the multiplier $\LagMult_k$, we first apply $l_2$ clipping to control its magnitude and then perform dual ascent as in ADMM using the updated parameter $\auxParam^{(t + 1)}$.
Therefore, we have the following generalized update scheme:
\begin{equation} \label{eq:update_aux}
\begin{aligned}
    \auxParam^{(t + 1)} &= \auxParam^{(t)} - \Delta_k^{(t)} = \auxParam^{(t)} - \eta_k^{(t)} \precondition^{(t)} \nabla_{\auxParam} \gL(\auxParam^{(t)}) \\
    \LagMult_k^{(t + 1)} &= \widehat{\LagMult}_k^{(t)} + \rhoK^{(t)} (\auxParam^{(t+1)} - \paramVec^{(t)}) = \frac{\clipBound^{(t)} \cdot \LagMult_k^{(t)}}{\max \{C^{(t)}, \|\LagMult_k^{(t)}\|_2\}} - \rhoK^{(t)} \eta_k^{(t)} \precondition^{(t)} \nabla_{\auxParam} \gL(\auxParam^{(t)})
\end{aligned}
\end{equation}
where $\clipBound^{(t)}$, $\precondition^{(t)}$ and $\eta_k^{(t)}$ are the clipped bound, the inverse preconditioner and the learning rate, respectively.
In addition, $\Delta_k^{(t)}$ and $\widehat{\LagMult}_k^{(t)}$ represent the parameter update and the clipped multiplier, respectively.
The update rule in (\ref{eq:update_aux}) is compatible with preconditioned gradient-based optimizers in deep learning.
The inverse preconditioner $\precondition^{(t)}$ is a diagonal and positive definite matrix. In Adam and RMSProp, the diagonal of $\precondition^{(t)}$ is the inverse of the moving average of squared gradients. In optimizers without a preconditioner like pure SGD, $\precondition^{(t)}$ is the identity matrix.
Without the loss of generality, we assume $0 < \precondition^{(t)} \leq \minCurv^{-1} \rmI$ where $\minCurv$ indicates the minimum estimated curvature. This is a benign assumption, because in practice, optimizers usually set a minimum value for preconditioners to avoid zero division.



As discussed in the top-level design, it is necessary to bound $\frac{1}{\numAux} \|\nabla_{\paramVec} \sum_{k = 1}^{\numAux} h(\paramVec^{(t)}, \auxParam^{(t+1)})\|_2$ to ensure differential privacy for the publishable parameter $\paramVec$. 
A sufficient condition for this is that $\|\nabla_{\paramVec} h(\paramVec^{(t)}, \auxParam^{(t+1)})\|_2 \leq \clipBound^{(t)}$ holds for any $k$ where $C^{(t)}$ is the clipped bound in the update rule (\ref{eq:update_aux}).
Here, we use $\paramVec^{(t)}$, $\auxParam^{(t+1)}$, because $\paramVec$ is updated after $\auxParam$ in Algorithm~\ref{alg:dp_decoupled}.
We adjust the learning rate $\eta^{(t)}_k$ in the update rule (\ref{eq:update_aux}) to ensure the bounded sensitivity.
Specifically, the adaptive learning rate $\eta^{(t)}_k$ is set as the largest value under a predefined threshold $\eta_k$ to satisfy $\|\nabla_{\paramVec} h(\paramVec^{(t)}, \auxParam^{(t+1)})\|_2 \leq \clipBound^{(t)}$. 
we have the following equation.
\begin{equation}
\begin{aligned}
\|\nabla_{\paramVec} h(\paramVec^{(t)}, \auxParam^{(t+1)})\|_2 &= \|\rhoK^{(t)}(\paramVec^{(t)} - \auxParam^{(t+1)}) - \LagMult_k^{(t+1)}\|_2 = \left\|\widehat{\LagMult}^{(t)} - 2\rhoK^{(t)} \eta_k^{(t)} \precondition^{(t)} \nabla_{\auxParam} \gL(\auxParam^{(t)})\right\|_2 \\
    &= \left\|\frac{\clipBound^{(t)} \cdot \LagMult_k^{(t)}}{\max \{C^{(t)}, \|\LagMult_k^{(t)}\|_2\}} - 2\rhoK^{(t)} \eta_k^{(t)} \precondition^{(t)} \nabla_{\auxParam} \gL(\auxParam^{(t)})\right\|_2
\end{aligned}
\end{equation}
Therefore, the adaptive learning rate $\eta_k^{(t)}$ is selected by the optimization problem below.
\begin{equation}
\begin{aligned} \label{eq:st}
    \max_{\eta_k^{(t)} \in [0, \eta_k]} \eta_k^{(t)},~ s.t.~ \left\|\frac{\clipBound^{(t)} \cdot \LagMult_k^{(t)}}{\max \{C^{(t)}, \|\LagMult_k^{(t)}\|_2\}} - 2\rhoK^{(t)} \eta_k^{(t)} \precondition^{(t)} \nabla_{\auxParam} \gL(\auxParam^{(t)})\right\|_2 \leq \clipBound^{(t)}
\end{aligned}
\end{equation}
Problem (\ref{eq:st}) involves a quadratic constraint, making it computationally efficient to solve. 
Notably, since $\LagMult_k^{(t)}$ is clipped by its $l_2$ norm, the feasible set of Problem (\ref{eq:st}) is always non-empty, ensuring that the problem is well-defined.

In summary, when updating the auxiliary model $\auxParam$ and its corresponding multiplier $\LagMult_k$, we first solve Problem~(\ref{eq:st}) to find the adaptive learning rate and then update $\auxParam$ and $\LagMult_k$ by rules in (\ref{eq:update_aux}) and the selected optimizer.

\textbf{Stage 2: Privacy-preserving Publishable Model Update.} 

The update rule in Stage 1 ensures bounded sensitivity for publishable parameters $\paramVec$, i.e., $\frac{1}{\numAux}\|\nabla_{\paramVec} \sum_{k = 1}^{\numAux} h(\paramVec^{(t)}, \auxParam^{(t+1)})\|_2 < C^{(t)}$.
In Stage 2, we use Gaussian mechanism to enforce differential privacy guarantees on $\paramVec$.
Considering that the regularization term $r(\paramVec)$ may be non-smooth, we employ noisy proximal gradient descent to update $\paramVec$:
\begin{equation}
\begin{aligned} \label{equ:theta_update}
    \paramVec^{(t+1)} &= \prox{\effStep}{\text{r}} \left( \paramVec^{(t)} - \effStep^{(t)} \left( \nabla_{\paramVec} h(\paramVec^{(t)}) + \mathbf{z}^{(t)} \right) \right)~\text{where}~ h(\paramVec^{(t)}) = \frac{1}{\numAux} \sum_{k = 1}^\numAux h\left(\paramVec^{(t)}, \auxParam^{(t+1)}\right)
\end{aligned}
\end{equation}
where $\mathbf{z}^{(t)} \sim \Ndist(\mathbf{0}, \noiseStd^2 \mathbf{I})$ denotes the Gaussian noise and $\effStep^{(t)} = \frac{K\cdot \eta_{\paramVec}}{\sum_{k = 1}^K \rho^{(t)}_k}$ is the effective step size. Coefficient $\noiseStd$ and $\lrGlobal$ denote the noise magnitude and the global learning rate.
In addition, the smooth component $h(\paramVec)$ is a quadratic function whose Hessian can be analytically obtained as $(\frac{1}{\numAux}\sum_{k = 1}^\numAux \rhoK^{(t)}) \mathbf{I}$.
Leveraging this curvature information, we can accelerate the convergence while maintaining training stability by setting the step size $\effStep^{(t)} = \frac{K\cdot \eta_{\paramVec}}{\sum_{k = 1}^K \rho^{(t)}_k}$. 

\textbf{Stage 3: Penalty Coefficient Update.} 

We increase the penalty parameter exponentially during training to enforce the consensus constraint gradually:
\begin{equation}
\label{equ:rho_update}
    \rhoK^{(t+1)} = \rhoK^{(t)} / \penDecay_k, \quad \text{where } 0 < \penDecay_k \leq 1, \forall k \in \{1, \ldots, \numAux\}
\end{equation}
In practice, we periodically update the $\rhoK$ to stabilize the training trajectory.
In our design, as the training progresses, the consistency constraint is gradually enforced, which is crucial to improve the utility of the published model.

\section{Convergence Analysis}

\subsection{Proof Roadmap}
We outline the proof strategy before presenting the technical details:

\textbf{Section~\ref{sec:error_quantification}} bounds the mini-batch gradient deviation (Assumption~\ref{assume:bounded_deviation}) and the bias from clipping the dual variables (Lemma~\ref{lemma:clipping_bound}).
\textbf{Section~\ref{sec:descent_lemma}} then studies how each ADMM update affects the augmented Lagrangian (Lemmas~\ref{lemma:descent_local}--\ref{lemma:expected_descent_dp}).

\textbf{Section~\ref{sec:lyapunov}} addresses a structural obstacle: dual ascent (Lemma~\ref{lemma:dual_variable}), alongside residuals from mini-batch errors (Lemma~\ref{lemma:descent_local}) and DP noise (Lemma~\ref{lemma:expected_descent_dp}), prevents the Lagrangian from decreasing monotonically.
To resolve this, the section constructs a Lyapunov potential (Lemma~\ref{lemma:global_descent}) that absorbs the conflicting terms; Theorem~\ref{thm:robbins_siegmund} then guarantees almost-sure convergence of the potential and square-summability of the parameter updates.

\textbf{Section~\ref{sec:stationary}} addresses a second obstacle: square-summability implies $\|\boldsymbol{\theta}^{(t+1)}-\boldsymbol{\theta}^{(t)}\|\to 0$, but does not rule out the iterates drifting along a flat region without approaching a stationary point.
The section closes this gap via the Attouch--Bolte--Svaiter framework~\cite{attouch2013convergence}: Lemma~\ref{lemma:subgradient_bound} links the subgradient norm to the step size, and the Kurdyka--\L{}ojasiewicz property translates the energy descent into finite path length, from which convergence to a unique limit point follows (Theorem~\ref{thm:global_convergence}).

\subsection{Error Quantification}
\label{sec:error_quantification}

Before analyzing the global optimization trajectory, we must properly bound the system errors introduced by local mini-batch updates and dual variable clipping. We observe that Algorithm~\ref{alg:dp_decoupled} effectively decouples sensitivity control from the noise calibration process. Specifically, inspired by the work of~\cite{zhou2025preconditioned}, we derive the deterministic bound for the local gradient deviation as below.

\begin{assumption}[Bounded Gradient Deviation]
\label{assume:bounded_deviation}
There exists a constant $\zeta < \infty$ such that for all micro-batches $\gD_m^k$ and all auxiliary model parameters $\auxParam^{(t)}$ encountered during training, the stochastic gradient deviation from the full-batch gradient is uniformly bounded:
\begin{equation*}
    \|\nabla_{\auxParam} \loss(\auxParam^{(t)}; \gD^{k}_m) - \nabla_{\auxParam} \loss(\auxParam^{(t)}; \gD_{\text{train}})\| \leq \zeta.
\end{equation*}
This is mild in practice: the loss $\loss$ is continuously differentiable, the ADMM penalty term $(\rho_k/2)\|\auxParam - \paramVec\|^2$ and global weight decay $r(\paramVec)$ jointly prevent parameters from diverging, and common losses (e.g., cross-entropy with softmax) have uniformly bounded gradients for compact input domains.
\end{assumption}

Moreover, the algorithm bounds the global model's gradient sensitivity by adaptively controlling the learning rate of the auxiliary model and clipping the dual variables. We subsequently derive an algebraic bound for this clipping bias.

\begin{lemma}[Bound for Clipping Bias]
\label{lemma:clipping_bound}
Consider Algorithm~\ref{alg:dp_decoupled} and update rule~(\ref{eq:update_aux}), if $\{\clipBound^{(t)}\}_{t=0}^\infty$ is a monotonically non-increasing sequence, i.e., $\forall t \geq 1$, $\clipBound^{(t)} \le \clipBound^{(t-1)}$, then $\forall k$ and $\forall t \geq 1$, the multiplier satisfies $\|\LagMult_k^{(t)}\|_2 \le \clipBound^{(t-1)}$. 
Consequently, the bias term $\kappa_k^{(t)} :=\LagMult_k^{(t)} - \widehat{\LagMult}_k^{(t)} $ is bounded by $\clipBound^{(t-1)} - \clipBound^{(t)}$, i.e.,$\|\kappa_k^{(t)}\|_2 \le \clipBound^{(t-1)} - \clipBound^{(t)}$.
\end{lemma}

The proof of Lemma \ref{lemma:clipping_bound} is deferred to Appendix \ref{proof:clipping}. Lemma \ref{lemma:clipping_bound} bounds the clipping bias in Algorithm \ref{alg:dp_decoupled}. Specifically, if $\clipBound^{(t-1)} = \clipBound^{(t)}$, then $\kappa_k^{(t)} = \mathbf{0}$, indicating the bias introduced by the clipping function vanishes. In contrast to classical DP-SGD convergence analyses \citep{bu2024automatic}, which rely on restrictive assumptions such as bounded gradient variance to account for clipping errors, our analysis provides a more general framework by directly addressing the clipping dynamics.

\subsection{Descent Lemmas for Alternative Updates} \label{subsec:descent_lemma}
\label{sec:descent_lemma}
To prove the convergence of DP-DT in Algorithm~\ref{alg:dp_decoupled}, we first study how the updates in $\auxParam$, $\LagMult_k$, $\paramVec$ affect the loss objective function $\Lag$, respectively.

We start with $\auxParam$, based on Algorithm~\ref{alg:dp_decoupled} and update rule~(\ref{eq:update_aux}), the auxiliary model is reset to the global publishable model, i.e., $\auxParam^{(t+1)} = \paramVec^{(t)}$, and then updated by $\auxParam^{(t)} = \auxParam^{(t)} - \rho^{(t)}_k \precondition^{(t)} \nabla_{\auxParam} \Lag(\auxParam^{(t)})$.
In this context, we derive the descent lemma for $\auxParam$. 

\begin{lemma}[Descent Lemma for the Local Auxiliary Model]
\label{lemma:descent_local}
Consider Algorithm~\ref{alg:dp_decoupled} and update rule (\ref{eq:update_aux}) where the preconditioner $0 < \precondition^{(t)} \leq \minCurv^{-1}\rmI$, under Assumption~\ref{assume:bounded_deviation}, that is, the loss $\loss(\auxParam)$ is Lipschitz smooth with a constant $L$, if the step size for $\auxParam$ satisfies $\eta^{(t)}_k \leq \eta_k < \frac{2\minCurv}{L+\rho^{(t)}_k}$. Let $\Delta\auxParam^{(t)} = \auxParam^{(t+1)} - \paramVec^{(t)}$ and $e_k^{(t)} = \nabla\loss(\auxParam^{(t)};\gD_m^k) - \nabla\loss(\auxParam^{(t)})$. Then, we have:
\begin{equation} \label{eq:conclusion_descent_lemma}
    \Lag(\auxParam^{(t+1)}) - \Lag(\auxParam^{(t)}) \le \frac{1}{K}\big[-c^{(t)}_k\|\Delta\auxParam\|^2 - \langle e_k^{(t)}, \Delta\auxParam^{(t)}\rangle\big] \leq \frac{\zeta^2}{4K\cdot c^{(t)}_k} \quad \text{where}~c^{(t)}_k = \frac{\minCurv}{\eta^{(t)}_k} - \frac{L+\rho^{(t)}_k}{2}
\end{equation}
Here, we slightly abuse the notation $\auxParam^{(t + 1)}$, $\auxParam^{(t)}$: the former indicates the auxiliary parameter after its update but before being reset to the global publishable parameter in the next iteration, while the latter indicates the auxiliary before its update but after being reset to the global publishable parameter in the current iteration.
\end{lemma}

The proof of Lemma~\ref{lemma:descent_local} is deferred to Appendix~\ref{proof:descent_lemma}. Then, we study how the update in the dual variable $\LagMult_k$ affects the loss objective function $\Lag$.

\begin{lemma}[Bound for Dual Variable Ascent] \label{lemma:dual_variable}
    Based on Algorithm \ref{alg:dp_decoupled} and the update rule (\ref{eq:update_aux}), we have the following:
    \begin{equation}
        \begin{aligned}
            \Lag(\paramVec^{(t)}, \auxParam^{(t+1)}, \LagMult_k^{(t+1)}) - \Lag(\paramVec^{(t)}, \auxParam^{(t+1)}, \widehat{\LagMult}_k^{(t)}) &= \frac{\rho^{(t)}_k}{K}\|\auxParam^{(t + 1)} - \paramVec^{(t)}\|^2. \\
            \Lag(\paramVec^{(t)}, \auxParam^{(t + 1)}, \widehat{\LagMult}_k^{(t)}) - \Lag(\paramVec^{(t)}, \auxParam^{(t+1)}, \LagMult_k^{(t)}) &= \frac{1}{K} \langle \kappa^{(t)}_k, \auxParam^{(t+1)} - \paramVec^{(t)} \rangle.
        \end{aligned} \label{eq:dual_bound}
    \end{equation}
    where $\kappa^{(t)}_k =  \LagMult^{(t)}_k - \widehat{\LagMult}^{(t)}_k $ is the clipping bias term as defined in Lemma~\ref{lemma:clipping_bound}.
\end{lemma}

The proof of Lemma~\ref{lemma:dual_variable} is deferred to Appendix~\ref{proof:dual_variable}. Finally, we focus on the noisy proximal gradient descent, which is the update rule on the global publishable model $\paramVec$ in (\ref{equ:theta_update}). First, we notice that $h(\paramVec^{(t)})$, as defined in (\ref{equ:theta_update}), is a quadratic function of $\paramVec^{(t)}$, and specifically, $h(\paramVec^{(t)})$ is Lipschitz smooth with a constant $L_h = \frac{1}{K}\sum_{k = 1}^K \rho^{(t)}_k$. 

\begin{lemma}[Descent Lemma for Global Model Update] \label{lemma:expected_descent_dp}
    Consider Algorithm~\ref{alg:dp_decoupled} and update rule~(\ref{equ:theta_update}), we let $L_h = \frac{1}{K}\sum_{k = 1}^K \rho^{(t)}_k$, if the effective step size $\effStep^{(t)} \leq \frac{1}{L_h}$, i.e., $\eta_\theta \leq 1$, the regularization term $r(\paramVec)$ a proper, lower semi-continuous and convex, then the expected descent of the objective function conditioned on the filtration $\gF_t$ after the update rule (\ref{equ:theta_update}) is bounded by the following inequality where $d$ is the dimension of $\paramVec$.
    \begin{equation} \label{eq:expected_descent_dp}
        \mathbb{E}[\Lag(\paramVec^{(t+1)}) \mid \gF_t] \le \Lag(\paramVec^{(t)}) - \left( \frac{1}{\effStep^{(t)}} - \frac{L_h}{2} \right) \mathbb{E}[\|\paramVec^{(t+1)} - \paramVec^{(t)}\|^2 \mid \gF_t] + \effStep^{(t)} d \noiseStd^2
    \end{equation}
\end{lemma}

The proof of Lemma~\ref{lemma:expected_descent_dp} is deferred to Appendix~\ref{proof:expected_descent_dp}.




\subsection{Lyapunov Energy Descent and Subgradient Bound}
\label{sec:lyapunov}

The augmented Lagrangian $\Lag$ does not decrease monotonically under the ADMM updates. Apart from the dual variable ascent (Lemma~\ref{lemma:dual_variable}) term,  the mini-batch gradient deviation (Assumption~\ref{assume:bounded_deviation}) and dual variable clipping(Lemma \ref{lemma:clipping_bound}) also introduce non-descent. To resolve these, we construct a Lyapunov potential $\widetilde{\Phi}$ that absorbs all effects and restores monotone descent. Let $c_{\text{aux}}^{(i)} \triangleq \min_k c_k^{(i)}$, where $c_k^{(i)}$ is defined in~(\ref{eq:conclusion_descent_lemma}); taking the minimum across auxiliary models yields the largest error bound. The potential is then defined as:
\begin{equation}
    \widetilde{\Phi}(\paramVec^{(t + 1)}, \gA^{(t + 1)}, \Pi^{(t + 1)}, \gP^{(t + 1)}) = \Lag(\paramVec^{(t + 1)}, \gA^{(t + 1)}, \Pi^{(t + 1)}, \gP^{(t + 1)})  + \sum_{i=t}^\infty \frac{\zeta^2}{2 c_{\text{aux}}^{(i)}}
\end{equation}
The  step-to-step difference of the tail sum contributes $\frac{\zeta^2}{2 c_{\text{aux}}^{(t)}}$, canceling out the mini-batch gradient error. The DP noise and clipping noise requires no compensating term in the potential, as its cumulative contribution over the entire training horizon is finite and therefore does not threaten convergence.
Here, $\paramVec^{(t+1)}, \gA^{(t+1)}, \Pi^{(t+1)}, \gP^{(t+1)}$ denote the values after the $t$-th iteration of Algorithm~\ref{alg:dp_decoupled}. For brevity, we write $\widetilde{\Phi}_{t + 1} = \widetilde{\Phi}(\mathbf{w}^{(t + 1)})$, where $\mathbf{w}^{(t + 1)} := (\gA^{(t+1)}, \paramVec^{(t+1)}, \Pi^{(t+1)}, \gP^{(t+1)})$ collects all system parameters after the $t$-th update; $\Lag_{t + 1} = \Lag(\mathbf{w}^{(t + 1)})$ is defined analogously. 

To rigorously analyze the non-smooth optimization landscape, we first define the distance metric used for the limiting subdifferential set.
Specifically, we use the standard notation $\text{dist}(\mathbf{0}, \partial\widetilde{\Phi}(\mathbf{w})) = \inf_{\mathbf{g}\in\partial\widetilde{\Phi}(\mathbf{w})}\|\mathbf{g}\|$ for the minimal subgradient norm. Subsequently, we have $\text{dist}(\mathbf{0}, \partial\widetilde{\Phi}(\mathbf{w})) = 0$ if and only if $\mathbf{w}$ is a limit point.

Now, we provide the global expected descent lemma for Lyapunov potential function.
\begin{lemma}[Global Expected Descent with Lyapunov function]
\label{lemma:global_descent}
Let $\gF_t$ be the filtration generated by Algorithm~\ref{alg:dp_decoupled} up to iteration $t$, $\effStep^{(t)2} d \noiseStd^2$ is the expected variance bound of the DP noise at iteration $t$. Due to the geometric growth of $\rho_k^{(t)}$ and $\effStep^{(t)} = \frac{K \cdot \eta_{\vtheta}}{\sum_{k = 1}^K \rho^{(t)}_k}$, the sequence $\{\effStep^{(t)2} d \noiseStd^2\}_{t = 1}^{\infty}$ is  summable, i.e., $\sum_{t=1}^\infty \effStep^{(t)2} d \noiseStd^2 < \infty$ \footnote{This is indicated by Proposition~\ref{prop:summability_Bt_exact} in Appendix~\ref{proof:lyapunov_convergence}}. In addition, we denote $A^{(t)}_k = \frac{1}{K}\left(\frac{c^{(t)}_k}{2} - \rho_k^{(t)} - \frac{1}{2}\right)$ and $B^{(t)} = \frac{1}{\effStep^{(t)}} - \frac{L_h}{2}$ as two constants, then the sequence of the potential function $\{\widetilde{\Phi}_t\}$ satisfies the following clean expected descent inequality:
\begin{equation}
    \label{eq:clean_descent}
    \mathbb{E}[\widetilde{\Phi}_{t+1} \mid \gF_t] \le \widetilde{\Phi}_t - \sum_{k=1}^K A^{(t)}_k \|\auxParam^{(t+1)} - \paramVec^{(t)}\|^2 - B^{(t)}\E\left[ \|\paramVec^{(t+1)} - \paramVec^{(t)}\|^2 | \gF_t \right] +C_{residual}^{(t)} 
\end{equation}
where $C_{residual}^{(t)}=\E[ G^{(t)} | \gF_t] + \effStep^{(t)2} d \noiseStd^2 + \sum_{i=1}^{K}\frac{\|\kappa_k^{(t)}\|^2}{2K}$ and $\E[ G^{(t)} | \gF_t]$ is reset error in Lemma \ref{lemma:reset}. If $A^{(t)}_k, B^{(t)} > 0$ are positive constants, then $\widetilde{\Phi}$ descents faster with larger parameter updates.
\end{lemma}

Lemma~\ref{lemma:global_descent} is based on the descent lemmas in Section~\ref{subsec:descent_lemma}.
Proof details are deferred to Appendix \ref{sec:proof_global_descent}. We now establish almost-sure convergence of the Lyapunov function.

\begin{theorem}[Almost Sure Convergence of Lyapunov function]
\label{thm:robbins_siegmund}
Let $\{\widetilde{\Phi}_t\}_{t \ge 0}$ be the sequence of the Lyapunov function evaluated at the iterates generated by Algorithm \ref{alg:dp_decoupled}. Assume that the potential function is bounded from below, i.e., there exists a constant $\widetilde{\Phi}_{\min} > -\infty$ such that $\inf_{\mathbf{w}} \widetilde{\Phi}(\mathbf{w}) \ge \widetilde{\Phi}_{\min}$. Given the filtration $\gF_t$ and the expected descent inequality from Lemma \ref{lemma:global_descent} in Equation (\ref{eq:clean_descent}) with $A_k^{(t)},B^{(t)}>0$, then:
\begin{enumerate}
    \item The sequence $\{\widetilde{\Phi}_t\}_{t \ge 0}$ converges almost surely (a.s.) to a finite random variable $\widetilde{\Phi}^*$ as $t \to \infty$.
    \item The sequence of parameter updates satisfies the square-summable condition almost surely:
    \begin{equation}
        \label{eq:square_summable}
        \sum_{t=0}^\infty \|\paramVec^{(t+1)} - \paramVec^{(t)}\|^2 < \infty \quad \text{a.s., \quad and \quad} \sum_{t=0}^\infty \|\auxParam^{(t+1)} - \auxParam^{(t)}\|^2 < \infty \quad \text{a.s., } \forall k \in [K].
    \end{equation}
    Consequently, we have the asymptotic regularity: $\lim_{t \to \infty} \|\paramVec^{(t+1)} - \paramVec^{(t)}\| = 0$ a.s. and $\lim_{t \to \infty} \|\auxParam^{(t+1)} - \auxParam^{(t)}\| = 0$ a.s.
\end{enumerate}
\end{theorem}

The proof of Theorem \ref{thm:robbins_siegmund} is deferred to Appendix \ref{proof:lyapunov_convergence}. It guarantees sufficient energy descent at each iteration. However, to understand the local geometry of the optimization landscape, we must link the magnitude of the parameter updates to the steepness of the potential function. We establish a relative error condition demonstrating that a small parameter update implies the system is approaching a limit point.

\begin{lemma}[Relative Error Condition and Subgradient Bound]
\label{lemma:subgradient_bound}
 There exists two constant $b > 0$, $ \widetilde{C}\ge 0$ and a non-negative stochastic error sequence $\{\xi^{(t)}\}_{t \ge 0}$, such that the subgradient norm of the Lyapunov function $\widetilde{\Phi}$ evaluated at the updated point $\mathbf{w}^{(t+1)}$ is bounded by the distance of the parameter updates:
\begin{equation}
    \label{eq:subgradient_bound}
    \text{dist}\left(\mathbf{0}, \partial \widetilde{\Phi}(\mathbf{w}^{(t+1)})\right) \le b \|\mathbf{w}^{(t+1)} - \mathbf{w}^{(t)}\| + \xi^{(t)} +  \widetilde{C}
\end{equation}
where $\widetilde{C}=C+ dist(0,\partial r(\paramVec^{(t+1)}))$ is finite, and the residual error sequence $\{\xi^{(t)}\}$ is almost surely summable over the infinite horizon(i.e. $\sum_{t=0}^\infty \xi^{(t)} < \infty$ a.s.). The specific formula of $\xi$ and $b$ are deferred to Appendix \ref{sec:subgradient}.
\end{lemma}

This lemma links the algorithmic dynamics to first-order optimality. It implies that the subgradient norm vanishes as the iterates stabilize. 

\subsection{Global Convergence via the Kurdyka-\L{}ojasiewicz Property}
\label{sec:stationary}

Although Lemma \ref{lemma:subgradient_bound} establishes a vital connection between the parameter displacement and the subgradient norm, it does not preclude the possibility of the sequence infinitely oscillating in a high-dimensional space. To mathematically eliminate this behavior, we leverage the Kurdyka-\L{}ojasiewicz (KL) property.

\begin{definition}[Kurdyka-\L{}ojasiewicz Property]
\label{def:kl_property}
A proper lower semicontinuous function $\widetilde{\Phi}$ satisfies the KL property at a limit point $\mathbf{w}^*$ if there exists a concave desingularizing function $\varphi \in C^1(0, r_{kl})$ with $\varphi(0)=0$, $\varphi'>0$, such that locally:
\begin{equation}
    \label{eq:kl_inequality}
    \varphi'(\widetilde{\Phi}(\mathbf{w}) - \widetilde{\Phi}(\mathbf{w}^*)) \cdot \text{dist}(\mathbf{0}, \partial\widetilde{\Phi}(\mathbf{w})) \ge 1
\end{equation}
\end{definition}
In our context, since the augmented Lagrangian $\Lag$ and the potential function $\widetilde{\Phi}$ are constructed via deep neural network components with typical activation functions (e.g., ReLU) and standard loss functions, they belong to the class of semi-algebraic functions. Any semi-algebraic function naturally satisfies the KL property with the desingularizing function taking the form $\varphi(s) = \frac{\bar{c}}{1-c_{kl}} s^{1-c_{kl}}$, where $\bar{c} > 0$ is a constant and $c_{kl} \in [0, 1)$ is the KL exponent.

By exploiting the geometric concavity of the KL desingularizing function, we elegantly translate the potential energy descent into a bounded path length.

\begin{lemma}[Finite Length Property of the Sequence]
\label{lemma:finite_length}
Suppose the subgradient is bounded as $\text{dist}\left(\mathbf{0}, \partial \widetilde{\Phi}_t(\mathbf{w}^{(t)})\right) \le C_2 \|\mathbf{w}^{(t)} - \mathbf{w}^{(t-1)}\| + \xi^{(t)}+\widetilde{C}$, with the constants satisfying $2C_1 > C_2$. Under the KL-based potential difference bound established in Lemma \ref{lemma:kl_difference}, the sequence $\{\mathbf{w}^{(t)}\}_{t \ge 0}$ has a finite length, i.e.,
\begin{equation}
    \sum_{t=1}^{\infty} \|\mathbf{w}^{(t+1)} - \mathbf{w}^{(t)}\| < \infty \quad \text{a.s.}
\end{equation}
where $C_1=\min\{\min_k \frac{A_k}{1+2\rho_k^2},B\}$ from Lemma \ref{lemma:global_descent} and $C_2=b$ from Lemma \ref{lemma:subgradient_bound}.
\end{lemma}

The proof of Lemma \ref{lemma:finite_length} is deferred to Appendix \ref{proof:finite_length}. The condition $2C_1 > C_2$ required therein is the most restrictive among all convergence requirements. Together with $A_k > 0$ and $B > 0$ from Lemma~\ref{lemma:global_descent}, these conditions translate directly into concrete learning-rate constraints, which we formalize in  Appendix~\ref{sec:global_convergence}. The finite length property established in Lemma~\ref{lemma:finite_length} fundamentally constrains the topological behavior of the sequence, ensuring it forms a Cauchy sequence. Armed with this property and the summable subgradient bounds, we now present the main theorem of our analysis.

\begin{theorem}[Global Convergence to a Unique Limit Point]
\label{thm:global_convergence}
Let $\{\mathbf{w}^{(t)}\}_{t \ge 0}$ be the sequence generated by Algorithm~\ref{alg:dp_decoupled}. Under the learning-rate conditions $\eta_k < 2\minCurv/(L + 5\rho_k + 2)$ and $\eta_\theta < 1$ for all $k \in [K]$, the sequence converges almost surely to a unique limit point $\mathbf{w}^*$, and
$\operatorname{dist}(\mathbf{0}, \partial\widetilde{\Phi}(\mathbf{w}^*)) \le \widetilde{C} \text{a.s.}$
\end{theorem}

Lemma~\ref{lemma:finite_length} implies $\{\mathbf{w}^{(t)}\}$ is Cauchy, hence converges to a unique $\mathbf{w}^*$ a.s.; Lemma~\ref{lemma:subgradient_bound} then yields $\operatorname{dist}(\mathbf{0}, \partial\widetilde{\Phi}(\mathbf{w}^*)) \le \widetilde{C} \quad \text{a.s},$. Full proof in Appendix~\ref{sec:global_convergence}.



\section{Privacy Loss Analysis}

\subsection{Preliminary}
\label{subsec:privacy_prelim}
As described in Section~\ref{subsec:pre}, our threat model is based on the hidden state assumption, where only the final state of the publishable parameters $\vtheta$, is accessible to adversaries. In the DP-DT framework introduced in Section~\ref{subsec:dpdt}, we employ the update rule in (\ref{equ:theta_update}) to enhance the differential privacy of $\vtheta$.
This update rule can be considered as a diffusion process~\citep{balle2019privacy, chourasia2021differential, ye2022differentially} consisting of three key steps: running clipped stochastic gradient descent, adding random calibrated noise, and applying proximal operators.

Differential privacy quantifies the distributional distance between outputs when given two adjacent input datasets. 
To analyze its impact, we study how the update rule in~(\ref{equ:theta_update}) alters the distribution of the publishable parameters $\vtheta$. 
Specifically, let $\Theta$ denote the distribution of $\vtheta$. After applying the one-step update in (\ref{equ:theta_update}), the updated distribution $\Theta'$ is:
\begin{equation} \label{equ:gaussian_process}
\begin{aligned}
&\Theta' = T_\# (F_\# (\Theta) * \gN(0, 2\widehat\eta \noiseLStd^2 \rmI)) 
\end{aligned}
\end{equation}
We define $\noiseStd^2 = 2\widehat \eta \noiseLStd^2$ to align the noise scale with the canonical form of Langevin dynamics implied by~(\ref{equ:theta_update}). In addition, $F_\#$ and $T_\#$ denote push-forward mappings, and $*$ represents the convolution operator between two distributions. These three operators directly correspond to the three steps above: $F_\#$ applies gradient descent to each sample, $*$ adds Gaussian noise, and $T_\#$ applies the proximal mapping.  Specifically, $F$ corresponds to the gradient descent update: $F(\vtheta) = \vtheta-\widehat \eta  \nabla h(\vtheta)$, while $T$ represents the convex bijective proximal operator mapping: $T(\vtheta) = \text{Prox}_{\widehat{\eta}, r}(\vtheta)$.

In DP-DT, we iteratively apply (\ref{equ:theta_update}) to update the publishable parameters $\vtheta$. Therefore, in privacy accounting, we investigate how the distributional distance between parameter distributions evolves when given two adjacent input datasets and applying (\ref{equ:gaussian_process}) iteratively. To facilitate analysis, we adopt R\'enyi differential privacy (RDP), which can be translated to standard differential privacy~\citep{mironov2017renyi}.

\subsection{Privacy Accounting}
\label{subsec:privacy_accounting}

In order to track the trajectory of $\Theta$,  we first establish the Lipschitz continuity properties of the operators $F$ and $T$.

\begin{lemma}[Lipschitz continuity for $F$] \label{lem:F_Lip} The gradient descent update function $F(\vtheta) = \vtheta - \widehat{\eta} \nabla h(\vtheta)$ is Lipschitz continuous with a constant $L_F = 1 - \eta_{\vtheta}$, where $h(\vtheta)$ is defined in (\ref{equ:theta_update}) and $\widehat{\eta} = \frac{K \cdot \eta_\vtheta}{\sum_{k = 1}^K \rho_k}$ is the effective step size.
\end{lemma}

\begin{lemma}[Lipschitz continuity for $T$] \label{lem:T_Lip}
    If $\widehat{\eta} > 0$ and $r(\vtheta)$ is a convex function, then the proximal operator function $T(\vtheta) = \text{Prox}_{\widehat\eta, r}(\vtheta) = \argmin_{\widetilde{\vtheta}} \left\{ r_\vtheta(\widetilde{\vtheta}) + \frac{1}{2\widehat\eta} \|\widetilde{\vtheta} - \vtheta\|^2_2 \right\}$ is Lipschitz continuous with a constant $L_T \leq 1$.
\end{lemma}

The proof of Lemma \ref{lem:F_Lip} and \ref{lem:T_Lip} is deferred to Appendix~\ref{proof:F_Lip} and \ref{proof:T_Lip} .
This work primarily employs $\ell_2$ regularization (weight decay): $r(\vtheta) = \frac{\reg}{2}\|\vtheta\|_2^2$. Under this setting, the proximal operator admits the following closed-form solution
$\text{Prox}_{\widehat{\eta}, r}(\vtheta) = \frac{1}{1 + \reg \widehat\eta} \vtheta$ and is $1$-Lipchitz. 
Nevertheless, it is worth noting that Lemma~\ref{lem:T_Lip} is general and supports other regularization schemes like $l_1$ regularization.
In addition, when there is no regularization, we have $r_\vtheta = 0$ and $T(\vtheta) = \vtheta$, which is also 1-Lipschitz.
Now we study the properties of the distribution of the publishable model $\vtheta$.
We assume that the initial distribution of $\vtheta$ satisfies Log-Sobolev Inequality~\citep{vempala2019rapid} defined as below:
\begin{definition}(Log-Sobolev Inequality) \label{def:lsi}
    A distribution $\nu$ over $\mathbb{R}^d$ satisfies the log-Sobolev inequality (LSI) with a constant $c$ if for any smooth function $g: \mathbb{R}^d \to \mathbb{R}$ with $\mathbb{E}_{\vtheta \sim \nu} [g(\vtheta)^2] < \infty$, the following holds:
    \begin{equation}
    \begin{aligned}
        \E_{\vtheta \sim \nu} [|g(\vtheta)|^2 \log (|g(\vtheta)|^2)] - \frac{2}{c} \E_{\vtheta \sim \nu} [|\nabla_\vtheta g(\vtheta)|^2]\leq \E_{\vtheta \sim \nu} [|g(\vtheta)|^2] \log \E_{\vtheta \sim \nu} [|g(\vtheta)|^2]
    \end{aligned}
    \end{equation}
    
\end{definition}

LSI is a benign assumption commonly used in existing literature~\citep{vempala2019rapid, ye2022differentially} for the distribution of parameter.
The LSI assumption is very mild, strongly log-concave distribution, such as Gaussian distribution, uniform distribution, and some non-logconcave distribution satisfy the LSI assumption~\citep{vempala2019rapid}.
Popular neural network initialization schemes~\citep{glorot2010understanding, he2015delving} are based on Gaussian or uniform distributions, so the initial distribution of $\vtheta$ satisfies LSI assumption.

Based on the LSI assumption and Lipschitz constants derived in Lemma \ref{lem:F_Lip}, \ref{lem:T_Lip}, we consider the recursive privacy dynamics for the noisy proximal gradient descent in (\ref{equ:theta_update}) and bound the change rate of RDP during one step.
Our formal result is demonstrated as Theorem~\ref{thm:rate_of_rdp} in Appendix~\ref{proof:rate_of_rdp}.
It is an extension of~\citet[Lemma 3.1]{ye2022differentially} to the noisy proximal gradient descent that we adopt in our update rule in (\ref{equ:theta_update}).





As discussed in Section~\ref{subsec:dpdt}, our proposed DP-DT in Algorithm~\ref{alg:dp_decoupled}  resets auxiliary parameters in the beginning of each iteration and apply clipping on the updates of both auxiliary parameters and their multipliers to control the magnitude of the gradients for the global parameters $\vtheta$.
These mechanisms effectively bound the data sensitivity on $\vtheta$ by a pre-defined constant.
Formally, the update rule on $\vtheta_k$ and $\vpi_k$ in DP-DT guarantees $\forall k$, $\|\nabla h(\vtheta, \vtheta_k)\|_2 \leq C$.
Considering $h(\vtheta) = \frac{1}{K} \sum_{k = 1}^K h(\vtheta, \vtheta_k)$, we conclude that adding or removing any single training data will have a bounded effect on the gradient $\nabla_\vtheta h(\vtheta)$ in the update rule (\ref{equ:theta_update}): $\|\nabla_\vtheta h(\vtheta)\| \leq C/K$.
That is to say, the data sensitivity bound for $\vtheta$ is $C/K$.

With bounded data sensitivity and the change rate of RDP derived in Theorem~\ref{thm:rate_of_rdp}, we can directly adopt \citet[Lemma 3.2]{ye2022differentially} to obtain the following Corollary to conclude how privacy loss $\epsilon$ in RDP changes for one iteration of DP-DT in Algorithm~\ref{alg:dp_decoupled}.

\begin{corollary} \label{coro:recursive} Let $\gD$, $\gD'$ be an arbitrary pair of the neighboring datasets that only differ in the $i_0$-th data point (i.e. $x_{i_0} \ne x_{i_0}'$). We denote $\vtheta$ and ${\vtheta'}$ as the intermediate global weight parameters. If the distributions of $\vtheta$ and $\vtheta'$ satisfy log-Sobolev inequality with a constant $c$, the update function $F(\vtheta) = \vtheta - \widehat \eta \nabla h(\vtheta)$ and the proximal operator $T(\vtheta) = \text{Prox}_{\widehat \eta, r}(\vtheta)$ are Lipschitz continuous with Lipschitz constant $L_F$ and $L_T$, respectively, then the following recursive bound for R\'enyi divergence holds for any order $\renyiOrder > 1$:

1) If $i_0 \in \gD_m$, then $\frac{R_\renyiOrder(\vtheta||{\vtheta'})}{\renyiOrder} \leq \frac{R_{\renyiOrder}(\vtheta || {\vtheta'})}{\renyiOrder} + \frac{\widehat\eta C^2}{2K^2 \noiseLStd^2}$.

2) Otherwise, $\frac{R_\renyiOrder(\vtheta||{\vtheta'})}{\renyiOrder} \leq \frac{R_{\renyiOrder'}(\vtheta||{\vtheta'})}{\renyiOrder'} \cdot \left( 1+ \frac{c \cdot 2 \widehat \eta \cdot \noiseLStd^2}{L_F^2} \right)^{-1 / L_T^2}$ where $\renyiOrder' = (\renyiOrder-1)\left(1+\frac{c \cdot 2\widehat\eta \noiseLStd^2}{L_F^2}\right)^{-1} + 1$.

\end{corollary}


By applying Corollary~\ref{coro:recursive} iteratively, we can obtain the algorithm's privacy loss for the whole training phase. The formal theorem is demonstrated below.

\begin{theorem} \label{theorem:privacy_loss}
    Suppose the initial distribution of the global publishable parameters $\vtheta$ follows the log-Sobolev inequality (LSI). 
    Let the update function $F$ and proximal operator $T$ be Lipschitz continuous with constants $L_F$ and $L_T$, respectively. 
    After training for $E \geq 1$ epochs using Algorithm~\ref{alg:dp_decoupled} with $M$ iterations per epoch and $K$ auxiliary models, the mechanism satisfies $(\renyiOrder, \varepsilon_E(\renyiOrder))$-R\'enyi differential privacy, where:
    \begin{equation}
        \varepsilon_E(\renyiOrder) \le \frac{1}{\renyiOrder-1}\log\left( \sum_{m_0=1}^{M} \frac{1}{M} e^{(\renyiOrder-1) \sum_{l=1}^{E}\varepsilon(\alpha, m_0, l)} \right).
    \end{equation}
    Here, the epoch-wise privacy cost $\varepsilon(\renyiOrder, m_0,l)$ is defined as:
    \begin{equation} \label{eq:eps_K}
       \varepsilon(\alpha, m_0, l) = \frac{\renyiOrder \widehat \eta C^2}{2 K^2 \noiseLStd^2} \cdot \left[ \Lambda(l, m_0) \right]^{-1/{L_T^2}}
    \end{equation}
    where $\Lambda(l, m_0)$ encapsulates the contraction factor derived from LSI constants:
    \begin{equation} \label{eq:lambda}
        \Lambda(l, m_0) = \frac{c^{m_0}_l}{c^{M-1}_{E-1}} \left( L_F^2 L_T^2 \right)^{M(E - l) - m_0} \prod_{p = l + 1}^{E} \frac{c^{m_0}_p}{c^{m_0 - 1}_p}
    \end{equation}
    The LSI constants $c^m_l$ represents the LSI constant for the distribution of $\vtheta$ in the $l$-th epoch and the $m$-th mini-batch. $c^m_l$ can be efficiently computed by Lemma~\ref{lemma:LSI sequence} in the Appendix.
\end{theorem}

\textit{Proof Sketch.} 
Although the update rule (\ref{equ:theta_update}) is only applied to $\vtheta$, it depends on other variables in $\gA$, $\Pi$ and $\sP$ as well.
Nevertheless, the data sensitivity $C / K$, the Lipschitz constant $L_F$ and $L_T$ is always satisfied regardless of the values of these variables.
Therefore, we can recursively use Corollary~\ref{coro:recursive} to analyze the algorithm's privacy loss.
In each epoch, there is one and only one different mini-batch for two neighboring datasets, so we assume it is the $m_0$-th batch without the loss of generality and apply case 1) of Corollary~\ref{coro:recursive} for this mini-batch update and case 2) of Corollary~\ref{coro:recursive} for the updates of other mini-batches.
Therefore, we can uniformly bound the value of $\frac{\varepsilon(\renyiOrder, m_0,l)}{\renyiOrder}$ for any $\renyiOrder$.
Finally, since $m_0$ is uniformly distributed among the index set $\{1, 2, \ldots, M\}$, we use the joint convexity of scaled exponentiation of R\'enyi divergence to bound the final privacy loss $\varepsilon_E(\renyiOrder)$. The complete proof is deferred to the Appendix\ref{proof:privacy_loss}.

$\varepsilon(\renyiOrder, m_0,l)$ in the Theorem~\ref{theorem:privacy_loss} represents the privacy loss when the only different instance of the two neighboring datasets is in the $m_0$-th mini-batch of each epoch.
The overall privacy loss is the summation of each epoch's privacy loss term with a discount term $\Lambda(l, m_0)$. 
Moreover, the discount term $\Lambda(l, m_0)$, as defined in (\ref{eq:lambda}), increases with $l$.
This indicates that the earlier epochs have less contribution to the overall privacy loss under the hidden state assumption.


\subsection{Convergence of Privacy Loss and Noise Calibration}

\label{sec:result_analysis}

\begin{figure}[htbp] 
    \centering
    \begin{subfigure}[b]{0.48\columnwidth}
        \centering
        \includegraphics[width=\linewidth]{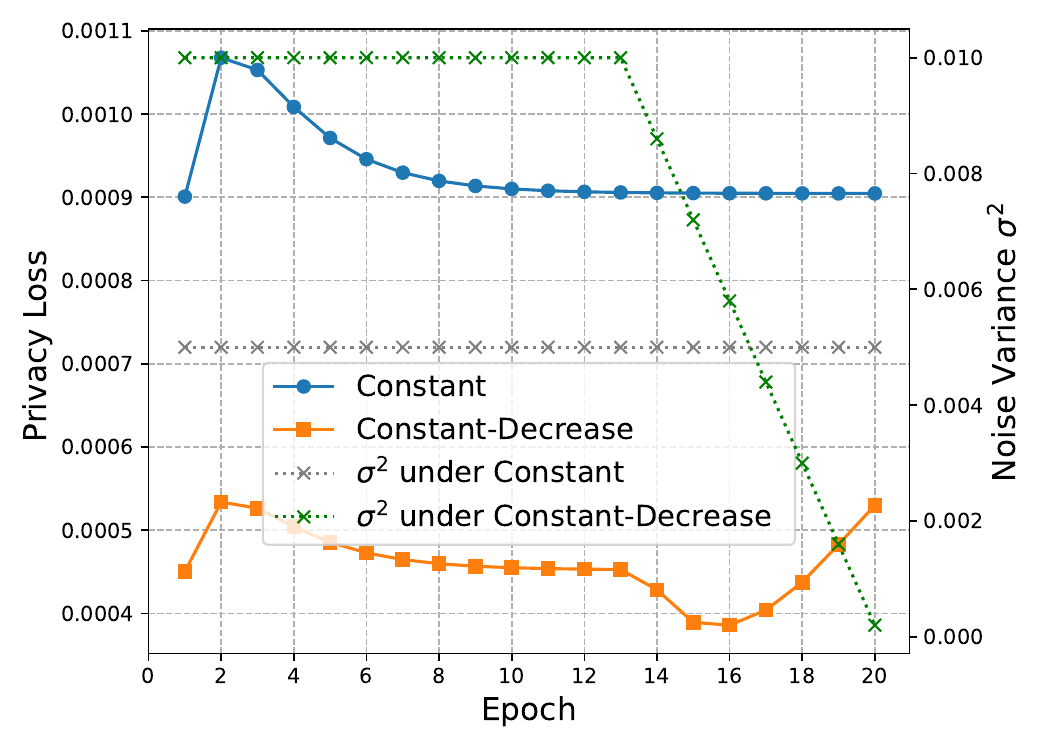}
        \caption{$\varepsilon_E(\alpha)$ for different $E$.}
        \label{fig:adaptive_noise}
    \end{subfigure}
    \hfill 
    \begin{subfigure}[b]{0.48\columnwidth}
        \centering
        \includegraphics[width=\linewidth]{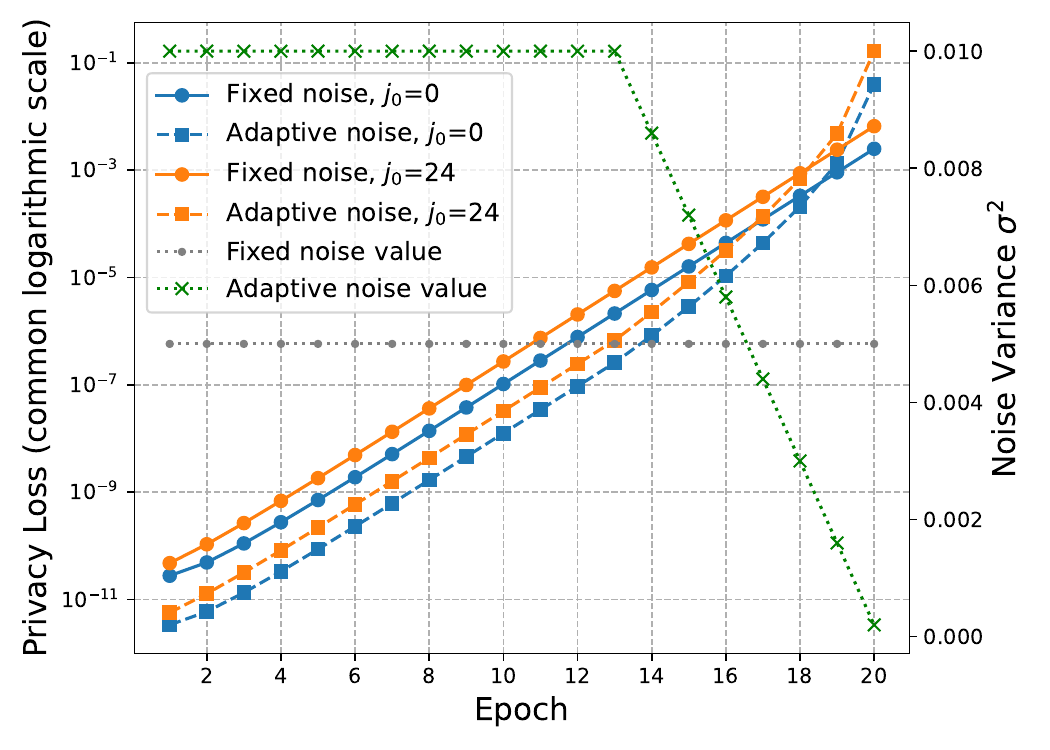}
        \caption{$\varepsilon(\alpha, m_0, l)$ for different $l$ and fixed $E$.}
        \label{fig:per_epoch}
    \end{subfigure}
    
    \caption{\small (a) We show the overall privacy loss $\varepsilon_E(\alpha)$ for different epoch numbers $E$. The horizontal axis represents $E$. (b) We show the privacy loss contribution $\varepsilon(\alpha, m_0, l)$ for the $l$-th epoch with a fixed epoch number $E = 20$ over different values of $m_0$. The horizontal axis represent $l$ and the left vertical axis is log-scale. The curves showing the variance of calibrated noise follow the right vertical axis. Hyper-parameters are as follow: Sensitivity $S_g=1$, batch size $b=100$ and sample size $n=2500$, Lipschitzness $L_F=0.98,L_T=1$, learning rate $\widehat \eta=0.05$, RDP order $\alpha=10$.}
    \label{fig:maintext}

\end{figure}

Theorem~\ref{theorem:privacy_loss} shows that the total privacy loss is a sum of per-epoch terms, each containing a leakage increment and a decay factor, both governed by the noise variance $\sigma^2$. The interdependency between these two components makes a closed-form characterization of convergence difficult, we therefore turn to numerical analysis.

Figure~\ref{fig:adaptive_noise} plots the total privacy loss $\varepsilon_E(\alpha)$ against the epoch number $E$ under constant and adaptive noise schedules. Figure~\ref{fig:per_epoch} fixes $E$ and shows the per-epoch contribution $\varepsilon(\alpha, m_0, l)$ as a function of $l$, averaged over $m_0$.
Figure~\ref{fig:per_epoch} shows that the per-epoch contribution decays linearly on the log-scale as the distance from the terminal epoch increases, confirming a constant exponential decay rate. Under constant noise, this fixed rate allows the leakage and decay to reach an equilibrium, and $\varepsilon_E$ converges to a finite limit---consistent with the prediction of~\citet[Theorem~23]{feldman2018privacy} that HSA-based privacy loss stabilizes under contractive dynamics. Adaptive noise disrupts this equilibrium: the decay rate varies with $\sigma$, so $\varepsilon_E$ never
stabilizes, as seen in Figure~\ref{fig:adaptive_noise}.

The same noise variance $\sigma^2$ couples the two analyses. In Corollary~\ref{coro:recursive}, $\sigma$ appears in the denominator of the per-epoch leakage: larger $\sigma$ tightens the privacy guarantee. In Lemma~\ref{lemma:global_descent}, the noise penalty term scales as $B_t \propto \sigma^2$: larger $\sigma$ slows the energy descent. The two
effects pull in opposite directions.

In summary, fixed hyperparameters yield converged privacy loss; larger $\sigma$ tightens the final bound but slows convergence. We therefore adopt a two-phase strategy. In early training, $\sigma$ is kept small to drive $\widetilde{\Phi}_t$ down rapidly; the resulting larger per-epoch leakage is heavily discounted by the remaining contraction steps. In later epochs, $\sigma$ is gradually increased to a target value and then held constant, so that $\kappa$ becomes fixed, the equilibrium re-establishes, and $\varepsilon_E$ converges to a new, tighter limit.



\section{Experiments}




\subsection{Experiment Setup}
\label{sec:exp-setup}
We conduct two sets of experiments.
First, we compare DP-DT against DP-SGD and non-private training across four
benchmarks.
Second, we ablate the two structural parameters that govern DP-DT's privacy--utility trade-off under the Hidden State Assumption: per-layer budget allocation and auxiliary ensemble configuration.

\paragraph{Datasets and models.}
We evaluate on \textbf{CIFAR-10} and \textbf{CIFAR-100} (ResNet-18),
\textbf{ImageNet-100} (ViT-Small), and \textbf{SST-2} (GPT-2~117M with LoRA,
$r=16$, $\alpha=32$).
Standard data augmentation (random crop, horizontal flip) is applied to all
vision benchmarks.

\paragraph{DP-DT configuration.}
Unless noted otherwise, all DP-DT experiments use $K = 40$ auxiliary models
($K = 16$ for SST-2), $\delta = 10^{-5}$, and per-layer RDP composition
under HSA.
BatchNorm is replaced by LayerNorm across all architectures.
Auxiliary models are optimized with RMSProp and synchronized to a global
model updated via SGD (lr~$=0.99$, wd~$=5\times10^{-5}$).
Per-dataset batch sizes, epoch counts, clipping thresholds, noise multipliers,
and the penalty decay and clipping schedules are detailed in
Appendix~\ref{appendix:hyperparameters}.

\paragraph{Baselines.}
We compare against non-private training (SGD for CNNs, AdamW for Transformers)
and DP-SGD (FastDP~\citep{bu2024automatic} implementation with automatic
clipping).
On CIFAR-10 we additionally reference PATE~\citep{papernot2016semi} and
GEP~\citep{yu2021not} as CNN-specific DP baselines, using the results
reported in their original papers (which use ResNet-20, an architecture
with marginally fewer parameters than ResNet-18).
All methods share the same architecture and $\delta = 10^{-5}$ per benchmark..

\subsection{Main Results}
\label{sec:main-results}
 




\begin{table}[htbp]
\centering
\small
\caption{%
  Privacy-utility trade-off across datasets and architectures.  PATE and GEP are only applicable to CNN-based architectures.
}
\label{tab:main_results}
\setlength{\tabcolsep}{6pt}
\begin{tabular}{l c c c c c c c c}
\toprule
& \multicolumn{2}{c}{\textbf{CIFAR-10}} & \multicolumn{2}{c}{\textbf{CIFAR-100}}
& \multicolumn{2}{c}{\textbf{ImageNet-100}} & \multicolumn{2}{c}{\textbf{SST-2}} \\
& \multicolumn{2}{c}{\small\itshape ResNet18} & \multicolumn{2}{c}{\small\itshape ResNet18}
& \multicolumn{2}{c}{\small\itshape ViT-Small} & \multicolumn{2}{c}{\small\itshape GPT-2 (Medium)} \\
\cmidrule(lr){2-3} \cmidrule(lr){4-5} \cmidrule(lr){6-7} \cmidrule(lr){8-9}
\textbf{Method} & $\varepsilon$ & Acc(\%) & $\varepsilon$ & Acc(\%) & $\varepsilon$ & Acc(\%) & $\varepsilon$ & Acc(\%) \\
\midrule
Non-private & $\infty$ & 92.85 & $\infty$ & 65.44 & $\infty$ & 60.38 & $\infty$ & 93.92 \\[0.3em]
DP-SGD      & 1.00     & 43.64 & 2.00     & 14.57 & 4.05    & 16.28 & 2.04     & 50.51 \\[0.3em]
PATE & 2.00 & 46.20 &2.00 &2.88 & - & - & - & - \\[0.3em]
GEP  & 2.00 & 56.60 &2.00 &18.13 & - & - & - & - \\[0.3em]
\textbf{DP-DT} & \textbf{0.64} & \textbf{82.59} & \textbf{1.51} & \textbf{51.48} & \textbf{4.05} & \textbf{53.62} & \textbf{2.04} & \textbf{92.54} \\
\bottomrule
\end{tabular}
\end{table}

Table~\ref{tab:main_results} reports the privacy-utility trade-off across four benchmarks.
DP-DT consistently outperforms baselines (DP-SGD, PATE, GEP) across all four benchmarks, narrowing the gap to non-private performance in model utility and therefore achieving Pareto improvements in privacy-utility trade-off.

The decoupled design in DP-DT, namely noise-free auxiliary training with privacy noise confined to the
global aggregation step under the Hidden State Assumption, underpins this broad
advantage.

Empirical privacy evaluation via membership inference attacks
(Appendix~\ref{sec:mia}) further confirms that DP-DT's practical privacy leakage
is consistent with its theoretical guarantee.

\subsection{Ablation Studies}

Under HSA, the privacy-utility trade-off shifts from training iterations to per-layer noise allocation and DP-DT hyperparameters; we examine both factors in this Section.

\subsubsection{Noise Allocation Strategies}
\label{sec:noise-allocation}
\paragraph{Design.}
We start from an \texttt{equal\_baseline} where the three layer types
(conv, ln, fc) contribute equally to the total privacy loss ($1:1:1$
ratio under per-layer composition).
We then create three \emph{boost} variants that increase the target
layer's share by a factor $r_g$, shifting the contribution ratio to
$r_g : 1 : 1$.
The noise multiplier $\sigma_g$ and clipping threshold $C_g$ are
derived from this allocation, with $C_g$ recalibrated to keep the
overall $(\varepsilon,\delta)$ guarantee comparable across all
configurations; accuracy differences therefore isolate the effect of
budget \emph{allocation} rather than a change in total privacy cost.

We sweep $r_g \in [0.5, 2.0]$ and evaluate two strategies:
adjusting noise alone (\emph{Noise}) and jointly adjusting noise
and clipping threshold (\emph{Joint}).
Table~\ref{tab:noise-allocation} reports the $r_g = 1.25$ case;
Figure~\ref{fig:ratio_vs_acc} shows the full sweep.

Because this ablation uses an equalized baseline to isolate the
allocation effect, the absolute $\varepsilon$ values differ from
the main result (which already applies a conv-favoring allocation).

\begin{table}[htbp]
\centering
\caption{%
  Test accuracy and MIA accuracy under different per-layer privacy-budget
  allocation strategies on CIFAR-10 and CIFAR-100.
  \emph{Noise-only}: only the per-layer noise multiplier $\sigma_g$ is scaled;
  \emph{Joint}: both $\sigma_g$ and the per-layer clipping threshold $C_g$ are
  jointly scaled.
  Each configuration fixes the RDP budget split (Conv:Norm:FC) as follows:
  \emph{Uniform} uses an equal split ($33\%{:}33\%{:}34\%$);
  \emph{Conv-weighted} allocates more budget to convolutional layers
  ($43\%{:}29\%{:}28\%$);
  \emph{Norm-weighted} favours normalisation layers ($29\%{:}43\%{:}28\%$);
  \emph{FC-weighted} favours the fully-connected layer ($29\%{:}29\%{:}42\%$).
  Bold denotes the best utility accuracy within each group.
}
\label{tab:noise-allocation}
\begin{adjustbox}{width=\linewidth}
\begin{tabular}{l l c c c c c c}
\toprule
& & \multicolumn{3}{c}{\textbf{CIFAR-10}} & \multicolumn{3}{c}{\textbf{CIFAR-100}} \\
\cmidrule(lr){3-5} \cmidrule(lr){6-8}
Strategy & Configuration & $\varepsilon$ & Acc (\%) & MIA (\%) & $\varepsilon$ & Acc (\%) & MIA (\%) \\
\midrule
\multirow{4}{*}{Noise-only}
 & Uniform        & 0.78 & 77.94          & $56.44 \pm 0.17$ & 1.04 & 41.73          & $55.32 \pm 0.30$ \\
 & Conv-weighted  & 0.84 & \textbf{80.84} & $56.50 \pm 0.40$ & 1.11 & \textbf{45.55} & $55.47 \pm 0.09$ \\
 & Norm-weighted  & 0.84 & 78.97          & $56.22 \pm 1.65$ & 1.11 & 43.13          & $55.41 \pm 0.15$ \\
 & FC-weighted    & 0.85 & 78.43          & $56.31 \pm 1.33$ & 1.12 & 41.45          & $55.27 \pm 0.23$ \\
\midrule
\multirow{4}{*}{Joint}
 & Uniform        & 0.79 & 78.51          & $56.45 \pm 1.52$ & 1.04 & 44.31          & $55.68 \pm 0.28$ \\
 & Conv-weighted  & 0.84 & \textbf{80.99} & $56.43 \pm 0.83$ & 1.12 & \textbf{47.34} & $55.74 \pm 0.27$ \\
 & Norm-weighted  & 0.84 & 78.25          & $56.37 \pm 0.42$ & 1.12 & 44.95          & $55.76 \pm 0.21$ \\
 & FC-weighted    & 0.85 & 78.00          & $56.57 \pm 0.28$ & 1.13 & 46.29          & $55.74 \pm 0.14$ \\
\bottomrule
\end{tabular}
\end{adjustbox}
\end{table}

\paragraph{Findings.}
Figure~\ref{fig:ratio_vs_acc} and Table~\ref{tab:noise-allocation}
show a clear sensitivity ranking across layer types in ResNet-18:
convolutional layers are the most noise-sensitive, followed by
normalization layers, with the FC head largely insensitive.
Concentrating the privacy budget on conv layers
($r_{\text{conv}}=1.25$) yields the highest accuracy on both datasets
under both strategies.
MIA accuracy remains stable across all 18 configurations
($56.4\% \pm 1.5\%$ on CIFAR-10, $55.5\% \pm 0.5\%$ on CIFAR-100),
confirming that the DP guarantee is unaffected.
A well-designed per-layer allocation thus improves utility  without increasing privacy cost.

\subsubsection{Utility With Respect To Training Hyper-parameters}
\label{sec:hp-ablation}

We ablate the number of auxiliary models $K$ and batch size $B$ on CIFAR-10 and
CIFAR-100, keeping privacy settings fixed and reducing training to 200 epochs for
efficiency (full sweep ranges in Appendix~\ref{sec:ablation-setup}).
Figure~\ref{fig:ablation} shows the trade-off with sliding-window smoothing applied
to privacy loss.
The privacy loss rises steeply in the initial phase due to the adaptive scheduler's
large clipping thresholds in early epochs (see Appendix~\ref{appendix:hyperparameters});
overall, the trajectory moves toward the top-right, indicating improved utility
under privacy constraints.
For a fixed batch size, fewer auxiliary models generally yield higher privacy loss
but superior utility.
The trajectory exhibits distinct dynamics: in the early phase (large clipping
threshold), more models accelerate convergence, whereas in the later phase (small
clipping threshold), fewer models enhance utility.
These results indicate that unlike standard DP-SGD, DP-DT's privacy loss (under the
Hidden State Assumption) is intrinsically coupled with the algorithm's convergence
state rather than strictly accumulating with steps.

\begin{figure}[H]
    \centering
    \begin{subfigure}[b]{0.47\textwidth}
        \centering
        \includegraphics[width=\linewidth]{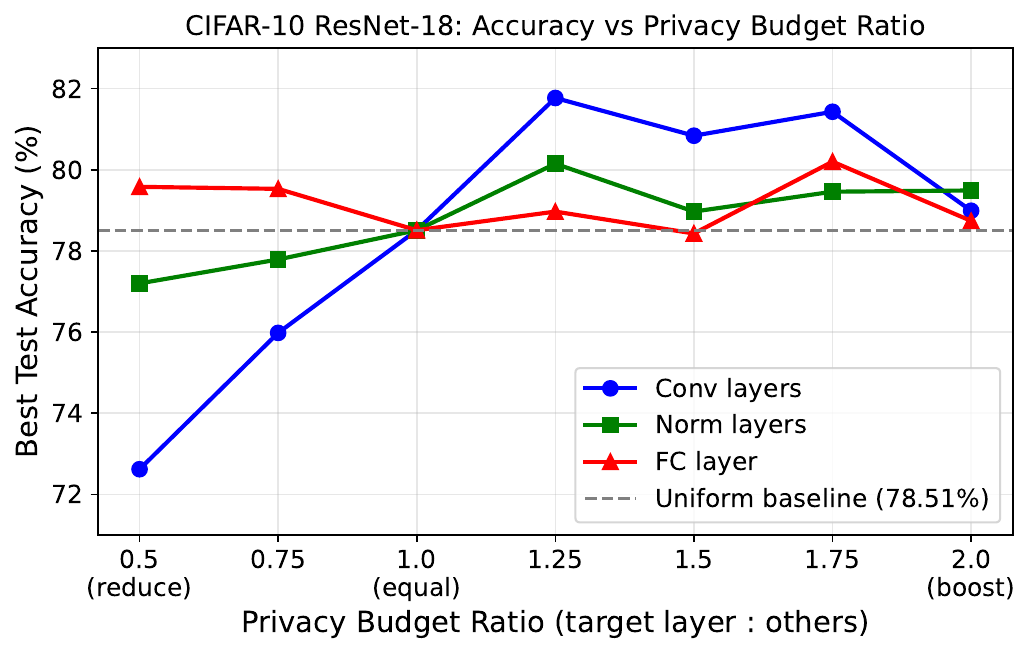}
        \phantomsubcaption
        \label{fig:ratio_vs_acc}  
    \end{subfigure}
    \hfill
    \begin{subfigure}[b]{0.48\textwidth}
        \centering
        \includegraphics[width=\linewidth]
        {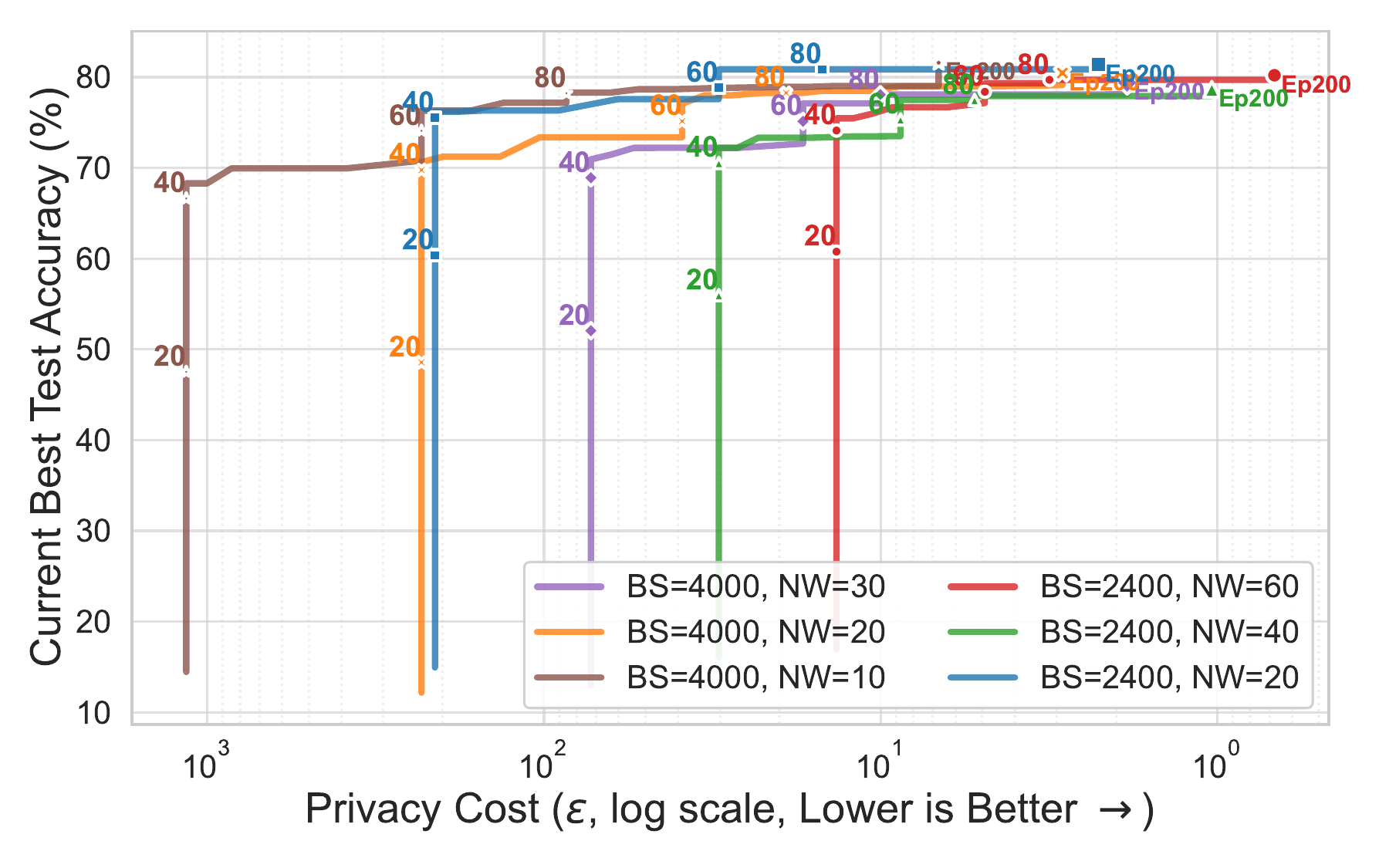}
        \phantomsubcaption
        \label{fig:ablation}
    \end{subfigure}

    \caption{
        \textbf{(a)} Test accuracy on CIFAR-10 as a function of the per-layer budget ratio $r_g$ ($r_g > 1$ = more budget / lower noise). Dashed line: equal-baseline. Conv layers peak near $r_{\text{conv}}=1.25$ and drop sharply below it; FC head is nearly flat.
        \textbf{(b)} Privacy-utility trade-off (smoothed). Markers: sequential epochs. Privacy axis inverted; top-right = optimal.
    }
    \label{fig:combined}  
\end{figure}

\section{Discussion and Future Work}
\label{sec:discussion}
The superior performance of DP-DT stems from its distinct threat model. While DP-SGD assumes an adversary who observes every gradient update—thereby accumulating privacy cost via composition theorems and necessitating a conservative privacy budget—DP-DT operates under the Hidden State Assumption. Under this assumption, the adversary only observes the final published model. Consequently, privacy loss is determined solely by the final state rather than the cumulative number of training steps, allowing the privacy cost to remain bounded rather than growing indefinitely. This threat model is well-suited for standard deployment scenarios where only the final model is released. However, it does not cover settings where intermediate states are exposed, such as federated learning with untrusted aggregators or systems that log intermediate checkpoints; in such cases, composition-based accounting remains necessary.

The primary computational overhead of DP-DT arises from training and storing $K$ auxiliary models. While this cost is manageable on modern hardware for the moderate-scale architectures evaluated in this work ($K=40$ for ResNet18 or ViT-Small), storing $K$ full replicas becomes prohibitive for full-parameter fine-tuning of large language models (LLMs). To mitigate this bottleneck in our SST-2 experiments, we leverage LoRA, which restricts trainable parameters to low-rank adapters and substantially reduces the per-model memory footprint. Scaling DP-DT to full-parameter LLM training remains an important direction for our future work.

\section{Conclusion}

In this work, we introduced \textit{Differentially Private Decoupled Training (DP-DT)} to address the critical utility bottleneck in private deep learning.  Departing from standard gradient perturbation paradigms, DP-DT leverages auxiliary models that are continuously synchronized with the global model to perform noise-free representation extraction. This strategy effectively isolates feature learning from privacy noise, ensuring that the final model retains high utility.
Theoretically, We derive a rigorous privacy accountant for DP-DT under the Hidden State Assumption (HSA), theoretically proving that the privacy loss converges to a stable bound. To the best of our knowledge, this represents the first extension of HSA's applicability to deep learning, moving beyond the convex assumptions of prior works.
Empirically, DP-DT establishes a new state-of-the-art on benchmarks such as CIFAR-10 and CIFAR-100, significantly outperforming existing baselines.




\bibliography{main}
\appendix
\section{Notation Table}
\label{sec:notation_table}

Table~\ref{tab:notation} summarizes the key symbols used throughout this paper.
The ``Defined in'' column links to the subsection where each symbol is first introduced.

\begin{table}[htbp]
    \small
    \centering
    \caption{Summary of Notation}
    \label{tab:notation}
    \begin{tabular}{c p{0.65\textwidth} l}
    \toprule
    \footnotesize
    \textbf{Symbol} & \textbf{Meaning} & \textbf{Defined in} \\
    \midrule
    \multicolumn{3}{c}{\textit{Core Variables}} \\
    $\paramVec$     & Global publishable model parameters & Section \ref{subsec:pre} \\
    $\auxParam$     & $k$-th auxiliary model parameters   & Section \ref{subsec:decoupling} \\
    $\LagMult_k$    & Lagrange multiplier for auxiliary $k$ & Section \ref{subsec:decoupling} \\
    $\Aset$         & Set of auxiliary models $\{\auxParam\}_{k=1}^{\numAux}$ & Section \ref{subsec:decoupling} \\
    $\PiSet$        & Set of Lagrange multipliers $\{\LagMult_k\}_{k=1}^{\numAux}$ & Section \ref{subsec:decoupling} \\
    $\Pset$         & Set of penalty coefficients $\{\rhoK\}_{k=1}^{\numAux}$ & Section \ref{subsec:decoupling} \\
    \midrule
    \multicolumn{3}{c}{\textit{Datasets \& Indices}} \\
    $\Dset_{\text{train}}$ & Training dataset & Section \ref{subsec:pre} \\
    $\numAux$       & Number of auxiliary models ($K$) & Section \ref{subsec:decoupling} \\
    $\numIter$      & Mini-batches per epoch ($M$)     & Section \ref{subsec:decoupling} \\
    $\numEpochs$    & Number of training epochs ($E$)   & Section \ref{subsec:dpdt} \\
    \midrule
    \multicolumn{3}{c}{\textit{Loss Functions}} \\
    $\loss$         & Empirical risk & Section \ref{subsec:pre} \\
    $\Lag$          & Augmented Lagrangian & Section \ref{subsec:decoupling} \\
    \midrule
    \multicolumn{3}{c}{\textit{Optimization}} \\
    $\effStep$      & Effective step size $\widehat{\eta} = \frac{K \cdot \eta_{\paramVec}}{\sum_{k=1}^K \rho_k}$ & Section \ref{subsec:dpdt} \\
    $\lrGlobal$     & Global learning rate $\eta_{\paramVec}$ & Section \ref{subsec:dpdt} \\
    $\lrAux$        & Auxiliary learning rate $\eta_k$ & Section \ref{subsec:dpdt} \\
    $\rhoK$         & Penalty coefficient $\rho_k$ & Section \ref{subsec:decoupling} \\
    $\penDecay$     & Penalty decay factor $\gamma_k \in (0,1]$ & Section \ref{subsec:dpdt} \\
    $\minCurv$      & Minimum estimated curvature $\minCurv = 1/\lambda_{\max}(\mathbf{V}^{(t)})$ & Section \ref{subsec:dpdt} \\
    $L$, $L_h$ & Lipschitz smoothness constants of $\loss$ and $h$ & Sections \ref{sec:descent_lemma} \\
    \midrule
    \multicolumn{3}{c}{\textit{Differential Privacy}} \\
    $\epsPriv, \dpDelta$ & Standard $(\varepsilon,\delta)$-DP parameters & Section \ref{subsec:pre} \\
    $\renyiOrder$   & R\'enyi divergence order $\alpha > 1$ & Section \ref{subsec:pre} \\
    $\noiseStd$     & Algorithm noise std $\sigma_{\text{noise}}$ (actual noise in Alg.~1); $\noiseStd^2 = 2\widehat{\eta}\,\noiseLStd^2$ & Section \ref{subsec:dpdt} \\
    $\noiseLStd$    & Langevin noise parameter $\sigma$ (privacy heat-flow analysis) & Section \ref{subsec:privacy_prelim} \\
    $\clipBound$    & Clipping threshold $C^{(t)}$ & Section \ref{subsec:dpdt} \\
    \midrule
    \multicolumn{3}{c}{\textit{Analysis Constants}} \\
    $\lipF$         & Lipschitz constant of $F$; $L_F = 1 - \eta_{\paramVec}$ & Section \ref{subsec:privacy_accounting} \\
    $\lipT$         & Lipschitz constant of $T$; $L_T \leq 1$ & Section \ref{subsec:privacy_accounting} \\
    $\lsiConst$     & Log-Sobolev inequality constant $c$ & Section \ref{subsec:privacy_accounting} \\
    $\smooth$       & Smoothness constant $\beta$ & Section \ref{sec:related_work} \\
    $\wdCoeff$      & Weight decay coefficient $\lambda$ & Section \ref{sec:related_work} \\
        $\zeta$         & Uniform bound on mini-batch gradient deviation (Assumption~\ref{assume:bounded_deviation}) & Section\ref{sec:error_quantification} \\
    $\xi^{(t)}$     & Stochastic error sequence; $\xi^{(t)} = \|\mathbf{z}^{(t)}\| + \sum_{k} \|e_k^{(t)}\| + c' \sum_{k} \|\kappa_k^{(t)}\|$, summable a.s.\ (Lemma~\ref{lemma:subgradient_bound}) & Section\ref{sec:lyapunov} \\
    $c_k^{(t)}$     & Descent margin constant per auxiliary model; $c_k^{(t)} = \frac{\minCurv}{\eta_k^{(t)}} - \frac{L+\rho_k}{2}$ & Section\ref{sec:descent_lemma} \\
    $c_{aux}^{(t)}$ & Minimum descent margin across auxiliary models; $c_{aux}^{(t)} = \min_k c_k^{(t)}$ & Section\ref{sec:lyapunov} \\
    $c_l^m$         & Per-iteration LSI constant at epoch $l$, mini-batch $m$ & Section\ref{subsec:privacy_accounting} \\
    $\bar{c}$       & KL desingularizing function constant ($\bar{c} > 0$) & Section\ref{sec:stationary} \\
    $c_{kl}$        & KL exponent ($c_{kl} \in [0,1)$) & Section\ref{sec:stationary} \\
    $r_{kl}$        & KL neighborhood radius ($r_{kl} \in (0, \infty]$) & Section\ref{sec:stationary} \\
    \bottomrule
    \end{tabular}
\end{table}

\section{Convergence Analysis}

\subsection{Proof of Lemma \ref{lemma:clipping_bound}} \label{proof:clipping}

\begin{proof} 
We prove two key conclusions one by one.

\textit{1. Bound the Multiplier$\|\LagMult_k^{(t)}\|$ via Triangle Inequality:} For notation simplicity, we let $\alpha^{(t)}_k = \rho^{(t)}_k \eta^{(t)}_k \precondition^{(t)} \nabla_{\auxParam} \gL(\auxParam^{(t)})$. Based on the way of selecting $\eta^{(t)}_k$ (Problem~\ref{eq:st}), we have  $\|{\widehat{\LagMult}^{(t)} - 2\alpha^{(t)}_k}\| \le C^{(t)}$. The update rule (\ref{eq:update_aux}) indicates $\LagMult^{(t+1)}_k = \widehat{\LagMult}^{(t)}_k - \alpha^{(t)}_k$.
By definition, we have $\|\widehat{\LagMult}^{(t)}_k\| \leq C^{(t)}$.
Based on the triangle inequality, we have the following inequality:
\begin{equation}
    \begin{aligned}
        \|\LagMult^{(t+1)}_k\| &= \left\|\frac{1}{2} (\widehat{\LagMult}^{(t)} - 2 \alpha^{(t)}_k) + \frac{1}{2} \widehat{\LagMult}^{(t)}\right\| \leq \frac{1}{2} \left\| (\widehat{\LagMult}^{(t)} - 2 \alpha^{(t)}_k) \right\| + \frac{1}{2} \left\|\widehat{\LagMult}^{(t)}\right\| = C^{(t)}
    \end{aligned}
\end{equation}

This establishes that $\|\LagMult^t\| \le C^{(t-1)}$ holds for any $t \geq 1$.

\textit{2. Bound the Bias Term:} The bias term at step $t$ is $\kappa_k^{(t)} = \LagMult_k^{(t)} -\widehat{\LagMult}_k^{(t)}$. 
Based on the relationship between $\|\LagMult_k^{(t)}\|$ and $C^{(t)}$, we discuss the following two cases:

\textbf{Case 1:} If $\|\LagMult^{(t)}\| \le C^{(t)}$, the clipping function is an identity mapping, i.e., $\LagMult_k^{(t)} = \widehat{\LagMult}_k^{(t)}$. Consequently, $\kappa_k^{(t)} = \mathbf{0}$, and we have $\|\kappa_k^{(t)}\| = 0 \le C^{(t-1)} - C^{(t)}$, since the sequence is non-increasing.

\textbf{Case 2:} If $\|\LagMult^{(t)}\| > C^{(t)}$, the clipping function scales the vector: $\widehat{\LagMult}^{(t)}_k =  \frac{C^{(t)} \cdot \LagMult^{(t)}_k}{\|\LagMult^{(t)}_k\|}$. We substitute this directly into the definition of $\kappa^{(t)}_k$ to calculate its norm:
\begin{equation*}
    \|\kappa^{(t)}_k\| = \left\|\LagMult^{(t)} - \frac{C^{(t)} \cdot \LagMult^{(t)}_k}{\|\LagMult^{(t)}_k\|}\right\| = \left(1 - \frac{C^{(t)}}{\|\LagMult^{(t)}_k\|}\right) \|\LagMult^{(t)}_k\| = \|\LagMult^{(t)}_k\| - C^{(t)}.
\end{equation*}
The second equality of the Equation above is based on the scalar multiplier $\left(1 - \frac{C^{(t)}}{\|\LagMult^t\|}\right) > 0$ , because $\|\LagMult^{(t)}\| > C^{(t)}$ for this case. 

Using our proven bound $\|\LagMult^{(t)}_k\| \le C^{(t-1)}$, we immediately obtain: $\|\kappa_k^{(t)}\| \le C^{(t-1)} - C^{(t)}$
\end{proof}

\subsection{Proof of Lemma~\ref{lemma:descent_local}} \label{proof:descent_lemma}
\begin{proof}
Since $\auxParam^{(t)}$ is set to $\paramVec^{(t)}$ in Algorithm~\ref{alg:dp_decoupled} before its update, the penalty term vanishes in $\gL(\auxParam^{(t)})$ as defined in Equation (\ref{equ:alm_form}), so the local gradient is $\LagMult_k^{(t)} + \nabla_{\auxParam} \loss(\auxParam^{(t)}; \gD^{k}_m)$.
In addition, we let $e^{(t)}_k = \nabla_{\auxParam} \loss(\auxParam^{(t)}; \gD^{k}_m) - \nabla_{\auxParam} \loss(\auxParam^{(t)})$. Assumption~\ref{assume:bounded_deviation} indicates $\|e^{(t)}_k\| \leq \zeta$.
The local update rule in (\ref{eq:update_aux}) yields:
\begin{equation*}
    \auxParam^{(t+1)} - \auxParam^{(t)} = \auxParam^{(t+1)} - \paramVec^{(t)} = - \eta^{(t)}_k \precondition^{(t)} \left( \nabla_{\auxParam} \loss(\auxParam^{(t)}) + e^{(t)}_k + \LagMult^{(t)}_k \right)
\end{equation*}
Rearranging this equation, we express the exact full gradient of the empirical risk as:
\begin{equation}
    \nabla_{\auxParam} \loss(\auxParam^{(t)}) = -(\eta^{(t)}_k \precondition^{(t)})^{-1} (\auxParam^{(t+1)} - \auxParam^{(t)}) - \LagMult^{(t)}_k - e_k^{(t)} \label{eq:grad_R_sub}
\end{equation}
Since $\loss(\auxParam)$ is $L$-smooth, applying the standard Descent Lemma yields:
\begin{equation} \label{eq:s_descent_lemma}
    \loss(\auxParam^{(t+1)}) \le \loss(\auxParam^{(t)}) + \langle \nabla_{\auxParam} \loss(\paramVec^{(t)}), \auxParam^{(t+1)} - \auxParam^{(t)} \rangle + \frac{L}{2} \|\auxParam^{(t+1)} - \auxParam^{(t)}\|^2
\end{equation}
Substituting Equation (\ref{eq:grad_R_sub}) into the inner product term and considering the preconditioner term $0 < \precondition^{(t)} \leq \minCurv^{-1} \rmI$, we have:
\begin{equation}
    \langle \nabla_{\auxParam} \loss(\auxParam^{(t)}), \auxParam^{(t+1)} - \auxParam^{(t)} \rangle \leq -\minCurv (\eta^{(t)}_k)^{-1} \|\auxParam^{(t+1)} - \auxParam^{(t)}\|^2 - \langle \LagMult^{(t)}_k + e_k^{(t)}, \auxParam^{(t+1)} - \auxParam^{(t)} \rangle \label{eq:inner_prod_R}
\end{equation}
Now, we evaluate the difference in the augmented Lagrangian $\Lag$ as defined in (\ref{equ:alm_form}). Considering $\auxParam^{(t)}$ is set to $\paramVec^{(t)}$, we have the following equation:
\begin{equation} \label{eq:loss_diff}
    \Lag(\auxParam^{(t+1)}) - \Lag(\auxParam^{(t)}) = \frac{1}{K} \bigg[\loss(\auxParam^{(t+1)}) - \loss(\auxParam^{(t)}) + \langle \LagMult^{(t)}_k, \auxParam^{(t+1)} - \auxParam^{(t)} \rangle + \frac{\rho_k^{(t)}}{2} \|\auxParam^{(t+1)} - \auxParam^{(t)}\|^2 \bigg]
\end{equation}
Substituting (\ref{eq:s_descent_lemma}), (\ref{eq:inner_prod_R}) into (\ref{eq:loss_diff}), we observe that the cross-terms involving the dual variable $\LagMult^{(t)}_k$ perfectly cancel out and obtain the following inequality:
\begin{equation}
\begin{aligned}
    \Lag(\auxParam^{(t+1)}) - \Lag(\auxParam^{(t)}) &\le \frac{1}{K} \bigg[-\minCurv(\eta^{(t)}_k)^{-1} \|\auxParam^{(t+1)} - \auxParam^{(t)}\|^2 - \langle \LagMult^{(t)}_k + e^{(t)}_k, \auxParam^{(t+1)} - \auxParam^{(t)} \rangle \\
    &\quad + \frac{L}{2} \|\auxParam^{(t+1)} - \auxParam^{(t)}\|^2 + \langle \LagMult^{(t)}_k, \auxParam^{(t+1)} - \auxParam^{(t)} \rangle + \frac{\rho_k^{(t)}}{2} \|\auxParam^{(t+1)} - \auxParam^{(t)}\|^2 \bigg] \\
    &= \frac{1}{K}\Bigg[- \underbrace{\left( \frac{\minCurv}{\eta^{(t)}_k} - \frac{L+\rho_k^{(t)}}{2} \right)}_{:= c^{(t)}_k} \|\auxParam^{(t+1)} - \auxParam^{(t)}\|^2 - \langle e_k^{(t)}, \auxParam^{(t+1)} - \auxParam^{(t)} \rangle \Bigg]
\end{aligned} \label{eq:bound_descent_lemma}
\end{equation}
The right hand side of the inequality above is a quadratic in terms of $\|\auxParam^{(t + 1)} - \auxParam^{(t)}\|$.
Since $\eta_k^{(t)} \leq \eta_k < \frac{2 \minCurv}{L+\rho_k^{(t)}}$, we have $c_k^{(t)} > 0$.
Therefore, we can further bound the right hand side of (\ref{eq:bound_descent_lemma}) by:
\begin{equation}
    \Lag(\auxParam^{(t+1)}) - \Lag(\auxParam^{(t)}) \leq \frac{\|e^{(t)}_k\|^2}{4 c_k^{(t)}} \leq \frac{\zeta^2}{4\left( \frac{\minCurv}{\eta^{(t)}_k} - \frac{L+\rho_k^{(t)}}{2} \right)}
\end{equation}

The second inequality is based on $\|e^{(t)}_k\| \leq \zeta$ as indicated in Assumption~\ref{assume:bounded_deviation}.

\end{proof}

\subsection{Proof of Lemma~\ref{lemma:dual_variable}} \label{proof:dual_variable}

\begin{proof}
We observe that $\Lag$ is linear with respect to $\LagMult_k$, and includes a scaling factor $\frac{1}{K}$. Therefore, we have the following:
\begin{equation*}
    \Lag(\paramVec^{(t)}, \auxParam^{(t+1)}, \LagMult_k^{(t+1)}) - \Lag(\paramVec^{(t)}, \auxParam^{(t+1)}, \widehat{\LagMult}_k^{(t)}) = \frac{1}{K} \langle \LagMult_k^{(t+1)} - \widehat{\LagMult}_k^{(t)}, \auxParam^{(t+1)} - \paramVec^{(t)} \rangle
\end{equation*}
Substituting $\LagMult_k^{(t+1)} - \widehat{\LagMult}_k^{(t)} = \rho_k^{(t)} (\auxParam^{(t + 1)} - \paramVec^{(t)})$ in update rule~\ref{eq:update_aux}, we have the following:
\begin{equation*}
    \Lag(\paramVec^{(t)}, \auxParam^{(t+1)}, \LagMult_k^{(t+1)}) - \Lag(\paramVec^{(t)}, \auxParam^{(t+1)}, \widehat{\LagMult}_k^{(t)}) = \frac{\rho^{(t)}_k}{K}\|\auxParam^{(t + 1)} - \paramVec^{(t)}\|^2
\end{equation*}
Similarly, 
\begin{equation*}
    \Lag(\paramVec^{(t)}, \auxParam^{(t + 1)}, \widehat{\LagMult}_k^{(t)}) - \Lag(\paramVec^{(t)}, \auxParam^{(t+1)}, \LagMult_k^{(t)}) = \frac{1}{K} \langle \widehat{\LagMult}_k^{(t)} - \LagMult_k^{(t)}, \auxParam^{(t+1)} - \paramVec^{(t)} \rangle
\end{equation*}
Substituting $\widehat{\LagMult}_k^{(t)} = \LagMult_k^{(t)} + \kappa_k^{(t)}$ into the linear expansion of $\Lag$ directly yields the result.
\end{proof}

\subsection{Proof for Lemma~\ref{lemma:expected_descent_dp}} \label{proof:expected_descent_dp}

\begin{proof}
$h(\paramVec)$ is $L_h$ Lipschitz smooth, so we have:
\begin{equation}
\begin{aligned} \label{eq:proof_update_theta_l}
    \Lag(\paramVec^{(t + 1)}) - \Lag(\paramVec^{(t)}) &= h(\paramVec^{(t + 1)}) + r(\paramVec^{(t + 1)}) - h(\paramVec^{(t)}) - r(\paramVec^{(t)}) \\
    &\leq \langle \nabla_{\paramVec} h(\paramVec^{(t)}), \paramVec^{(t + 1)} - \paramVec^{(t)} \rangle + \frac{L_h}{2} \|\paramVec^{(t + 1)} - \paramVec^{(t)}\|^2 + r(\paramVec^{(t + 1)}) - r(\paramVec^{(t)})
\end{aligned}
\end{equation}
We have $\paramVec^{(t + 1)} = \argmin_{\paramVec} \big\{r(\paramVec) + \frac{1}{2\effStep^{(t)}} \|\paramVec - (\paramVec^{(t)} - \effStep^{(t)} (\nabla_{\paramVec} h(\paramVec) + \vz_t))\|^2\big\}$ based on the proximal operator in the update rule (\ref{equ:theta_update}). We let $g_r$ be a subgradient for $r$, i.e., $g_r \in \partial r(\paramVec^{(t + 1)})$, then we have:
\begin{equation} \label{eq:proof_proximal}
    g_r + \frac{1}{\effStep^{(t)}} \big(\paramVec^{(t + 1)} - (\paramVec^{(t)} - \effStep^{(t)} (\nabla_{\paramVec} h(\paramVec) + \vz_t))\big) = 0
\end{equation}
Substituting (\ref{eq:proof_proximal}) into (\ref{eq:proof_update_theta_l}), we have:
\begin{equation}
\begin{aligned} \label{eq:prox_standard_descent}
    \Lag(\paramVec^{(t + 1)}) &- \Lag(\paramVec^{(t)}) \\
    &\leq -\left(\frac{1}{\effStep^{(t)}} - \frac{L_h}{2}\right) \|\paramVec^{(t + 1)} - \paramVec^{(t)}\|^2 - \langle g_r + \vz_t, \paramVec^{(t + 1)} - \paramVec^{(t)} \rangle + r(\paramVec^{(t + 1)}) - r(\paramVec^{(t)}) \\
    &\leq - \left(\frac{1}{\effStep^{(t)}} - \frac{L_h}{2}\right) \|\paramVec^{(t + 1)} - \paramVec^{(t)}\|^2 - \langle  \vz_t, \paramVec^{(t + 1)} - \paramVec^{(t)} \rangle
\end{aligned}
\end{equation}
The last inequality arises from the convexity of $r(\paramVec)$: $r(\paramVec^{(t+1)}) + \langle g_r, \paramVec^{(t)} - \paramVec^{(t + 1)} \rangle \leq r(\paramVec^{(t)})$.

To decouple the noise dependency in $\langle \bm{z}_t, \paramVec^{(t+1)} - \paramVec^{(t)} \rangle$, introduce the virtual noise-free update $\widetilde{\vtheta}^{(t+1)} = \text{Prox}_{\effStep^{(t)}, r}(\paramVec^{(t)} - \effStep^{(t)}\nabla h(\paramVec^{(t)}))$ and write:
\begin{equation*}
    -\langle \bm{z}_t, \paramVec^{(t+1)} - \paramVec^{(t)} \rangle = \underbrace{-\langle \bm{z}_t, \widetilde{\vtheta}^{(t+1)} - \paramVec^{(t)} \rangle}_{\text{Term A}} \;+\; \underbrace{-\langle \bm{z}_t, \paramVec^{(t+1)} - \widetilde{\vtheta}^{(t+1)} \rangle}_{\text{Term B}}
\end{equation*}
Term~A has zero mean since $\widetilde{\vtheta}^{(t+1)}$ is $\mathcal{F}_t$-measurable and $\bm{z}_t$ is independent zero-mean noise. For Term~B, the proximal operator is 1-Lipschitz, so $\|\paramVec^{(t+1)} - \widetilde{\vtheta}^{(t+1)}\| \le \|\effStep^{(t)}\bm{z}_t\|$, giving the bound $\effStep^{(t)}\|\bm{z}_t\|^2$ via Cauchy--Schwarz. Taking expectations, $\mathbb{E}[\|\bm{z}_t\|^2 \mid \mathcal{F}_t] = d\noiseStd^2$, yielding the final inequality.
\end{proof}

\subsection{Proof for Lemma~\ref{lemma:global_descent}}
\label{sec:proof_global_descent}

\begin{lemma}[Reset Gap Bound]
\label{lemma:reset}
Let $\delta_k^{(t+1)}= \auxParam^{(t+1)} - \paramVec^{(t+1)}$, $\Delta\auxParam^{(t)} = \auxParam^{(t+1)} - \paramVec^{(t)}$, $\Delta\paramVec^{(t)} = \paramVec^{(t+1)} - \paramVec^{(t)}$(where $\paramVec^{(t)}=\auxParam^{(t)}$), the Reset operation in Algorithm \ref{alg:dp_decoupled} leads to following gap on each iteration 
\begin{equation}
\begin{aligned}
 \sum_{t=0}^{\infty}\E[ G^{(t)} | \gF_t]  &= \E\left[ \gL(\paramVec^{(t+1)}, \paramVec^{(t + 1)}, \LagMult^{(t + 1)}) - \gL(\paramVec^{(t+1)}, \auxParam^{(t+1)}, \LagMult^{(t+1)}) | \gF_t\right] <  \infty \label{equ:reset}
\end{aligned}
\end{equation}
\end{lemma}

\begin{proof}
Due to the sensitivity constraint Equation  (\ref{eq:st}) in Algorithm~\ref{alg:dp_decoupled}, we have
$\|\text{clip}(\LagMult_k^{(t)}) + 2\rho_k^{(t)}\Delta\auxParam^{(t)}\| \le C^{(t)}$.
By the triangle inequality, $2\rho_k^{(t)}\|\Delta\auxParam^{(t)}\| \le C^{(t)} + \|\text{clip}(\LagMult_k^{(t)})\| \le 2C^{(t)}$.
Hence $\|\Delta\auxParam^{(t)}\| \le C^{(t)}/\rho_k^{(t)} \le \frac{C^{(0)}}{\rho_k^{(0)}}\gamma^t$.
For $\|\Delta\paramVec^{(t)}\|$, the global update Equation (\ref{equ:theta_update}) with the proximal operator's contrastive property gives
$\|\Delta\paramVec^{(t)}\| \le \widehat{\eta}^{(t)}\|\nabla h(\paramVec^{(t)}) + \mathbf{z}^{(t)}\|$.
From the definition of $h$, $\|\nabla h\| \le \frac{1}{K}\sum_k(\|\LagMult_k^{(t+1)}\| + \rho_k^{(t)}\|\Delta\auxParam^{(t)}\|) \le 2C^{(0)}$.
Since $\widehat{\eta}^{(t)} \propto \gamma^t$, we obtain $\E[\|\Delta\paramVec^{(t)}\|\mid\gF_t] \leq O(\gamma^t)$.
Moreover, due to the smoothness of $\loss(\paramVec)$, we have $\|\nabla \loss(\paramVec^{(t+1)})\| \le \|\nabla \loss(\paramVec^{(0)}) \| + L(\|\paramVec^{(t+1)}-\paramVec^{(0)}\|) = \|\nabla \loss(\paramVec^{(0)}) \| + L \|\sum_{i = 1}^t \Delta \paramVec^{(t)}\|$. Since $\|\Delta \paramVec^{(t)}\| \leq O(\gamma^t)$ and $\gamma \in (0, 1)$,  $\exists G_{\loss} < \infty$ $\|\nabla \loss(\paramVec^{(t+1)})\| \leq G_{\loss}$ is then bounded.
Finally $\delta_k^{(t+1)}= \Delta\auxParam^{(t)} - \Delta\paramVec^{(t)}$, it follows that
$\E[\|\delta_k\|\mid\gF_t] \leq O(\gamma^t)$ and $\E[\|\delta_k\|^2\mid\gF_t] \leq  O(\gamma^{2t})$. Now we have :
\begin{equation}
\begin{aligned}
\E[|G^{(t)}|\mid\gF_t]&=  \frac{1}{K}\sum_{k=1}^{K}\E\Big[\loss(\paramVec^{(t+1)}) - \loss(\auxParam^{(t+1)}) - \langle\LagMult_k^{(t)},\delta_k\rangle - \frac{\rho_k^{(t)}}{2}\|\delta_k\|^2 |\gF_t \Big]     \\
&\le \frac{1}{K}\sum_{k=1}^{K}\E\Big[\big|\langle\nabla\loss(\auxParam^{(t+1)}) + \LagMult_k^{(t+1)}, -\delta_k\rangle\big| + \frac{L-\rho_k^{(t)}}{2}\|\delta_k\|^2 | \gF_t \Big] \\
&\le \frac{1}{K}\sum_{k=1}^{K}\Big[(G_{\loss}+C^{(0)})\,\E[\|\delta_k\|\mid\gF_t] + \frac{L-\rho_k^{(t)}}{2}\E[\|\delta_k\|^2\mid\gF_t]\Big] \\
& \le \frac{1}{K}\sum_{k=1}^{K}\Big[\left(G_{\loss}+C^{(0)} - \frac{\rho_k^{(0)}}{2}\right)O(\gamma^t) + \frac{L}{2} O(\gamma^{2t})\Big]
\end{aligned} \label{eq:bound_G}
\end{equation}

The first inequality is based on the Lipschitz smoothness. The second inequality is based on the bound of $\nabla\loss(\auxParam^{(t+1)})$. The last inequality is based on $\rho_k^{(t)} = \gamma^t \rho^{(0)}_t$.
Since $\gamma \in (0, 1)$, the bound on the right hand side of (\ref{eq:bound_G}) is dominated by the first-order term. Therefore, $\exists M<\infty$ independent of $t$ such that
$\E[|G^{(t)}|\mid\gF_t] \le M\gamma^t$. Summing over all iterations:
\begin{equation}
\sum_{t=0}^{\infty} \E[|G^{(t)}|\mid\gF_t] \le M\sum_{t=0}^{\infty}\gamma^t = \frac{M}{1-\gamma} < \infty,
\end{equation}
which completes the proof.

\end{proof}

\begin{proof}
From Lemma~\ref{lemma:descent_local} and Lemma~\ref{lemma:dual_variable}, for each auxiliary model indexed by $k$, we have 
\begin{equation} 
\begin{aligned}
\gL(\paramVec^{(t)}, \auxParam^{(t + 1)}, \LagMult^{(t + 1)}) - \gL(\paramVec^{(t)}, \auxParam^{(t)}, \LagMult^{(t)}) &\leq \\
\frac{1}{K}\big[ -c^{(t)}_k\|\Delta\auxParam^{(t)}\|^2 - \langle e^{(t)}_k,\Delta\auxParam^{(t)}\rangle &+ \rho^{(t)}_k\|\Delta\auxParam^{(t)}\|^2 + \langle\kappa^{(t)}_k,\Delta\auxParam^{(t)}\rangle \big]
\end{aligned}
\end{equation}
where $\Delta\auxParam^{(t)} = \auxParam^{(t+1)}-\paramVec^{(t)}$ and $c^{(t)}_k = \frac{\minCurv}{\eta^{(t)}_k} - \frac{L+\rho^{(t)}_k}{2}$ as they are defined in Lemma~\ref{lemma:descent_local}. By Young's inequality,
$-\langle e^{(t)}_k,\Delta\auxParam^{(t)}\rangle \le \frac{c^{(t)}_k}{2}\|\Delta\auxParam^{(t)}\|^2 + \frac{\|e^{(t)}_k\|^2}{2c^{(t)}_k}$,
$\langle\kappa^{(t)}_k,\Delta\auxParam^{(t)}\rangle \le \frac12\|\Delta\auxParam^{(t)}\|^2 + \frac{1}{2}\|\kappa^{(t)}_k\|^2$.
Collecting $\|\Delta\auxParam^{(t)}\|^2$ terms, where 
$A^{(t)}_k  = \frac{1}{K}\Bigl(\frac{c^{(t)}_k}{2} - \rho^{(t)}_k - \frac12\Bigr)$,
leaving $\frac{\|e^{(t)}_k\|^2}{2Kc^{(t)}_k} + \frac{\|\kappa^{(t)}_k\|^2}{2K}$ as residuals per model:
\begin{equation}\label{equ:delta_k}
    \gL(\paramVec^{(t)}, \auxParam^{(t + 1)}, \LagMult^{(t + 1)}) - \gL(\paramVec^{(t)}, \auxParam^{(t)}, \LagMult^{(t)}) \leq -A^{(t)}_k \|\Delta \auxParam^{(t)}\|^2 + \frac{1}{K} \big[ \frac{1}{2}\|\kappa^{(t)}_k\|^2 + \frac{1}{2 c^{(t)}_k}\|e^{(t)}_k\|^2 \big]
\end{equation}
By Assumption~\ref{assume:bounded_deviation}, we have $\|e^{(t)}_k\|^2 \leq \zeta^2$. Therefore, we have $\sum_{k=1}^K \frac{\|e^{(t)}_k\|^2}{2Kc^{(t)}_k} \le \frac{\zeta^2}{2c_{\text{aux}}^{(t)}}$ where $c_{\text{aux}}^{(t)} = \min_k c_k^{(t)}$.
Summing over auxiliary update (Equation (\ref{equ:delta_k})), global update (Inequality~(\ref{eq:expected_descent_dp}) in Lemma \ref{lemma:expected_descent_dp}) and reset operator(Inequality~(\ref{equ:reset}) in Lemma \ref{lemma:reset}), taking conditional expectations, we obtain
\begin{equation*}
    \mathbb{E}[\widetilde{\Phi}_{t+1} \mid \mathcal{F}_t]
\le \widetilde{\Phi}_t - \sum_{k=1}^K A^{(t)}_k\|\Delta\auxParam^{(t)}\|^2 - B^{(t)}\,\mathbb{E}[\|\paramVec^{(t + 1)} - \paramVec^{(t)}\|^2 \mid \mathcal{F}_t] + \widehat{\eta}^{(t)2} d \sigma^2_{noise} + \sum_{k=1}^K\frac{\|\kappa^{(t)}_k\|^2}{2K}
\end{equation*}
which is (\ref{eq:clean_descent}), where $A_k^{(t)}$ , $B^{(t)}$ is geometrically summable. In addition,  we have $\sum_t\|\kappa^{(t)}_k\|^2 \le (C^{(0)}-C^{(\infty)})^2 < \infty$ (Lemma~\ref{lemma:clipping_bound}) since $\{C^{(t)}\}_t$ is monotonically non-increasing. Finally, we consider the reset-gap error and add $\E[ G^{(t)} | \gF_t]$ in above inequality.

Considering the definition of $c_k^{(t)}$ in Lemma~\ref{lemma:descent_local}, when $A^{(t)}_k>0$, we have $\eta^{(t)}_k < 2\minCurv/(L+5\rho^{(t)}_k+2)$.


\end{proof}

\subsection{Almost Sure Convergence of Lyapunov Function}
\label{proof:lyapunov_convergence}


Denote $\widehat{\LagMult}_k^{(t)} = \text{clip}(\LagMult_k^{(t)}, \clipBound)$. For the dual ascent part in Algorithm \ref{alg:dp_decoupled}, we also have the following lemma:

\begin{proposition}[Summability of DP Noise Variance via Penalty Update]
\label{prop:summability_Bt_exact}
The infinite series $\sum_{t=1}^{\infty} \effStep^{(t)2} d \noiseStd^2$ converges.
\end{proposition}
\begin{proof}
With $\rho_k^{(t)} = \rho_k^{(0)}\gamma^{-t}$ ($\gamma \in (0,1)$), the effective step size is $\effStep^{(t)} = K\eta_\theta / \sum_k \rho_k^{(0)} \gamma^{-t} = \effStep^{(t)}_0 \gamma^t$. Hence $\effStep^{(t)2} d \noiseStd^2 = (\effStep^{(t)}_0)^2 d \noiseStd^2 (\gamma^2)^t$, a geometric series with ratio $\gamma^2 < 1$, so $\sum_t \effStep^{(t)2} d \noiseStd^2 < \infty$

\end{proof}

\begin{proof}[Proof of Theorem \ref{thm:robbins_siegmund}]
 To apply the classical Robbins-Siegmund supermartingale convergence theorem\cite{robbins1971convergence}, we construct a non-negative sequence. Let $U_t \triangleq \widetilde{\Phi}_t - \widetilde{\Phi}_{\min}$. Since $\widetilde{\Phi}_t$ is bounded from below by $\widetilde{\Phi}_{\min}$, it strictly holds that $U_t \ge 0$ for all $t \ge 0$.

Subtracting the constant $\widetilde{\Phi}_{\min}$ from both sides of the expected descent inequality in Lemma \ref{lemma:global_descent}, we obtain:
\begin{equation*}
    \label{eq:rs_standard_form}
    \mathbb{E}[U_{t+1} \mid \mathcal{F}_t] \le U_t - X_t + C_{residual}^{(t)}
\end{equation*}
where $X_t$ is defined as the total deterministic descent evaluated at step $t$:
\begin{equation*}
    X_t \triangleq \sum_{k=1}^K A_k^{(t)} \|\auxParam^{(t+1)} - \paramVec^{(t)}\|^2 + B^{(t)} \|\paramVec^{(t+1)} - \paramVec^{(t)}\|^2 \ge 0
\end{equation*}
From our previous analysis, we have $\E[ G^{(t)} | \gF_t] < \infty$. 
Equation \ref{eq:rs_standard_form} perfectly matches the standard form of the Robbins-Siegmund theorem: a non-negative stochastic process $\{U_t\}$ subjected to a non-negative measurable penalty $X_t$ and a summable positive perturbation $\effStep^{(t)2} d \noiseStd^2$. According to the Robbins-Siegmund theorem, we draw two immediate conclusions:
First, the sequence $U_t$ converges almost surely to a finite, non-negative random variable $U^*$. Since $\widetilde{\Phi}_t = U_t + \widetilde{\Phi}_{\min}$, this implies that $\lim_{t \to \infty} \widetilde{\Phi}_t = \widetilde{\Phi}^*$ a.s., where $\widetilde{\Phi}^* = U^* + \widetilde{\Phi}_{\min}$ is finite. Second, the infinite series of the penalty term $X_t$ is almost surely convergent:
\begin{equation*}
    \sum_{t=0}^\infty X_t = \sum_{t=0}^\infty \left( \sum_{k=1}^K A_k^{(t)} \|\auxParam^{(t+1)} - \paramVec^{(t)}\|^2 + B \|\paramVec^{(t+1)} - \paramVec^{(t)}\|^2 \right) < \infty \quad \text{a.s.}
\end{equation*}
Given that the descent constants $A_k^{(t)}$ and $B$ are strictly positive ($A_k^{(t)} > 0, B > 0$), we can lower bound the sum by scaling out the minimum coefficient. This directly necessitates that the individual sum of squared norms for both global and auxiliary parameter updates must be finite:
\begin{equation*}
    \sum_{t=0}^\infty \|\paramVec^{(t+1)} - \paramVec^{(t)}\|^2 < \infty \quad \text{a.s., \quad and \quad} \sum_{t=0}^\infty \|\auxParam^{(t+1)} - \paramVec^{(t)}\|^2 < \infty \quad \text{a.s.}
\end{equation*}
By the necessary condition for the convergence of an infinite series, the general terms must converge to zero, yielding the asymptotic regularity $\lim_{t \to \infty} \|\paramVec^{(t+1)} - \paramVec^{(t)}\| = 0$ a.s.\ and $\lim_{t \to \infty} \|\auxParam^{(t+1)} - \paramVec^{(t)}\| = 0$ a.s. This completes the proof.   
\end{proof}

\subsection{Proof of Relative Error Condition and Subgradient Bound}
\label{sec:subgradient}

\begin{proof}[Proof of Lemma \ref{lemma:subgradient_bound}]
Since $C^{(t)}$ and the tail sum are independent of $\mathbf{w}$, $\partial\widetilde{\Phi}^{(t+1)}(\mathbf{w}^{(t+1)}) = \partial\Lag^{(t+1)}(\mathbf{w}^{(t+1)})$. We bound each partial subgradient of $\Lag$.

\paragraph{1. Auxiliary models $\auxParam$:}
From Lemma~\ref{lemma:descent_local} (Eq.~\ref{eq:grad_R_sub}), $\nabla\lossk(\vtheta^{(t)}) = -\frac{\minCurv}{\eta_k}(\auxParam^{(t+1)} - \vtheta^{(t)}) - \vpi_k^{(t)} - e_k^{(t)}$.
Adding this into $\nabla_{\auxParam}\Lag(\mathbf{w}^{(t+1)})$, substituting the dual ascent rule $\vpi_k^{(t+1)} = \widehat{\vpi}_k^{(t)} + \rho_k^{(t)}(\auxParam^{(t+1)} - \vtheta^{(t)})$, and expanding $\rho_k^{(t)}(\auxParam^{(t+1)} - \vtheta^{(t+1)}) = \rho_k^{(t)}(\auxParam^{(t+1)} - \vtheta^{(t)}) - \rho_k^{(t)}(\vtheta^{(t+1)} - \vtheta^{(t)})$ yields:
\begin{equation*}
    K\nabla_{\auxParam}\Lag = [\nabla\lossk(\auxParam^{(t+1)}) - \nabla\lossk(\vtheta^{(t)})] - (\tfrac{\minCurv}{\eta_k} - 2\rho_k^{(t)})(\auxParam^{(t+1)} - \vtheta^{(t)}) - \rho_k^{(t)}(\vtheta^{(t+1)} - \vtheta^{(t)}) - \kappa_k^{(t)} - e_k^{(t)}
\end{equation*}
With $L$-smoothness and the triangle inequality, $\|\nabla_{\auxParam}\Lag\| \le \frac{1}{K}(L + |\frac{\minCurv}{\eta_k} - 2\rho_k^{(t)}|)\|\auxParam^{(t+1)} - \vtheta^{(t)}\| + \frac{\rho_k^{(t)}}{K}\|\vtheta^{(t+1)} - \vtheta^{(t)}\| + \frac{1}{K}\|\kappa_k^{(t)}\| + \frac{1}{K}\|e_k^{(t)}\|$

\paragraph{2. Multipliers $\vpi_k$:}
$\|\nabla_{\vpi_k}\Lag\| = \frac{1}{K}\|\auxParam^{(t+1)} - \vtheta^{(t+1)}\| \le \frac{1}{K}(\|\auxParam^{(t+1)} - \vtheta^{(t)}\| + \|\vtheta^{(t+1)} - \vtheta^{(t)}\|)$. Reversing the dual update yields $\|\auxParam^{(t+1)} - \vtheta^{(t)}\| \le \frac{1}{\rho_k^{(t)}}(\|\vpi_k^{(t+1)} - \vpi_k^{(t)}\| + \|\kappa_k^{(t)}\|)$, so $\|\nabla_{\vpi_k}\Lag\| \le \frac{1}{K\rho_k^{(t)}}\|\vpi_k^{(t+1)} - \vpi_k^{(t)}\| + \frac{1}{K}\|\vtheta^{(t+1)} - \vtheta^{(t)}\| + \frac{1}{K\rho_k^{(t)}}\|\kappa_k^{(t)}\|$

\paragraph{3. Global model $\vtheta$:}
From $\partial_{\vtheta}\Lag = \partial r(\vtheta^{(t+1)}) + \nabla h(\vtheta^{(t+1)})$ and $g_r\in\partial r(\vtheta^{(t+1)})$, we have $\partial_{\vtheta}\Lag \ni \nabla h(\vtheta^{(t+1)}) + g_r$. By $L_h$-smoothness of $h$ and the sensitivity bound $\|\nabla h(\vtheta^{(t)})\| \le C$, $\|\partial_{\vtheta}\Lag\| \le L_h\|\vtheta^{(t+1)}-\vtheta^{(t)}\| + C + \|g_r\|$

\paragraph{4. Aggregation:}
The bound becomes $\text{dist}(\mathbf{0}, \partial\widetilde{\Phi}^{(t+1)}) \le b\|\mathbf{w}^{(t+1)} - \mathbf{w}^{(t)}\| + \xi^{(t)} + \widetilde{C}$, where $\widetilde{C}=C + \max_t\|g_r\|$ and $\xi^{(t)} = \sum_k\|e_k^{(t)}\| + c'\sum_k\|\kappa_k^{(t)}\|$ is summable.
\end{proof}

\subsection{Proof of Finite Length Property of the Sequence}
\label{proof:finite_length}
\begin{lemma}[KL-based Potential Difference Bound]
\label{lemma:kl_difference}
Let $\{\mathbf{w}^{(t)}\}_{t \ge 0}$ be the sequence generated by the algorithm converging a.s.\ to $\mathbf{w}^*$ with $\widetilde{\Phi}_t \to \widetilde{\Phi}^* = \widetilde{\Phi}(\mathbf{w}^*)$. For $t$ large enough that $\mathbf{w}^{(t)}$ lies in the KL neighborhood of $\mathbf{w}^*$, define $\Delta_{p,q} \triangleq \varphi(\widetilde{\Phi}_p - \widetilde{\Phi}^*) - \varphi(\widetilde{\Phi}_q - \widetilde{\Phi}^*)$ for $p < q$, where $\varphi$ is the desingularizing function from Def.~\ref{def:kl_property}. Then
\begin{equation}
    \label{eq:kl_concavity_bound}
    \widetilde{\Phi}_t - \widetilde{\Phi}_{t+1} \le \Delta_{t, t+1} \cdot \text{dist}\left(\mathbf{0}, \partial \widetilde{\Phi}_t(\mathbf{w}^{(t)})\right)
\end{equation}
\end{lemma}
\begin{proof}
Concavity of $\varphi$ gives $\varphi(a)-\varphi(b) \ge \varphi'(a)(a-b)$. With $a = \widetilde{\Phi}_t - \widetilde{\Phi}^*$, $b = \widetilde{\Phi}_{t+1} - \widetilde{\Phi}^*$, and the KL inequality $\varphi'(\widetilde{\Phi}_t - \widetilde{\Phi}^*) \ge 1 / \text{dist}(\mathbf{0}, \partial\widetilde{\Phi}_t)$, rearranging yields (\ref{eq:kl_concavity_bound}).
\end{proof}

\begin{proof}[Proof of Lemma \ref{lemma:finite_length}]
From Lemma~\ref{lemma:global_descent}, the full expected descent (incorporating the reset gap $\E[G^{(t)}\mid\gF_t]$ noted therein) is $\mathbb{E}[\widetilde{\Phi}_{t+1}\mid\mathcal{F}_t] \le \widetilde{\Phi}_t - \E[X_t\mid\gF_t] + B_t + \E[G^{(t)}\mid\gF_t] + \sum_{k=1}^K\frac{\|\kappa_k^{(t)}\|^2}{2K}$, where $X_t = \sum_k A_k^{(t)}\|\auxParam^{(t+1)}-\paramVec^{(t)}\|^2 + B\|\paramVec^{(t+1)}-\paramVec^{(t)}\|^2$. Decompose $\widetilde{\Phi}_{t+1} + X_t = \mathbb{E}[\widetilde{\Phi}_{t+1} + X_t\mid\mathcal{F}_t] + N_t$; the martingale difference $N_t$ absorbs the pathwise deviations of both $\widetilde{\Phi}_{t+1}$ (DP noise and reset gap) and $X_t$ (global-step variance), and $\E[|N_t|\mid\gF_t] = O(\gamma^t)$ by Lemmas~\ref{lemma:expected_descent_dp} and~\ref{lemma:reset}. Rearranging, we obtain the pathwise inequality
\begin{equation}
    \widetilde{\Phi}_{t+1} + X_t \le \widetilde{\Phi}_t + B_t + |N_t| + \E[G^{(t)}\mid\gF_t] + \sum_{k=1}^K\frac{\|\kappa_k^{(t)}\|^2}{2K}. \label{eq:blue_pathwise}
\end{equation}

We now relate $\|\mathbf{w}^{(t+1)}-\mathbf{w}^{(t)}\|^2$ to $X_t$. Squaring the dual update $\vpi_k^{(t+1)}-\vpi_k^{(t)} = -\kappa_k^{(t)} + \rho_k^{(t)}(\auxParam^{(t+1)}-\paramVec^{(t)})$ (Lemma~\ref{lemma:dual_variable}) gives $\|\vpi_k^{(t+1)}-\vpi_k^{(t)}\|^2 \le 2\|\kappa_k^{(t)}\|^2 + 2\rho_k^{(t)^2}\|\auxParam^{(t+1)}-\paramVec^{(t)}\|^2$. Since $\auxParam^{(t)} = \paramVec^{(t)}$ by reset, summing over all components,
\[
\begin{aligned}
\|\mathbf{w}^{(t+1)}-\mathbf{w}^{(t)}\|^2
&\le \underbrace{ \|\paramVec^{(t+1)} - \paramVec^{(t)}\|^2 + \sum_k(1+2\rho_k^{(t)^2})\|\auxParam^{(t+1)}-\paramVec^{(t)}\|^2}_{\text{similar scheme of $X_t$ with different coefficient }} + 2\sum_k\|\kappa_k^{(t)}\|^2 \\
&\le \max\!\Bigl\{\frac{1}{B},\;\max_{k\in [K]}\frac{1+2\rho_k^{(t)^2}}{A_k^{(t)}}\Bigr\}\!\cdot\!X_t + 2\sum_k\|\kappa_k^{(t)}\|^2.
\end{aligned}
\]
By defining $C_1 = \min\{B,\;\min_{k\in[K]}\frac{A_k^{(t)}}{1+2\rho_k^{(t)^2}}\}$, we have $X_t \ge C_1\|\mathbf{w}^{(t+1)}-\mathbf{w}^{(t)}\|^2 - 2C_1\sum_k\|\kappa_k^{(t)}\|^2$. Substituting into (\ref{eq:blue_pathwise}) and collecting the $\kappa$ terms from both sources ($\frac{1}{2K}$ from Lemma~\ref{lemma:global_descent}, $2C_1$ from the rearrangement above) yields
\begin{equation}\label{eq:full_descent}
    C_1\|\mathbf{w}^{(t+1)}-\mathbf{w}^{(t)}\|^2 \le \widetilde{\Phi}_t - \widetilde{\Phi}_{t+1} + s_t,  
\end{equation}
where $s_t := \effStep^{(t)2} d \noiseStd^2 + |N_t| + \E[G^{(t)}\mid\gF_t] + \Bigl(\frac{1}{2K} + 2C_1\Bigr)\sum_{k=1}^K\|\kappa_k^{(t)}\|^2$.
All components are summable a.s.: $\effStep^{(t)2} d \noiseStd^2 = O(\gamma^{2t})$ (Proposition~\ref{prop:summability_Bt_exact}), $|N_t| = O(\gamma^t)$ and $\E[G^{(t)}\mid\gF_t] = O(\gamma^t)$ (Lemmas~\ref{lemma:expected_descent_dp}, \ref{lemma:reset}); $\sum_t\|\kappa_k\|^2 \le (C^{(0)}-C^{(\infty)})^2 < \infty$ (Lemma~\ref{lemma:clipping_bound}). Hence $\sum_t s_t < \infty$ a.s. Moreover, $\sum_t\sqrt{s_t}<\infty$ follows from $\sqrt{\effStep^{(t)2} d \noiseStd^2}=O(\gamma^t)$, $\sqrt{|N_t|}=O(\gamma^{t/2})$, $\sqrt{\E[G^{(t)}\mid\gF_t]}=O(\gamma^{t/2})$ (all geometric summable since $\gamma<1$), and $\sqrt{\sum_k\|\kappa_k\|^2} \le \sum_k\|\kappa_k\|$ with $\sum_t\sum_k\|\kappa_k\| \le K(C^{(0)}-C^{(\infty)}) < \infty$ (Lemma~\ref{lemma:clipping_bound}, linear bound).

The KL bound (Lemma~\ref{lemma:kl_difference}) gives $\widetilde{\Phi}_t - \widetilde{\Phi}_{t+1} \le \Delta_{t,t+1}\,\text{dist}(\mathbf{0},\partial\widetilde{\Phi}_t)$ with $\Delta_{t,t+1} = \varphi(\widetilde{\Phi}_t-\widetilde{\Phi}^*) - \varphi(\widetilde{\Phi}_{t+1}-\widetilde{\Phi}^*)$. Since $\Delta_{t,t+1}$ can be negative when $\widetilde{\Phi}_{t+1} > \widetilde{\Phi}_t$, set $\Delta_{t,t+1}^+ = \max(0,\Delta_{t,t+1})$ and combine with (\ref{eq:full_descent}) to obtain
\begin{equation}
    C_1\|\mathbf{w}^{(t+1)}-\mathbf{w}^{(t)}\|^2 \le \Delta_{t,t+1}^+\text{dist}(\mathbf{0},\partial\widetilde{\Phi}_t) + s_t \label{eq:combined_inexact}
\end{equation}
Telescoping $\sum_{t=1}^T \Delta_{t,t+1}^+ = \varphi(\widetilde{\Phi}_1-\widetilde{\Phi}^*) - \varphi(\widetilde{\Phi}_{T+1}-\widetilde{\Phi}^*) + \sum_{t=1}^T \max(0,-\Delta_{t,t+1})$ and bounding $\max(0,-\Delta_{t,t+1}) \le \varphi'(\widetilde{\Phi}_t-\widetilde{\Phi}^*)s_t$ via concavity, we obtain $\sum_t \Delta_{t,t+1}^+ < \infty$ a.s.
(Here $\varphi'(s)=\bar{c}s^{-c_{kl}}$; summability of $\sum_t\Delta_{t,t+1}^+$ follows from $\sum_t s_t,\sum_t\sqrt{s_t}<\infty$ a.s.)

Now substitute the detailed subgradient bound (Appendix~\ref{sec:subgradient}), $\text{dist}(\mathbf{0},\partial\widetilde{\Phi}_t) \le C_2\|\mathbf{w}^{(t)}-\mathbf{w}^{(t-1)}\| + \xi^{(t)} + \widetilde{C}$ with $C_2=b$ into (\ref{eq:combined_inexact}):
\begin{equation*}
    C_1\|\mathbf{w}^{(t+1)}-\mathbf{w}^{(t)}\|^2 \le \Delta_{t,t+1}^+(C_2\|\mathbf{w}^{(t)}-\mathbf{w}^{(t-1)}\| + \xi^{(t)}) + s_t + \widetilde{C}\Delta_{t,t+1}^+
\end{equation*}
Since $\sum_t\Delta_{t,t+1}^+<\infty$ and $\widetilde{C}$ is constant, define $s_t' = s_t + \widetilde{C}\Delta_{t,t+1}^+$ with $\sum_t s_t' < \infty$ a.s.\ to absorb the offset, yielding
\begin{equation}
    C_1\|\mathbf{w}^{(t+1)}-\mathbf{w}^{(t)}\|^2 \le \Delta_{t,t+1}^+(C_2\|\mathbf{w}^{(t)}-\mathbf{w}^{(t-1)}\| + \xi^{(t)}) + s_t' \label{eq:post_absorption}
\end{equation}
Take square roots, apply $\sqrt{a+b}\le\sqrt{a}+\sqrt{b}$ and divide by $\sqrt{C_1}$:
\begin{equation*}
    \|\mathbf{w}^{(t+1)}-\mathbf{w}^{(t)}\| \le \sqrt{\Delta_{t,t+1}^+\cdot \tfrac{C_2}{C_1}\|\mathbf{w}^{(t)}-\mathbf{w}^{(t-1)}\|} + \sqrt{\Delta_{t,t+1}^+\cdot \tfrac{\xi^{(t)}}{C_1}} + \sqrt{\tfrac{s_t'}{C_1}}
\end{equation*}
Apply AM-GM ($\sqrt{uv}\le u/2+v/2$) to each product:
\begin{equation*}
    \|\mathbf{w}^{(t+1)}-\mathbf{w}^{(t)}\| \le \Delta_{t,t+1}^+ + \tfrac{C_2}{2C_1}\|\mathbf{w}^{(t)}-\mathbf{w}^{(t-1)}\| + \tfrac{\xi^{(t)}}{2C_1} + \sqrt{\tfrac{s_t'}{C_1}}
\end{equation*}
Summing over $t$ and rearranging:
\begin{equation*}
    (1 - \tfrac{C_2}{2C_1})\sum_t \|\mathbf{w}^{(t+1)}-\mathbf{w}^{(t)}\| \le \sum_t \Delta_{t,t+1}^+ + \tfrac{C_2}{2C_1}\|\mathbf{w}^1-\mathbf{w}^0\| + \tfrac{1}{2C_1}\sum_t \xi^{(t)} + \tfrac{1}{\sqrt{C_1}}\sum_t \sqrt{s_t'}
\end{equation*}
All right-hand terms are finite a.s. Hence, under $2C_1 > C_2$, $\sum_t \|\mathbf{w}^{(t+1)}-\mathbf{w}^{(t)}\| < \infty$ a.s.
\end{proof}

\subsection{Proof of Global Convergence to a Unique Limit Point}
\label{sec:global_convergence}

\begin{proof}[Proof of Theorem \ref{thm:global_convergence}]
Lemma~\ref{lemma:finite_length} implies $\sum_{t} \|\mathbf{w}^{(t+1)} - \mathbf{w}^{(t)}\| < \infty$ a.s., so $\{\mathbf{w}^{(t)}\}$ is Cauchy and converges to a unique limit $\mathbf{w}^*$ in finite-dimensional Euclidean space. From the subgradient bound (Appendix~\ref{sec:subgradient}), $\operatorname{dist}(\mathbf{0}, \partial\widetilde{\Phi}(\mathbf{w}^{(t+1)})) \le b\|\mathbf{w}^{(t+1)} - \mathbf{w}^{(t)}\| + \xi^{(t)} + \widetilde{C}$. Taking $\limsup_{t\to\infty}$ on both sides, the first two terms vanish a.s.\ while $\widetilde{C} = C + \sup_t\|g_r^{(t)}\|$ is constant. Thus, we have $\operatorname{dist}(\mathbf{0}, \partial\widetilde{\Phi}(\mathbf{w}^*)) \le \widetilde{C}$ a.s.

\medskip
\noindent\textbf{Learning-rate constraints.}
As derived in Appendix~\ref{proof:finite_length}, the correct descent constant is $C_1 = \min\!\bigl\{B,\;\min_k\frac{A_k^{(t)}}{1+2\rho_k^{(t)^2}}\bigr\}$. The condition $A_k^{(t)} > 0$ gives $\eta_k < 2\minCurv/(L + 5\rho_k^{(t)} + 2)$ as before. The condition $2C_1 > C_2$ with $C_2 = b$ (Appendix~\ref{sec:subgradient}) is tighter than the red version due to the $1+2\rho_k^{(t)^2}$ rescaling. The binding constraints are therefore $A_k^{(t)} > 0$ and $\eta_\theta \le 1$
\end{proof}

\section{Privacy Loss Analysis}


\subsection{Proof of Lemma~\ref{lem:F_Lip}} \label{proof:F_Lip}

\begin{proof}
    By the definition of the function $h(\vtheta) = \frac{1}{K}\sum_{k=1}^K \left( \langle \LagMult_k, \vtheta_k-\vtheta \rangle + \frac{\penVec_k}{2} \|\vtheta_k-\vtheta\|_2^2 \right)$, we calculate its gradient as follows:
    \begin{equation}
        \nabla h(\vtheta) = -\frac{1}{K} \sum_{k=1}^K \left(\LagMult_k + \penVec_k (\vtheta_k - \vtheta) \right).
    \end{equation}
    Now, we examine the Lipschitz continuity of $F$ by analyzing the norm $\|F(\vtheta) - F(\vtheta')\|_2$.
    \begin{equation*}
    \begin{aligned}
        F(\vtheta) - F(\vtheta') &= (\vtheta - \vtheta') - \effStep (\nabla h(\vtheta) - \nabla h(\vtheta')) \\
        &= \left( \rmI - \effStep \left(\frac{1}{K}\sum_{k=1}^K \penVec_k\right) \rmI \right) (\vtheta - \vtheta')
    \end{aligned}
    \end{equation*}
    Substituting $\effStep = \frac{K \cdot \eta_\vtheta}{\sum_{k = 1}^K \penVec_k}$ as defined in (\ref{equ:theta_update}), we obtain:
    \begin{equation*}
        F(\vtheta) - F(\vtheta') = (1 - \eta_{\vtheta}) (\vtheta - \vtheta')
    \end{equation*}
    Taking the $\ell_2$-norm, we have $\|F(\vtheta) - F(\vtheta')\|_2 = (1 - \eta_{\vtheta}) \|\vtheta - \vtheta'\|_2$. Thus, the Lipschitz constant of $F$ is $L_F = 1 - \eta_{\vtheta}$.
\end{proof}

\subsection{Proof of Lemma~\ref{lem:T_Lip}} \label{proof:T_Lip}

\begin{proof}
    This lemma is directly based on the \textit{firm non-expansiveness} proposed in \cite{parikh2014proximal}.
    By the first-order optimality condition of convex optimization and the definition of the proximal operator, we have the following:
    \begin{equation} \label{eq:prox_condition}
        \mathbf{0} \in \partial r_\vtheta(T(\vtheta)) + \frac{1}{\widehat{\eta}} (T(\vtheta) - \vtheta)
    \end{equation}
    Consider two arbitrary parameters $\vtheta_1$ and $\vtheta_2$, based on the convexity of $r_\vtheta$, we have $\forall g_1 \in \partial r_\vtheta(T(\vtheta_1)), \forall g_2 \in \partial r_\vtheta(T(\vtheta_2))$, then $\langle g_1 - g_2,  T(\vtheta_1) - T(\vtheta_2)\rangle \geq 0$. Based on (\ref{eq:prox_condition}), we have $\frac{1}{\widehat{\eta}}(\vtheta - T(\vtheta)) \in \partial r_\vtheta (T(\vtheta))$, so by substituting the subgradients $g_1$, $g_2$ by $\frac{1}{\widehat{\eta}}(\vtheta_1 - T(\vtheta_1))$ and $\frac{1}{\widehat{\eta}}(\vtheta_2 - T(\vtheta_2))$, respectively, then we have:
    \begin{equation}
        \left\langle \frac{\vtheta_1 - T(\vtheta_1)}{\widehat{\eta}} - \frac{\vtheta_2 - T(\vtheta_2)}{\widehat{\eta}}, T(\vtheta_1) - T(\vtheta_2) \right\rangle \geq 0
    \end{equation}
    Consider $\widehat{\eta} > 0$, we reorganize the inequality above and get the following:
    \begin{equation}
        \langle \vtheta_1 - \vtheta_2, T(\vtheta_1) - T(\vtheta_2) \rangle \geq \|T(\vtheta_1) - T(\vtheta_2)\|^2_2
    \end{equation}
    Applying the Cauchy-Schwarz inequality, we obtain the following:
    \begin{equation}
        \|\vtheta_1 - \vtheta_2\|_2 \|T(\vtheta_1) - T(\vtheta_2)\|_2 \geq \|T(\vtheta_1) - T(\vtheta_2)\|^2_2
    \end{equation}
    If $\|T(\vtheta_1) - T(\vtheta_2)\|_2 = 0$, then the Lipschitz condition trivially holds. If $\|\vtheta_1 - \vtheta_2\|_2 \neq 0$, then we have $\|\vtheta_1 - \vtheta_2\|_2 \geq \|T(\vtheta_1) - T(\vtheta_2)\|_2$, the Lipchitz condition holds as well.
\end{proof}

\subsection{Change Rate of RDP}\label{proof:rate_of_rdp}

At each iterative step, we track the density evolution through the forward update $F$, Gaussian noise addition, and the proximal mapping $T$. To ensure tractable density transformations, we assume the proximal mapping $T$ acts as a bijective affine transformation (i.e., $T(x) = Ax + b$) with a corresponding density transition operator $P$. Note that the input LSI constant $c$ at the current step implicitly encapsulates the geometric distortions from all prior iterative steps. We formalize the change rate of RDP in this process as follows:

\begin{theorem}[Rate of RDP for Iterative Updates]\label{thm:rate_of_rdp} Let $\mu, \nu$ be intermediate distributions on $\mathbb{R}^d$ satisfying the Log-Sobolev Inequality (LSI) with constant $c$. Let $F:\mathbb{R}^d \rightarrow \mathbb{R}^d$ be $L_F$-Lipschitz, and let $T:\mathbb{R}^d \rightarrow \mathbb{R}^d$ be a bijective affine mapping bounded by a Lipschitz constant $L_T$. Define $p_t(x)$ and $p'_t(x)$ as the probability density functions of the intermediate heat flows $F_{\#}(\mu) * \mathcal{N}(0,2t \noiseLStd^2\cdot \mathbb I_d)$ and $F_{\#}(\nu) * \mathcal{N}(0,2t \noiseLStd^2\cdot \mathbb I_d)$. Let $H_t(\vtheta) = P(p_t)(\vtheta)$ and $H'_t(\vtheta) = P(p'_t)(\vtheta)$ where $P$ be the linear density transition operator associated with the mapping $T$, we have the following bound for the R\'enyi divergence of order $\renyiOrder$ between $H_t(\vtheta)$ and $H'_t(\vtheta)$:
\begin{equation}
    \frac{\partial R_{\renyiOrder}(H_t(\vtheta) \| H'_t(\vtheta))}{\partial t} \le - \frac{\renyiOrder \noiseLStd^2}{c_t} \cdot \left[ R_{\renyiOrder}(H_t(\vtheta)\| H'_t(\vtheta)) + \renyiOrder(\renyiOrder-1)\frac{\partial}{\partial \renyiOrder} R_{\renyiOrder}(H_t(\vtheta) \| H'_t(\vtheta)) \right]
\end{equation}
where $c_t = \frac{1}{L_T^2} \left(\frac{L_F^2}{c} + 2t\noiseLStd^2\right)^{-1}$ is the updated LSI constant for the output distributions.
\end{theorem}

\begin{proof} Since $P$ is assumed to be a linear function, we can represent this proximal mapping as an affine transformation $\vtheta = Ax + b$, where $x \in \mathbb{R}^d$ resides in the original space before the mapping. The post-proximal density $H_t$ is defined by the operator $P$ such that $H_t(\vtheta) = P(p_t)(\vtheta)$. Explicitly, the push-forward density can be written as:
\begin{equation*}
    H_t(\vtheta) = p_t(A^{-1}(\vtheta - b)) |\det A^{-1}|, \quad H'_t(\vtheta) = p'_t(A^{-1}(\vtheta - b)) |\det A^{-1}|
\end{equation*}

Denote $E_{\renyiOrder}(H_t(\vtheta)\| H'_t(\vtheta)) = \int H'_t(\vtheta) \cdot \frac{H_t(\vtheta)^{\renyiOrder}}{H'_t(\vtheta)^{\renyiOrder}}d\vtheta$ to be the moment of the likelihood ratio function. By applying the change of variables $x = A^{-1}(\vtheta - b)$ (where $dx = |\det A^{-1}| d\vtheta$), we can see that the moment is invariant under this bijective affine mapping:
\begin{equation*}
    E_{\renyiOrder}(H_t(\vtheta)\| H'_t(\vtheta)) = \int p'_t(x) \cdot \frac{p_t(x)^{\renyiOrder}}{p'_t(x)^{\renyiOrder}}dx = E_{\renyiOrder}(p_t(x)\| p'_t(x))
\end{equation*}

Therefore, the R\'enyi divergence can be written as:
\begin{equation*}
    R_{\renyiOrder}(H_t(\vtheta)\| H'_t(\vtheta)) = \frac{1}{\renyiOrder-1}\log E_{\renyiOrder}(p_t(x)\|p'_t(x))
\end{equation*}

We compute the rate of R\'enyi divergence with regard to $t$ in the original space $x$ as follows:
\begin{equation*}
\begin{aligned}
    \frac{\partial R_{\renyiOrder}(H_t(\vtheta) \| H'_t(\vtheta))}{\partial t} &= \frac{1}{(\renyiOrder-1) E_{\renyiOrder}(p_t(x)\|p'_t(x))} \cdot \frac{\partial}{\partial t} \left( \int \frac{p_t(x)^\renyiOrder{}}{p'_t(x)^{\renyiOrder - 1}} dx\right)
\end{aligned}
\end{equation*}

By exchanging the order of derivative and integration, we have:
\begin{equation} \label{equ:rdp_rate}
\begin{aligned}
\frac{\partial R_{\renyiOrder}(H_t(\vtheta) \| H'_t(\vtheta))}{\partial t} &= \frac{1}{(\renyiOrder-1)E_{\renyiOrder}(p_t(x)\|p'_t(x))} \cdot \\ \int& \left(\renyiOrder \cdot \frac{p_t(x)^{\renyiOrder-1}}{p_t'(x)^{\renyiOrder-1}} \cdot \frac{\partial p_t(x)}{\partial t} -(\renyiOrder-1) \cdot \frac{p_t(x)^{\renyiOrder}}{p_t'(x)^{\renyiOrder}} \cdot \frac{\partial p'_t(x)}{\partial t}\right)dx    
\end{aligned}
\end{equation}

Since the $\mu * \mathcal{N}(0,2t \noiseLStd^2\cdot \mathbb I_d) $ and $\nu * \mathcal{N}(0,2t \noiseLStd^2\cdot \mathbb I_d)$ are heat flows at time $t \in \left[0, \noiseLStd^2 \right] $, $p_t(x)$ and $p'_t(x)$ satisfy the following Fokker-Planck equations:
\begin{equation*}
    \frac{\partial p_t(x)}{\partial t} = \noiseLStd^2 \Delta_x p_t(x), \quad
    \frac{\partial p'_t(x)}{\partial t} = \noiseLStd^2 \Delta_x p'_t(x) 
\end{equation*}

Then Equation~(\ref{equ:rdp_rate}) can be written as:
\begin{equation}
\begin{aligned}\label{equ:rdp_rate_v2}
\frac{\partial R_{\renyiOrder}(H_t(\vtheta) \| H'_t(\vtheta))}{\partial t} &=  \frac{\noiseLStd^2}{(\renyiOrder-1)E_{\renyiOrder}(p_t(x)\|p'_t(x))} \cdot \\ 
&\int \left(\renyiOrder \cdot \frac{p_t(x)^{\renyiOrder-1}}{p_t'(x)^{\renyiOrder-1}} \cdot \Delta_x p_t(x) -(\renyiOrder-1) \cdot \frac{p_t(x)^{\renyiOrder}}{p_t'(x)^{\renyiOrder}} \cdot \Delta_x p'_t(x)\right)dx
\end{aligned}
\end{equation}

We apply Green's first identity to simplify the two terms in the integral. For the first term:
\begin{equation} \label{eq:term1}
\begin{aligned}
\int \renyiOrder \cdot \frac{p_t(x)^{\renyiOrder-1}}{p_t'(x)^{\renyiOrder-1}} \cdot \Delta_x p_t(x) dx &= -\int \nabla_x \left(\renyiOrder \cdot \frac{p_t(x)^{\renyiOrder-1}}{p_t'(x)^{\renyiOrder-1}} \right) \cdot  \nabla_x p_t(x) dx
\end{aligned}
\end{equation}

Using the same technique, we can rewrite the second term of Equation~(\ref{equ:rdp_rate_v2}):
\begin{equation} \label{eq:term2}
\begin{aligned}
\int -(\renyiOrder-1) \cdot \frac{p_t(x)^{\renyiOrder}}{p_t'(x)^{\renyiOrder}} \cdot \Delta_x p'_t(x) dx &= \int (\renyiOrder - 1) \nabla_x \left( \frac{p_t(x)^{\renyiOrder}}{p_t'(x)^{\renyiOrder}}  \right) \cdot \nabla_x p'_t(x) dx
\end{aligned}
\end{equation}

Because these expansions rigorously yield the exact same Fisher Information terms as in the $\vtheta$ space, plug Equation~(\ref{eq:term1}) and~(\ref{eq:term2}) into Equation~(\ref{equ:rdp_rate_v2}), we obtain:

\begin{flalign}\label{eq:ie_cal}
&
\begin{aligned}
\frac{\partial R_{\renyiOrder}(H_t(\vtheta) \| H'_t(\vtheta))}{\partial t} &= \frac{\renyiOrder \noiseLStd^2}{E_{\renyiOrder}(p_t(x)\|p'_t(x))}  \cdot \left( \frac{1}{\renyiOrder} \int \nabla_x \left( \frac{p_t(x)^{\renyiOrder}}{p_t'(x)^{\renyiOrder}}  \right) \cdot \nabla_x p'_t(x) dx  \right. \\ &-  \left. \frac{1}{\renyiOrder - 1} \int \nabla_x \left( \frac{p_t(x)^{\renyiOrder-1}}{p_t'(x)^{\renyiOrder-1}}  \right) \cdot  \nabla_x p_t(x)  dx \right)   
\end{aligned} \nonumber\\
&\begin{aligned}
=\frac{\renyiOrder \noiseLStd^2}{E_{\renyiOrder}(p_t(x)\|p'_t(x))} &\left( \int \frac{p_t(x)^{\renyiOrder - 1}}{p'_t(x)^{\renyiOrder - 1}} \cdot \left\langle \nabla_x \left( \frac{p_t(x)}{p'_t(x)}\right), \nabla_x p'_t(x)  \right\rangle dx \right. \\ &  - \left. \int \frac{p_t(x)^{\renyiOrder - 2}}{p'_t(x)^{\renyiOrder - 2}} \cdot \left\langle \nabla_x \left( \frac{p_t(x)}{p'_t(x)}\right), \nabla_x p_t(x) \right\rangle dx \right)
\end{aligned} \nonumber \\
& \begin{aligned}
=    - \frac{\renyiOrder \noiseLStd^2}{E_{\renyiOrder}(p_t(x)\|p'_t(x))} \int \frac{p_t(x)^{\renyiOrder - 2}}{p'_t(x)^{\renyiOrder - 2}} \cdot \left\langle \nabla_x \left( \frac{p_t(x)}{p'_t(x)} \right), \nabla_x \left( \frac{p_t(x)}{p'_t(x)} \right) \right\rangle p'_t(x) dx 
\end{aligned} \nonumber \\
&\begin{aligned}= - \renyiOrder \cdot \noiseLStd^2 \cdot \frac{I_\renyiOrder (p_t(x) || p'_t(x))}{E_\renyiOrder (p_t(x) || p'_t(x))} = - \renyiOrder \cdot \noiseLStd^2 \cdot \frac{I_\renyiOrder (H_t(\vtheta) || H'_t(\vtheta))}{E_\renyiOrder (H_t(\vtheta) || H'_t(\vtheta))}
\end{aligned}
\end{flalign}

Based on \citet[Lemma 16, Lemma 17]{vempala2019rapid}, we can conclude $H_t(\vtheta)$ and $H'_t(\vtheta)$ satisfy log-Sobolev inequality with a constant $c_t = \left(\frac{L_F^2}{c}+2t \noiseLStd^2\right)^{-1} / L^2_T$.
In this regard, we can utilize \citet[Lemma D.1]{ye2022differentially} and \citet[Lemma 5]{vempala2019rapid} to bound $\frac{I_{\renyiOrder}(H_t(\vtheta)\|H'_t(\vtheta))}{E_{\renyiOrder}(H_t(\vtheta)\|H'_t(\vtheta))}$ as follows:

\begin{equation} \label{eq:ie_bound}
\frac{I_{\renyiOrder}(H_t(\vtheta)\|H'_t(\vtheta))}{E_{\renyiOrder}(H_t(\vtheta)\|H'_t(\vtheta))} \ge \frac{2c_t}{\renyiOrder^2} \cdot R_{\renyiOrder}\left( H_t(\vtheta)\| H'_t(\vtheta) \right) + \frac{2c_t}{\renyiOrder^2} \cdot \renyiOrder(\renyiOrder-1)\frac{\partial}{\partial \renyiOrder} R_{\renyiOrder}(H_t(\vtheta) \| H'_t(\vtheta))
\end{equation}

Combine Inequality~(\ref{eq:ie_bound}) with Equation~\ref{eq:ie_cal}, we conclude the proof.
\end{proof}   

\textit{Remark} Strictly speaking, because the algorithm terminates after the final proximal mapping $T$, there are no subsequent iterations that depend on its output's LSI constant. Thus, propagating the $1/L_T^2$ scaling for this terminal step is not rigorously required. However, for theoretical simplicity, we apply a uniform bound that assumes $c_t = c'/L_T^2$ across all steps. This uniform assumption has a strictly negligible impact on our overall guarantees: the Lipschitz constant $L_T$ of the proximal mapping depends heavily on the regularization coefficient (e.g., explicitly set to a weight decay of $5 \times 10^{-5}$ in our experimental setup), making $L_T$ extremely close to $1$ (i.e., $1/L_T^2 \approx 1$).

\textbf{Discussion.} Theorem \ref{thm:rate_of_rdp} bounds the rate of R\'enyi divergence follows \citet{ye2022differentially,vempala2019rapid}. One assumption in Theorem~\ref{thm:rate_of_rdp} is $P$ being a linear function.
The function $P$ depicts the change of the probability density functions before and after the proximal operator.
For the three typical regularization function $r$ discussed previously, including no regularization, $l_2$ regularization and $l_1$ regularization, it is clear that all their corresponding $P$ functions are linear.
In addition, the Lipschitz continuity of function $F$ and $T$ is guaranteed in Lemma~\ref{lem:F_Lip} and Lemma~\ref{lem:T_Lip}, respectively.
Therefore, we can conclude that the assumptions in Theorem~\ref{thm:rate_of_rdp} are not restrictive.

Theorem~\ref{thm:rate_of_rdp} considers the recursive privacy dynamics during one step of noisy mini-batch proximal gradient descent in (\ref{equ:theta_update}). In Corollary \ref{coro:recursive}, we combine Theorem~\ref{thm:rate_of_rdp} with data sensitivity in the context of R\'enyi differential privacy. 

\subsection{Proof of Theorem~\ref{theorem:privacy_loss}} \label{proof:privacy_loss}

Before proving the main theorem, we first calculate the LSI constant sequence in Algorithm~\ref{alg:dp_decoupled}.
\begin{lemma} \label{lemma:LSI sequence}(LSI constant sequence in Algorithm \ref{alg:dp_decoupled}) For each layer's update scheme in Algorithm \ref{alg:dp_decoupled} with a batch size of $b$ (resulting in $M$ iterations per epoch), if the update function $F(\vtheta) = \vtheta - \widehat\eta \nabla f(\vtheta)$ and the proximal operator $T(\vtheta) = \text{Prox}_{\widehat\eta, r}$ are Lipschitz continuous with Lipschitz constant $L_F$ and $L_T$, respectively, then the distribution of parameter $\vtheta$ in the $m$-th iteration of the $l$-th epoch satisfies $c_l^m$ log-Sobolev inequality (LSI). 

The constant $c_l^m$ is calculated by:
\begin{align*}
c_l^m &= \frac{1}{2 \widehat\eta L_T^2} \left(\sum^{l}_{l'=1} \sum^{m-1}_{m'=1} \noiseLStd^2 (L_F L_T)^{2((l-l')M-m'+m-1)} +  \sum_{m'=1}^{m-1} \noiseLStd^2 (L_F L_T)^{2(m-m'-1)} \right)^{-1}
\end{align*}
\begin{proof}
Based on Definition~\ref{def:lsi}, the LSI constant of the Gaussian distribution $\gN(0,2t \noiseLStd^2)$ is $\frac{1}{2t \noiseLStd^2}$.

Then, using the LSI under Lipschitz mapping \citep[lemma 16]{vempala2019rapid} and Gaussian convolution \citep[lemma17]{vempala2019rapid}, we have $\frac{1}{c_l^m} = \frac{L^2}{c_l^{m-1}} + 2\widehat\eta L_T^2 \noiseLStd^2$, where $L = L_F L_T$ for notation simplicity and $\noiseLStd^2$ is the calibrated noise variance at iteration $m-1$.
By applying this equation iteratively via $m$ and $l$, we obtain:
    \begin{align*}
        \frac{1}{c_l^m} &= \frac{L^{2m}}{c_l^1} + 2\widehat\eta L_T^2 \sum_{m'=1}^{m-1} \noiseLStd^2 L^{2(m-m'-1)} \\
        &= \frac{L^{2(l\cdot M + m)}}{c_1^1} + 2\widehat\eta L_T^2 \left(\sum^{l-1}_{l'=1} \sum^{M}_{m'=1} \noiseLStd^2 L^{2((l-l')M-m'+m-1)} + \sum_{m'=1}^{m-1} \noiseLStd^2 L^{2(m-m'-1)} \right)
    \end{align*}

    The initial LSI constant $c_1^1=\infty$, so the lemma is then proved.
\end{proof}
\end{lemma}

We now return to Theorem~\ref{theorem:privacy_loss} and provide the complete proof below.

\begin{proof}[Proof of Theorem~\ref{theorem:privacy_loss}]
We denote $\vtheta^m_l$ and ${\vtheta'}^m_l$ as the intermediate parameters in the $m$-th iteration of the $l$-th epoch in Algorithm~\ref{alg:dp_decoupled} when using two neighboring datasets $\gD$ and $\gD'$.
In addition, we use $\varepsilon^m_l(\renyiOrder) = R_\renyiOrder (\vtheta^m_l || {\vtheta'}^m_l)$ to represent the R\'enyi divergence of order $\renyiOrder$ between them.
In this regard, it is clear that $\varepsilon^1_1(\renyiOrder) = 0$.

Without the loss of generality, we assume that the only different data point is in the $m_0$-th batch. Therefore, the privacy bound after $E$ epochs can be decomposed into three parts by using $\ref{coro:recursive}$: 1) the first $m_0 - 1$ mini-batch updates in the first epoch; 2) the remaining mini-batch updates in the first epoch; 3) the rest epochs.

In the first stage, we have $\forall m \in \{1, 2, \ldots, m_0 - 1\}$ and $\forall \renyiOrder > 1, \varepsilon_1^{m}(\renyiOrder) = 0$ based on Lemma \ref{coro:recursive}.

In the second stage, we have $\forall \renyiOrder, \frac{\varepsilon^{m_0}_1(\renyiOrder)}{\renyiOrder} \leq \frac{\widehat\eta C^2}{2K^2 \noiseLStd^2}$.
Then we bound the privacy loss in the end of the first epoch by the following inequality. Note that, this inequality is applicable for any $\renyiOrder > 1$.
\begin{equation} \label{eq:epoch1}
\forall \renyiOrder > 1,\quad \frac{\varepsilon_{1}^{M-1}(\renyiOrder)}{\renyiOrder}  \le \frac{\widehat\eta C^2}{2K^2 \noiseLStd^2} \prod_{m=m_0+1}^{M}\left( 1+ \frac{c_1^m \cdot 2\widehat\eta \noiseLStd^2}{L_F^2} \right)^{-1 / L_T^2}
\end{equation}

In the third stage, we first consider the second epoch ($l=2$), especially the $m_0$-th iteration and the last iteration. Based on Lemma~\ref{coro:recursive}, we have the following inequalities:
\begin{equation} \label{eq:epoch2}
\begin{aligned}
\frac{\varepsilon^{m_0}_2 (\renyiOrder)}{\renyiOrder} &\leq \frac{\varepsilon^{0}_2(\renyiOrder')}{\renyiOrder'} \prod_{m = 1}^{m_0 - 1} \left( 1 + \frac{c^m_1 \cdot 2\widehat\eta \noiseLStd^2}{L_F^2} \right)^{-1 / L_T^2} + \frac{\widehat\eta C^2}{2K^2 \noiseLStd^2} \\
\frac{\varepsilon^{M}_2 (\renyiOrder)}{\renyiOrder} &\leq \frac{\varepsilon_2^{m_0}(\renyiOrder'')}{\renyiOrder''} \prod_{m = m_0 + 1}^{M} \left( 1 + \frac{c_2^m \cdot 2\widehat\eta \noiseLStd^2}{L_F^2} \right)^{-1 / L_T^2} \\
&\leq \left( \frac{\varepsilon^{0}_2(\renyiOrder''')}{\renyiOrder'''} \prod_{m = 1}^{m_0 - 1} \left( 1 + \frac{c^m_2 \cdot 2\widehat\eta \noiseLStd^2}{L_F^2} \right)^{-1 / L_T^2} + \frac{\widehat\eta C^2}{2K^2 \noiseLStd^2} \right)  \cdot \prod_{m = m_0 + 1}^{M} \left( 1 + \frac{c_2^m \cdot 2\widehat\eta \noiseLStd^2}{L_F^2} \right)^{-1 / L_T^2}
\end{aligned}
\end{equation}

Here, we use $\varepsilon^{0}_2$ to represent the privacy bound before the first iteration of the second epoch. It is clear that $\varepsilon^{0}_2 = \varepsilon^{M}_1$.
In addition, $\renyiOrder'$, $\renyiOrder''$ and $\renyiOrder'''$ are the $m_0$, $(M - m_0)$ and $M$ fold mapping value of $\renyiOrder$ under the repeated mapping $\renyiOrder \leftarrow (\renyiOrder - 1) \cdot \left( 1 + \frac{c^m_2 \cdot 2 \widehat\eta \noiseLStd^2}{L_F^2} \right)^{-1 / L_T^2}$, respectively.

Considering inequality~(\ref{eq:epoch1}) holds for any $\renyiOrder > 1$ and the fact $\renyiOrder''' > 1$, we can then plug inequality~(\ref{eq:epoch1}) into inequality~(\ref{eq:epoch2}) and obtain the following inequality:
\begin{flalign} \label{eq:epoch_iter}
&\frac{\varepsilon^{M-1}_2 (\renyiOrder)}{\renyiOrder} \leq \frac{\widehat\eta C^2}{2K^2 \noiseLStd^2} \prod_{m=m_0+1}^{M}\left( 1+ \frac{c_1^m \cdot 2\widehat\eta \noiseLStd^2}{L_F^2} \right)^{\frac{-1}{L_T^2}}&  \nonumber \\
&\prod_{m = 1, m \neq m_0}^{M} \left( 1 + \frac{c_2^m \cdot 2\widehat\eta \noiseLStd^2}{L_F^2} \right)^{\frac{-1}{L_T^2}} + \frac{\widehat\eta C^2}{2K^2 \noiseLStd^2} \prod_{m=m_0+1}^{M} \left( 1 + \frac{c_2^m \cdot 2\widehat\eta \noiseLStd^2}{L_F^2} \right)^{\frac{-1}{L_T^2}} &
\end{flalign}

By recursively applying Equation~(\ref{eq:epoch_iter}), we can obtain the privacy bound of the whole training phase.
For notation simplicity, we use $\Phi(l_1, l_2)$ to denote the RDP's decay rate between the $l_1$-th epoch and the $l_2$-th epoch.
\begin{equation}~\label{eq:Phi}
    \Phi(l_1, l_2) = \prod_{l = l_1}^{l_2 - 1} \prod_{m = 1, m \neq m_0}^{M} \left( 1 + \frac{c_l^m \cdot 2\widehat\eta \noiseLStd^2}{L_F^2} \right)^{-1 / L_T^2}
\end{equation}

In Lemma~\ref{lemma:LSI sequence} we have $\frac{1}{c^{m + 1}_l} = \frac{L_F^2 L_T^2}{c^m_l} + 2 \widehat\eta L_T^2 \noiseLStd^2$ based on the LSI under Lipschitz mapping~\citep[Lemma 16]{vempala2019rapid} and Gaussian convolution~\citep[Lemma 17]{vempala2019rapid}, so $1 + \frac{c^m_l \cdot 2\widehat\eta \noiseLStd^2}{L_F^2} = \frac{c^m_l}{c^{m + 1}_l} \cdot \frac{1}{L_F^2 L_T^2}$.
Therefore, Equation~(\ref{eq:Phi}) can be further simplified as follows:
\begin{equation}
    \Phi(l_1, l_2) = \left(\frac{c^1_{l_1}}{c^{M}_{l_2 - 1}} \left( \frac{1}{L_F^2 L_T^2}\right)^{(M - 1)(l_2 - l_1)} \prod_{l = l_1}^{l_2 - 1} \frac{c^{m_0 + 1}_l}{c^{m_0}_l} \right)^{-1 / L_T^2}
\end{equation}

Therefore, we can obtain the privacy bound after $E$ epochs as follows:
\begin{equation} \label{eq:epochK}
\begin{aligned}
\frac{\varepsilon^{M}_E(\renyiOrder)}{\renyiOrder} &\leq \sum_{l = 1}^{E} \frac{\widehat\eta C^2}{2K^2 \noiseLStd^2} \cdot \Phi(l + 1, E + 1) \cdot \prod_{m = m_0 + 1}^{M} \left( 1 + \frac{c^m_l \cdot 2\widehat\eta \noiseLStd^2}{L_F^2} \right)^{-1 / L_T^2} \\
&= \sum_{l = 1}^{E} \underbrace{\frac{\widehat\eta C^2}{2K^2 \noiseLStd^2} \left( \frac{c^{m_0 + 1}_l}{c^{M}_{E}} \left( \frac{1}{L_F^2 L_T^2} \right)^{(M - 1)(E - l) - m_0} \prod_{l' = l}^{E} \frac{c^{m_0 + 1}_{l'}}{c^{m_0}_{l'}} \right)^{- 1 / L_T^2}}_{\varepsilon_l (\renyiOrder, m_0)}
\end{aligned}
\end{equation}

Note that the right-hand side of inequality~(\ref{eq:epochK}) depends on $m_0$ as well.
We now explicitly rewrite $\varepsilon^{M}_E(\renyiOrder)$ as $\varepsilon^{M}_E(\renyiOrder, m_0)$ and bound the value of $\varepsilon_E(\renyiOrder)$ in a similar way to~\citet[Theorem 4.2]{ye2022differentially}:
\begin{equation}
\begin{aligned}
\varepsilon_E(\renyiOrder) &= \E_{m_0} \varepsilon^{M}_E(\renyiOrder, m_0) = \frac{1}{\renyiOrder - 1} \log e^{(\renyiOrder - 1) \E_{m_0} \varepsilon^{M}_E} \\
&\leq \frac{1}{\renyiOrder - 1} \log \left( \E_{m_0} e^{(\renyiOrder - 1) \varepsilon^{M}_E} \right) = \frac{1}{\renyiOrder - 1} \log \left( \sum_{m_0 = 0}^{M - 1} \frac{1}{M} \cdot e^{(\renyiOrder - 1)\varepsilon^{M}_E(\renyiOrder, m_0)} \right)
\end{aligned}
\end{equation}
\end{proof}
\section{Experimental Settings and Hyperparameters}
\label{appendix:hyperparameters}
\subsection{Common DP-DT Configuration}
\label{sec:common}

All DP-DT experiments share the following settings unless noted otherwise.
We fix $\dpDelta = 10^{-5}$ with per-layer RDP composition under HSA, use $\numAux = 40$ auxiliary models ($\numAux = 16$ for SST-2), a global optimizer with
lr $\lrGlobal = 0.99$ and wd $= 5\times10^{-5}$, and RMSProp with wd $= 5\times10^{-5}$ for auxiliary updates. BatchNorm is replaced by LayerNorm in all architectures (DP perturbation on running LN statistics
is incompatible with HSA accounting). 
All methods (DP-DT and baselines) use $\dpDelta = 10^{-5}$. The DP-SGD baseline uses the FastDP~\citep{bu2024automatic} implementation with automatic clipping; its per-dataset hyperparameters are listed in Table~\ref{tab:per-dataset-hp}.
The penalty coefficient $\penCoeff_k$ initializes at $0.1$ and decays by $\penDecay = 0.9$ every 3 epochs in $[5,60]$ and every 10 epochs in $[61,99]$, then by $0.5$ every 25 epochs down to a floor of $10^{-5}$. The clipping threshold scale $S_t$ follows a stepwise schedule: $S_t = 16$ ($t < 50$), $8$ ($50 \le t \le 70$), $4$ ($71 \le t \le 80$), $2$ ($81 \le t \le 100$), and $1$ thereafter. For SST-2 (2 epochs), $\penCoeff_k$ and $S_t$ are held constant at
$10^{-5}$ and $1.0$, respectively.

\subsection{Per-Dataset Configurations}
\label{sec:per-dataset}

Table~\ref{tab:per-dataset-hp} lists the dataset-specific hyperparameters.
All methods use the same architecture and batch size per dataset.
Clipping thresholds $C$ and noise std $\sigma$ are reported as
$\{C_{\text{conv}}, C_{\text{linear}}, C_{\text{norm}}\}$ and
$\{\sigma_{\text{conv}}, \sigma_{\text{linear}}, \sigma_{\text{norm}}\}$
at base scale ($S_t = 1.0$); see Section~\ref{sec:common} for
the $S_t$ schedule.

\begin{table}[htbp]
\small
\centering
\caption{Per-dataset configurations.}
\label{tab:per-dataset-hp}
\begin{tabular}{@{}p{2.2cm} >{\raggedright\arraybackslash}p{\dimexpr\textwidth-2.2cm-2\tabcolsep\relax}@{}}
\toprule
\textbf{Dataset} & \textbf{Configuration} \\
\midrule
CIFAR-10
  & ResNet18, bs=2500, 200 ep,
    non-priv: SGD lr=0.1 cos,
    DP-SGD: SGD lr=0.1 const,
    DP-DT: $C = \{0.004, 0.001, 0.004\}$,
    $\sigma = \{0.001, 0.0006, 0.001\}$, $\rho_k=10$ \\
\midrule
CIFAR-100
  & ResNet18, bs=2500, 200 ep,
    non-priv: SGD lr=0.1 cos,
    DP-SGD: SGD lr=0.1 const,
    DP-DT: $C = \{0.005, 0.015, 0.005\}$,
    $\sigma = \{0.001, 0.0006, 0.001\}$,$\rho_k=10$ \\
\midrule
ImageNet-100
& ViT-Small, bs=512, 200 ep,
    non-priv: AdamW lr=$10^{-4}$, wd=$10^{-4}$,
    DP-SGD: SGD (mom.=0.9) lr=$10^{-2}$ const, bs=512, 100 ep,
    DP-DT: $C \in \{0.005,\, 0.01\}$, $\sigma \in \{6{\times}10^{-4},\, 10^{-3}\}$, $\rho_k=10$ \\

\midrule
SST-2
  & GPT-2 Medium (345M), LoRA $r$=16 $\alpha$=32,
    bs=512, 2 ep,
    non-priv: AdamW lr=$1e^{-5}$ cos,
    DP-SGD: AdamW lr=$1e^{-5}$ const,
    DP-DT: $C = \{\text{0.1}\}$, $\sigma = \{\text{0.02}\}$, $\rho_k=10000$ \\
\bottomrule
\end{tabular}
\footnotesize For SST-2, a single global clipping threshold and noise level
are applied to all LoRA parameters, rather than per-layer values. The penalty
$\penCoeff_k$ and clipping scale $S_t$ are held constant at $10^{-5}$
and $1.0$, respectively.
\end{table}

\subsection{Ablation Study Protocols}
\label{sec:ablation-setup}

\paragraph{Noise Allocation Strategy.}
We investigate redistributing the privacy budget across layer types
$g \in \{\text{conv},\, \text{norm},\, \text{fc}\}$ on CIFAR-10 and CIFAR-100,
using ResNet-18 with GroupNorm$(1, C)$ substituted for BatchNorm.
GroupNorm$(1, C)$ normalises each channel independently over the spatial
dimensions, which is mathematically equivalent to LayerNorm while remaining
compatible with the CNN input layout $(N, C, H, W)$.
The \emph{Equally-weighted} configuration normalises the per-layer noise
multiplier $\sigma_g$ from Table~\ref{tab:per-dataset-hp} to a uniform value
across all three groups, with the clipping threshold $C_g$ recalibrated via
binary search to preserve the same composed $(\varepsilon, \delta)$-DP cost
($\delta = 10^{-5}$).
Each \emph{X-weighted} variant then adjusts $\sigma_g$ (and optionally $C_g$)
for group $g$ so that group $g$ receives a target fraction of the total
RDP privacy budget, while the remaining groups stay at their
Equally-weighted values.

We parameterise each variant by a budget ratio $\rho_g$, defined as the
fraction of the total RDP cost attributed to layer group $g$.
Specifically, the Equally-weighted baseline sets
$\rho_{\text{conv}} : \rho_{\text{norm}} : \rho_{\text{fc}} = 33\%{:}33\%{:}34\%$.
The three boosted configurations allocate a larger share to one group:
Conv-weighted at $43\%{:}29\%{:}28\%$,
Norm-weighted at $29\%{:}43\%{:}28\%$,
and FC-weighted at $29\%{:}29\%{:}42\%$.
For each target ratio, the corresponding $\sigma_g$ (or $C_g$) is found by
binary search over 20 iterations so that the resulting $\varepsilon$ remains
within $0.01$ of the baseline.

Two noise injection modes are evaluated.
\emph{Noise-only} scales $\sigma_g$ alone, keeping $C_g$ fixed.
\emph{Joint} scales both $\sigma_g$ and $C_g$ simultaneously.
We sweep $r_g \in \{0.5, 0.75, 1.25, 1.5, 1.75, 2.0\}$ as the relative
noise reduction factor for the boosted group ($r_g = 1$ corresponds to the
Equally-weighted baseline; $r_g > 1$ implies lower noise for that group),
yielding 18 configurations per dataset ($6$ ratios $\times$ $3$ layer types).
Table~\ref{tab:noise-allocation} reports results at $r_g = 1.5$, the value
at which Conv-weighted allocation achieves peak test accuracy on both datasets;
Figure~\ref{fig:ratio_vs_acc} presents the full sweep.
MIA accuracy is averaged over 5 independent attack evaluations on CIFAR-100
and 3 on CIFAR-10, with $\pm$ denoting the 95\% confidence interval.

\paragraph{Clipping Warm-up Schedule.}
The per-layer clipping threshold $C_g$ follows a warm-up schedule during
training to accommodate the large gradient magnitudes in early epochs.
The effective threshold is set to $16 C_g$ for epochs~$0$--$50$, then
reduced to $8 C_g$, $4 C_g$, and $2 C_g$ at epochs~$50$, $70$, and $80$
respectively, reaching the nominal value $C_g$ at epoch~$100$.
This schedule is held constant across all ablation configurations so that
differences in utility arise solely from the noise and threshold settings.

\paragraph{Training Hyper-parameter Sweep.}
We ablate the number of auxiliary models $K$ and batch size $B$.
Training is reduced to 200 epochs with all other privacy settings identical
to the main experiment.
Sliding-window smoothing is applied to the privacy loss curves in
Figure~\ref{fig:ablation}.

\subsection{Empirical Privacy Audit}
\label{sec:mia}

We evaluate membership inference attack (MIA) resistance using
MetricMIA~\citep{chen2025miabench}, one of the most effective MIA methods
benchmarked for DP training.
We restrict the audit to CIFAR-10 and CIFAR-100, as these are the standard
benchmarks where non-private and DP-SGD MIA baselines are well-established;
extending MIA evaluation to ImageNet-100 and SST-2 is left for future work.

\begin{table}[H]
    \centering
    \small
    \caption{Empirical privacy audit via MIA.
    MIA accuracy near 50\% indicates near-random guessing.}
    \label{tab:mia_results}
    \begin{tabular}{ll c c c c}        \toprule
        \textbf{Dataset} & \textbf{Method} & MIA AUC & MIA Acc (\%) & Acc (\%) & $\epsPriv$ \\
        \midrule
        \multirow{3}{*}{CIFAR-10}
        & SGD    & 63.23($\pm 0.27$) & 59.0 & 92.85 & $\infty$ \\
        & DP-SGD & 56.60($\pm 1.08$) & 50.9 & 43.64 & 1.00 \\
        & DP-DT  & 56.89($\pm 0.77$) & 50.5 & 82.59 & 0.64 \\
        \midrule
        \multirow{3}{*}{CIFAR-100}
        & SGD    & 80.48($\pm 1.18$) & 82.1 & 65.44 & $\infty$ \\
        & DP-SGD & 55.28($\pm 0.25$) & 50.9 & 14.57 & 2.00 \\
        & DP-DT  & 55.70($\pm 1.09$) & 51.5 & 51.48 & 1.51 \\
        \bottomrule
    \end{tabular}
\end{table}

As shown in Table~\ref{tab:mia_results}, DP-DT's MIA accuracy is close to 50\%
(random guessing) on both datasets, comparable to DP-SGD, while maintaining
substantially higher utility.
This confirms that the empirical privacy leakage of DP-DT is consistent with
its theoretical DP guarantee under the Hidden State Assumption.

\vskip 0.2in

\end{document}